\pdfoutput=1
\documentclass[pdflatex,sn-mathphys-ay]{sn-jnl}

\usepackage[T1]{fontenc}
\usepackage{graphicx}
\usepackage{booktabs}
\usepackage{xcolor}
\usepackage{multirow}
\usepackage{amsmath,amssymb,amsfonts}
\usepackage{enumitem}
\usepackage{float}
\usepackage{tikz}
\usepackage{subcaption}
\usepackage{bm}
\usepackage{tabularx}
\usepackage{pifont}
\usepackage[most]{tcolorbox}

\newcommand{\cmark}{\ding{51}}
\newcommand{\xmark}{\ding{55}}
\newcommand{\tmark}{$\triangle$}

\definecolor{learningadapt}{RGB}{196,139,0}
\definecolor{policyadapt}{RGB}{128,82,150}
\definecolor{scenarioadapt}{RGB}{205,72,57}
\newcommand{\learnadapt}{\textcolor{learningadapt}{Learning adaptability}}
\newcommand{\policyadapt}{\textcolor{policyadapt}{Policy adaptability}}
\newcommand{\scenarioadapt}{\textcolor{scenarioadapt}{Scenario-driven adaptability}}

\setlist{nosep, topsep=0pt}
\setlist[enumerate]{label=\arabic*., font=\normalfont, itemsep=0pt, topsep=2pt, leftmargin=2.5em}
\renewcommand{\arraystretch}{1.12}
\tcbset{
  colback=gray!5,
  colframe=black,
  boxrule=0.5pt,
  arc=2pt,
  left=6pt,
  right=6pt,
  top=4pt,
  bottom=4pt,
  enhanced,
}

\begin{document}

\title[Toward Adaptable Multi-Agent Reinforcement Learning]{Toward Adaptable Multi-Agent Reinforcement Learning: An Assumption-Aware Review}

\author*[1]{\fnm{Siyi} \sur{Hu}}\email{siyi.hu@curtin.edu.au}
\author[2]{\fnm{Mohamad A.} \sur{Hady}}\email{mohamad.hady@adelaide.edu.au}
\author[3]{\fnm{Jianglin} \sur{Qiao}}\email{jianglin.qiao@sydney.edu.au}
\author[2]{\fnm{Zehong} \sur{Cao}}\email{jimmy.cao@adelaide.edu.au}
\author[2]{\fnm{Mahardhika} \sur{Pratama}}\email{dhika.pratama@adelaide.edu.au}
\author[2,4]{\fnm{Ryszard} \sur{Kowalczyk}}\email{ryszard.kowalczyk@adelaide.edu.au}

\affil*[1]{\orgname{Curtin University}, \orgaddress{\city{Perth}, \state{Western Australia}, \country{Australia}}}
\affil[2]{\orgname{Adelaide University}, \orgaddress{\city{Adelaide}, \state{South Australia}, \country{Australia}}}
\affil[3]{\orgname{The University of Sydney}, \orgaddress{\city{Sydney}, \state{New South Wales}, \country{Australia}}}
\affil[4]{\orgname{Systems Research Institute Polish Academy of Sciences}, \orgaddress{\city{Warsaw}, \country{Poland}}}

\abstract{Multi-Agent Reinforcement Learning (MARL) has achieved strong performance in simulated benchmarks, yet real deployments often violate the assumptions under which algorithms are designed and evaluated. Agent populations may change, objectives may shift, centralized information may be unavailable, execution may become asynchronous, and partner policies may be unfamiliar. Existing surveys discuss related desiderata such as scalability, robustness, generalization, and transferability, but these terms often refer to different objects of analysis and different kinds of distributional or structural shift.
This survey proposes \textit{adaptability} as an assumption-aware taxonomy for organizing these shifts, rather than as a universal requirement that every MARL algorithm should succeed in every setting. We distinguish three dimensions: \textit{learning adaptability}, which concerns the applicability of learning paradigms under changed training or system assumptions; \textit{policy adaptability}, which concerns the reuse or adaptation of learned policies under deployment-time changes; and \textit{scenario-driven adaptability}, which concerns whether benchmarks and evaluation protocols expose controlled, diagnostically useful shifts. By separating what changes, when the change occurs, what adaptation is allowed, and what success means, the framework clarifies how established concepts fit together and identifies where current MARL evaluation remains underspecified.}

\keywords{multi-agent reinforcement learning, adaptability, multi-agent systems, policy generalization, scenario-driven evaluation, reinforcement learning}

\maketitle

\section{Introduction}

Multi-Agent Reinforcement Learning (MARL) extends reinforcement learning (RL) to settings involving multiple learning agents. It has become a key framework for addressing sequential decision-making problems in real-world scenarios~\cite{busoniu2008comprehensive, feriani2021single, zhang2021multi, cui2022survey,  gronauer2022multi, orr2023multi, yuan2023survey, zhu2024survey}. Compared to single-agent RL~\cite{sutton1998reinforcement, mnih2015human, arulkumaran2017deep}, the multi-agent setting introduces additional complexities through agent interaction structures~\cite{tan1993multi, yang2018mean}, heterogeneous roles~\cite{wangrode, kuba2022heterogeneous}, and coupled observability constraints. Although partial observability also arises in single-agent reinforcement learning, MARL amplifies it because each agent's observations are shaped by other agents' hidden states, policies, intentions, and communication constraints~\cite{qmix, ganapathi2021partially}.

To address these complexities, MARL algorithms are often developed under specific structural and operational assumptions. Prominent examples include cooperative MARL under centralised training with decentralised execution (CTDE)~\cite{qmix, mappo, hatrpohappo}, mean-field methods for large agent populations~\cite{yang2018mean, guo2019learning}, offline MARL based on static datasets~\cite{yang2021believe, pan2022plan, formanek2023ogmarl}, networked MARL with local communication~\cite{zhang2018networked}, \cite{zhang2018fully}, and \cite{jiang2018graph}, model-based methods that incorporate predictive models of the multi-agent systems~\cite{willemsen2021mambpo, xu2022mingling, egorov2022scalable}, and constrained MARL frameworks for safe coordination~\cite{zhang2019mamps, liu2021cmix, gu2023safe}. While these paradigms demonstrate strong performance within their respective domains, they are often evaluated under narrowly defined or fixed conditions that align with their specific assumptions. As a result, it is difficult to assess generality or cross-paradigm robustness of a MARL algorithm.

In this work, we argue that MARL algorithms should be evaluated not only by benchmark performance, but also by the assumptions under which they remain applicable. These assumptions include fixed population size, centralized access to global information, synchronized updates, shared reward structures, stable observability patterns, and familiar partner policies. A method need not be expected to perform well when all of these assumptions are violated. Instead, its scope of applicability should be made explicit: which assumptions are essential, which can be relaxed, what form of adaptation is allowed, and how performance degradation should be interpreted. This requirement is motivated by the fact that real-world multi-agent systems are inherently dynamic. Agent populations fluctuate, objectives evolve, and execution-time constraints vary across deployments. We highlight this issue through three representative cases:

\begin{enumerate}
    \item \textit{CTDE: Coordination Without Flexibility?}: CTDE methods enable a group of agents to coordinate by optimizing centralized critics~\cite{maddpg, mappo, coma}, joint value functions~\cite{qmix, vdn, qplex}, and trust-region updates~\cite{hasac, harl, hatrpohappo}. However, they are typically restricted to fixed agent sets and full observability during training. As a result, these approaches often struggle to adapt when population structures or observability patterns change at deployment~\cite{li2022metadrive, SMARTS, xia2021multi}.

    \item \textit{Mean-field or Networked MARL: Scalable But Still Centralized?}
     Mean-field approximations~\cite{barabasi1999mean, bensoussan2013mean, yang2018mean} or networked local communication~\cite{zhang2018networked}, \cite{zhang2018fully}, and \cite{jiang2018graph} enable scalable learning by reducing agent interactions to statistical aggregates, but they all assume access to the full population during training. This assumption limits their effectiveness when centralized learning and synchronous policy update is infeasible~\cite{aoki1965optimal, hanabi, Poker}.

    \item \textit{Offline MARL: Trained Offline, Evaluated Under Shift?}
     Offline MARL offers a compelling alternative by training policies on fixed datasets~\cite{kostrikov2021offline, pan2022plan, formanek2023ogmarl}. Yet, these methods often fail to generalize in environments with unseen agent combinations or evolving dynamics. Without updated trajectories, offline policies become susceptible to extrapolation errors~\cite{fujimoto2019off, kumar2020conservative, kostrikov2021offline}.
\end{enumerate}

Existing literature has attempted to characterize desirable properties that can mitigate the issue above. Terms such as \textit{scalability}, \textit{robustness}, \textit{generalization}, and \textit{transferability} appear frequently across papers and surveys~\cite{bloembergen2015evolutionary, da2019survey, zhu2024survey, oroojlooy2023review, cui2022survey, yuan2023survey, geng2026scaling}. However, these notions often emphasize different objects of analysis. Scalability typically concerns how learning or inference behaves as the number of agents increases, but may say little about reward shifts, partner shifts, or asynchronous execution~\cite{cui2022survey, geng2026scaling}. Transferability emphasizes reusing learned knowledge across tasks or environments, but often assumes compatible agent interfaces or coordination protocols~\cite{da2019survey, yuan2023survey}. Robustness often refers to resilience under noise or perturbations, but does not always specify whether the perturbation affects the training process, the learned policy, or the benchmark design~\cite{gu2022review}. These concepts are valuable, but they need a clearer map of what changes, when it changes, and what kind of response is permitted.

In this survey, we introduce \textit{adaptability} as that organizing map. We do not use the term to claim that a MARL algorithm should remain effective under arbitrary setting changes. Rather, adaptability denotes the relationship between a MARL object of analysis (a learning paradigm, a learned policy, or an evaluation scenario) and a specified shift in assumptions. This distinction separates three interrelated dimensions:

\begin{itemize}
    \item \emph{Learning Adaptability} concerns whether a learning paradigm remains applicable when training or system-design assumptions change, such as population size, reward structure, centralized information access, or synchronization requirements.

    \item \emph{Policy Adaptability} concerns whether a learned policy can be reused, conditioned, fine-tuned, or otherwise adapted under deployment-time changes, such as novel tasks, altered agent roles, or unfamiliar partner policies.

    \item \emph{Scenario-Driven Adaptability} concerns whether environments and evaluation protocols provide controlled shifts that reveal where algorithms or policies succeed, fail, or require additional adaptation.
\end{itemize}

\begin{table}[t]
  \caption{Conceptual scope of MARL adaptability in this survey.}
  \label{tab:adaptability-scope}
  \centering
  \footnotesize
  \setlength{\tabcolsep}{3pt}
  \renewcommand{\tabularxcolumn}[1]{>{\raggedright\arraybackslash}m{#1}}
  \begin{tabularx}{\linewidth}{@{}>{\raggedright\arraybackslash}m{0.18\linewidth}>{\raggedright\arraybackslash}m{0.22\linewidth}>{\raggedright\arraybackslash}m{0.20\linewidth}>{\raggedright\arraybackslash}X@{}}
    \toprule
    \textbf{Dimension} & \textbf{Object evaluated} & \textbf{When the shift occurs} & \textbf{Diagnostic question} \\
    \midrule
    \learnadapt & Learning paradigm or algorithm class & Training, system design, or learning infrastructure & Which design assumptions must hold for the paradigm to remain trainable and applicable? \\
    \midrule
    \policyadapt & Trained policy or policy family & Deployment, transfer, or test-time reuse & Can the policy handle related unseen conditions, and what adaptation cost is required? \\
    \midrule
    \scenarioadapt & Benchmark, dataset, or evaluation protocol & Experimental design and evaluation & Does the scenario expose controlled shifts that diagnose success, failure, and assumption boundaries? \\
    \bottomrule
  \end{tabularx}
\end{table}

Together, these perspectives offer a diagnostic lens for evaluating MARL without collapsing distinct expectations into a single universal criterion. For example, an offline MARL method should not automatically be expected to perform well in an online setting; the relevant question is whether the evaluation explicitly tests offline-to-online shift, what additional interaction or fine-tuning is allowed, and which failure modes emerge when the data distribution no longer matches deployment.

This survey makes four contributions. First, it develops an assumption-aware taxonomy that distinguishes learning, policy, and scenario-driven adaptability. Second, it clarifies how existing concepts such as scalability, robustness, generalization, and transferability map onto different shift types rather than forming interchangeable labels. Third, it reviews representative MARL paradigms and policy-generalization methods through this taxonomy, with particular attention to the assumptions under which each method class remains applicable. Fourth, it identifies benchmark and reporting gaps that currently prevent systematic assessment of adaptability in dynamic multi-agent systems.

The remainder of this survey is organized as follows. Section~\ref{sec:priorwork} reviews relevant literature and situates our contribution within existing surveys. Section~\ref{sec:learning} introduces the concept of \textit{learning adaptability}, examining how different training paradigms respond to structural variation. Section~\ref{sec:policy} explores \textit{policy adaptability}, analysing how policies generalize across tasks, roles, and agent populations. Section~\ref{sec:scenario} presents \textit{scenario-driven adaptability}, discussing how benchmark environments support or hinder diagnostic evaluation under controlled shifts. Finally, Section~\ref{sec:future} outlines open research challenges and future directions for assumption-aware MARL design and evaluation, and Section~\ref{sec:conclusion} concludes with a summary of key insights.

\begin{figure}[t]
  \centering
  \includegraphics[width=\linewidth]{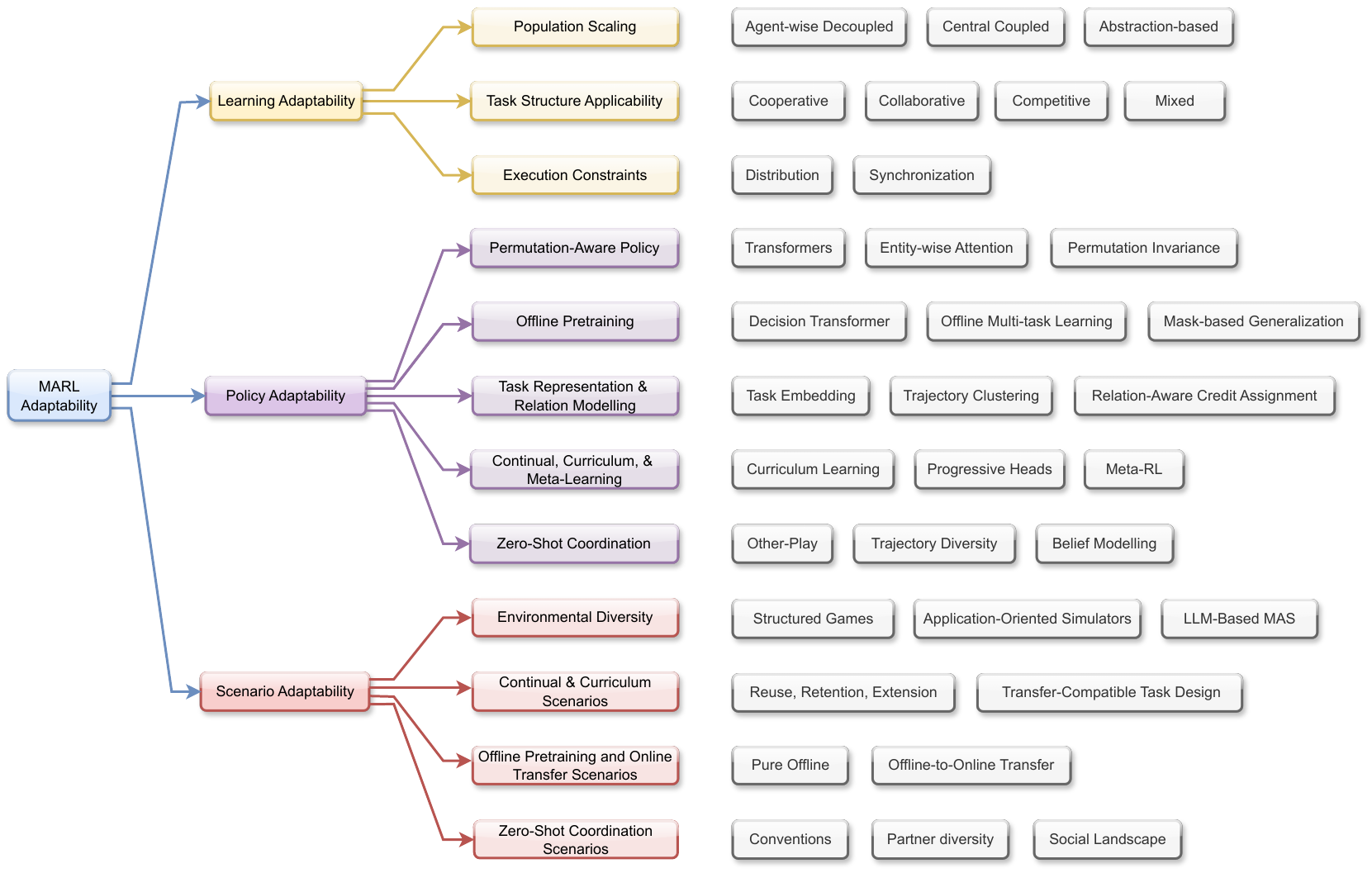} 
\caption{A structured overview of MARL adaptability across three dimensions with key topics.}
\label{fig:outline}
\end{figure}

\section{Background and Motivation}
\label{sec:priorwork}

\subsection{A Primer on MARL Paradigms}

A number of foundational paradigms structure the design of MARL algorithms.
\textit{Centralized Training with Decentralized Execution (CTDE)} is the dominant paradigm in cooperative MARL. It assumes access to global information during training, often via centralized critics~\cite{maddpg, mappo, coma} or joint value functions~\cite{qmix, vdn, qplex}, while preserving decentralized execution at test time.
\textit{Independent Learning (IL)} removes inter-agent coupling by training each agent in isolation based only on local observations~\cite{tan1993multi, de2020independent}. Despite its scalability, IL often suffers from non-stationarity and poor coordination in cooperative tasks~\cite{hernandez2017survey}.
\textit{Centralized Critics (CC)} such as MADDPG~\cite{maddpg} or MAPPO~\cite{mappo} stabilize training by conditioning policy gradients on joint observations and actions, enabling global coordination across agents~\cite{papoudakis1benchmarking, coma}.
\textit{Value Decomposition (VD)} methods, including VDN~\cite{vdn}, QTRAN~\cite{son2019qtran}, QMIX~\cite{qmix}, WQMIX~\cite{wqmix}, and QPLEX~\cite{qplex}, decompose a global value function into per-agent utilities using constraints such as monotonicity to retain coordination signal while training individual agents.
\textit{Heterogeneous-Agent (HA)} methods like HATRPO~\cite{hatrpohappo}, HAPPO~\cite{harl}, and HASAC~\cite{hasac}, stabilize training by updating agents sequentially and adjusting for changes in co-agent policies. This trust-region-based approach enables effective gradient updates even when agents have distinct roles and observation spaces.
\textit{Offline MARL} methods~\cite{yang2021believe, pan2022plan, formanek2023ogmarl} learn policies from pre-collected datasets without active exploration. They promise fast deployment but face extrapolation errors in unseen scenarios.
\textit{Mean-field MARL}~\cite{yang2018mean, guo2019learning, mondal2022approximation} approximates the influence of many agents by aggregating their behaviours into statistical summaries. This supports tractable training in large populations but often assumes homogeneity and synchronized execution~\cite{ganapathi2021partially, ganapathi2020multi}.
Other paradigms include \textit{model-based MARL}~\cite{willemsen2021mambpo, pasztorefficient, zhang2021centralized, xu2022mingling, yu2022model} and \textit{safe MARL}~\cite{zhang2019mamps, liu2021cmix, lu2021decentralized, melcer2022shield, ying2023scalable}, which prioritize sample efficiency and safety constraints, respectively.
These foundational paradigms form the basis for algorithmic exploration in MARL and serve as anchors for understanding MARL adaptability in learning and execution.

\subsection{Positioning Relative to Existing MARL Surveys}

MARL has already been surveyed from several mature perspectives. General surveys introduce core algorithms, training paradigms, and theoretical foundations~\cite{busoniu2008comprehensive, zhang2021multi, gronauer2022multi}. Other surveys focus on cooperative MARL~\cite{oroojlooy2023review}, open-environment cooperative settings~\cite{yuan2023survey}, communication mechanisms~\cite{zhu2024survey}, transfer learning~\cite{da2019survey}, large populations and scalability~\cite{cui2022survey}, or the combined difficulty of large teams and long-horizon tasks~\cite{geng2026scaling}. These works are valuable because they organize a rapidly expanding literature by algorithm family, application domain, interaction structure, or scale.

The contribution of this survey is different. We do not aim to provide the broadest catalogue of MARL algorithms, nor do we argue that adaptability supersedes existing concepts. Instead, we use adaptability to make explicit the assumption boundaries that are often implicit across these concepts. For example, a scalability-oriented method may handle more agents while still requiring homogeneous roles, synchronized execution, or access to global population statistics. A transfer-learning method may reuse knowledge across tasks while still assuming compatible observation-action interfaces. A robust policy may tolerate stochastic noise while failing under a change in partner convention or reward structure. These are not contradictions; they are different shift types.

\begin{table}[t]
  \caption{Relationship between established MARL concepts and the assumption-shift view adopted in this survey.}
  \label{tab:concept-map}
  \centering
  \footnotesize
  \setlength{\tabcolsep}{3pt}
  \renewcommand{\tabularxcolumn}[1]{>{\raggedright\arraybackslash}m{#1}}
  \begin{tabularx}{\linewidth}{@{}>{\raggedright\arraybackslash}m{0.16\linewidth}>{\raggedright\arraybackslash}m{0.24\linewidth}>{\raggedright\arraybackslash}m{0.25\linewidth}>{\raggedright\arraybackslash}X@{}}
    \toprule
    \textbf{Concept} & \textbf{Typical focus} & \textbf{Commonly underspecified boundary} & \textbf{Adaptability dimension most directly involved} \\
    \midrule
    Scalability & Performance or tractability as agent number, state space, or planning horizon grows & Whether reward structure, synchrony, agent heterogeneity, and information access also remain fixed & \learnadapt; \scenarioadapt \\ \hline
    Robustness & Stability under noise, perturbations, uncertainty, or adversarial variation & Whether the perturbation affects observations, partners, rewards, dynamics, training data, or evaluation scenarios & \learnadapt; \;  \;  \;\policyadapt \\ \hline
    Generalization & Performance on unseen tasks, layouts, agent configurations, or domains & Whether success is zero-shot, conditioned on context, obtained through fine-tuning, or achieved by retraining & \policyadapt \\ \hline
    Transferability & Reuse of learned representations, policies, or experience across tasks or environments & Whether source and target tasks share compatible interfaces, agent identities, conventions, and objectives & \policyadapt; \scenarioadapt \\ \hline
    Benchmark diversity & Availability of multiple environments, task families, and evaluation settings & Whether shifts are controlled, comparable, and diagnostic rather than merely heterogeneous & \scenarioadapt \\
    \bottomrule
  \end{tabularx}
\end{table}

Table~\ref{tab:concept-map} therefore positions adaptability as a diagnostic layer over existing terminology. The framework asks authors and evaluators to state the object being assessed, the assumption being shifted, the timing of the shift, and the permitted adaptation mechanism. This framing also prevents overgeneralization: failing under an untested or incompatible shift does not necessarily invalidate an algorithm, but it does define the boundary of its applicability.

\subsection{From Scalability to Adaptability}

Among the various terms used to describe a MARL algorithm's ability to handle diverse tasks, \textit{scalability} has long been a central focus, particularly in domains involving large agent populations or limited communication bandwidth. A broad range of methods has been proposed to address this challenge, spanning four primary directions.
\textit{Graph-based factorizations} decompose global objectives into local components using factored MDPs~\cite{guestrin2001, guestrin2003efficient}, coordination graphs~\cite{guestrin2002coordinated}, and sparse Q-learning variants~\cite{kok2004sparse, kok2006}, with extensions to partially observable settings via ND-POMDPs~\cite{nair2005networked} and factored Dec-POMDPs~\cite{oliehoek2013approximate}. Recent efforts have advanced scalability via decentralised actor-critic consensus~\cite{zhang2018networked} and \cite{zhang2018fully}, correlation decay techniques~\cite{qu2020scalable, lin2021multi}, and hypergraph-based coordination~\cite{bai2021value, zhang2022efficient}.
\textit{Mean-field approximations} simplify learning by modelling interactions through population-level statistics~\cite{yang2018mean}, enabling tractable coordination in large-scale systems~\cite{zheng2018magent}, with extensions to partially observable~\cite{subramanian2020partially} and heterogeneous-agent environments~\cite{ganapathi2020multi}. Theoretical advances in mean-field control~\cite{carmona2019model, gu2021mean} further recast the MARL problem as a single-agent control problem in the large-agent limit.
\textit{Swarm intelligence} approaches, inspired by biological collectives, rely on decentralised heuristics for search and coordination~\cite{brambilla2013swarm}. Benchmarks such as MAgent~\cite{zheng2018magent} and Neural MMO~\cite{suarez2023neural} demonstrate emergent behaviours among thousands of agents. While inherently scalable, swarm-based methods typically require heavy domain-specific reward shaping and are often task-specific, with limited generalisability across scenarios.
Although these methods have advanced scalability in terms of sample efficiency and population size, they often assume static settings in other aspects such as agent roles, task objectives, and interaction patterns—conditions that rarely hold in real-world application~\cite{li2022metadrive, stephenson2024bsk, wang2021multi}. When settings become dynamic and learning objective starts shifting, these algorithms frequently exhibit performance degradation or behavioural instability~\cite{updet, iqbal2021randomized, jianye2022boosting}.

To address this gap, we treat scalability as one important case within a broader assumption-shift perspective. Unlike scalability, which primarily addresses computational or representational efficiency as the agent count grows, adaptability asks which assumptions are being changed, which object is being evaluated, and what form of response is permitted. In the following sections, we describe MARL adaptability via three interrelated dimensions: \textit{Learning Adaptability}, \textit{Policy Adaptability}, and \textit{Scenario-Driven Adaptability}, which together provide a more explicit basis for assessing the scope and limits of MARL methods.
\footnote{This survey on MARL adaptability is based on algorithms published in top-tier venues such as NeurIPS, ICML, ICLR, AAAI, TNNLS, JMLR, TPAMI, and AAMAS, primarily from 2017 onward. Rather than providing an exhaustive enumeration, we focus on well-cited foundational works and representative recent approaches.}

\section{Learning Adaptability}
\label{sec:learning}

We begin by examining \textit{learning adaptability}, which concerns the applicability of a MARL learning paradigm or algorithm class when training and system-design assumptions change. The object of analysis here is not a single trained policy, but the learning setup itself: the information available during training, the form of the reward structure, the population or interaction topology, and the synchronization or infrastructure assumed by the algorithm. A learning paradigm is more adaptable when it can be instantiated under a wider range of such assumptions with limited redesign, or when its required redesign can be made explicit. This dimension therefore addresses four diagnostic questions:

\begin{tcolorbox}[title=Key Questions for Learning Adaptability]
\begin{enumerate}
    \item \textsf{What happens to trainability, stability, and coordination cost when the agent population or interaction topology changes?}
    \item \textsf{Which reward and objective structures does the paradigm assume: cooperative, collaborative, competitive, or mixed?}
    \item \textsf{Does training require centralized information, synchronized updates, fixed identities, or shared parameters?}
    \item \textsf{When these assumptions change, what response is required: no modification, parameter tuning, architectural redesign, data recollection, or a different algorithm class?}
\end{enumerate}
\end{tcolorbox}

To systematically evaluate these questions, we analyse representative MARL paradigms across three shifting axes:

\begin{itemize}
    \item \textit{Population and Topology Shifts} (Sec.~\ref{sec:population}): Examines how learning paradigms cope with growing, shrinking, or structurally different agent populations and the associated scalability-coordination trade-offs.
    \item \textit{Applicability Across Task Structures} (Sec.~\ref{sec:objectives}): Investigates which reward and interaction assumptions are built into cooperative, collaborative, competitive, and mixed settings.
    \item \textit{Execution Constraints on Distribution and Synchronization} (Sec.~\ref{sec:exec_constraints}): Assesses whether training remains feasible under distributed infrastructure, limited global information, or asynchronous updates.
\end{itemize}

Each axis highlights a different assumption boundary. This focus separates learning adaptability from policy adaptability: a paradigm may be trainable across several problem classes even if each trained policy remains task-specific, and a policy may transfer across related tasks even if its original training procedure depends on restrictive assumptions.

\begin{figure}[t]
  \centering
  \includegraphics[width=0.95\linewidth]{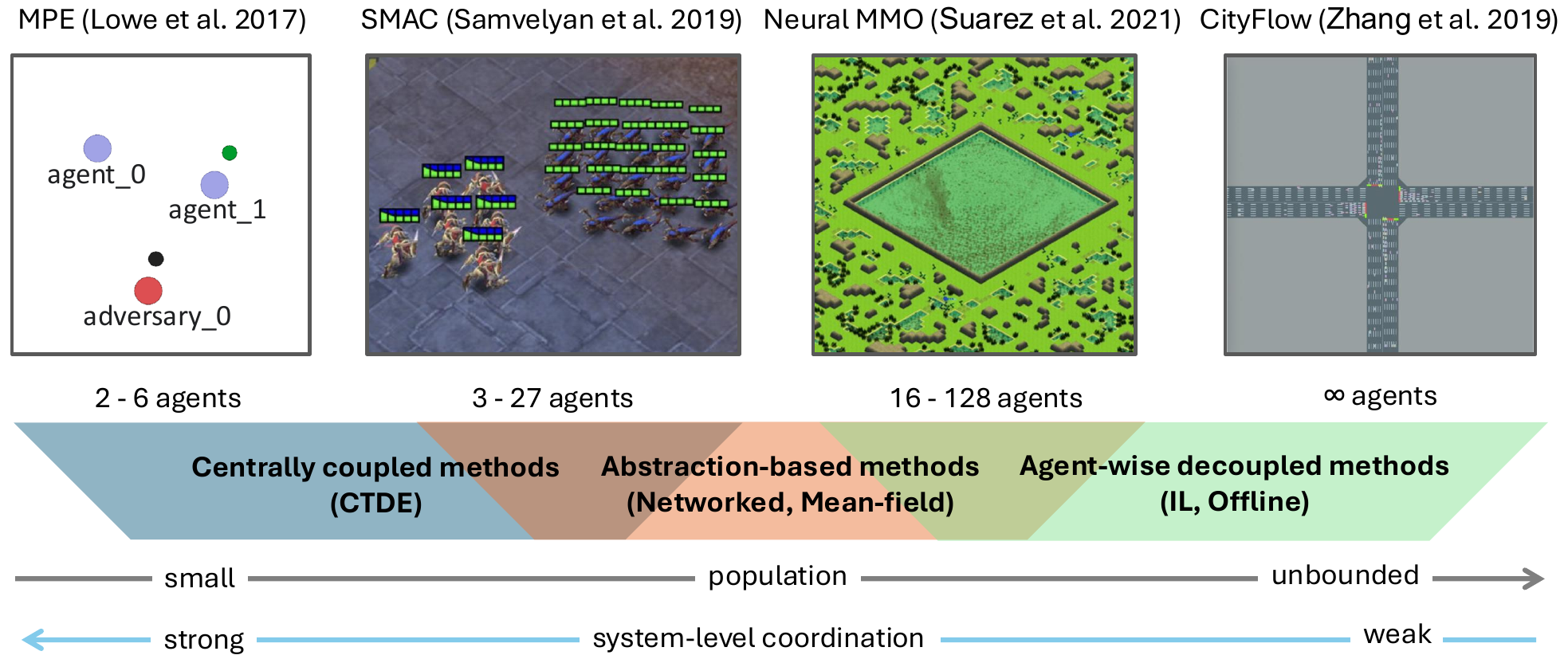}
  \caption{MARL environments exhibit significant variability in agent population scale, ranging from small-scale settings like MPE~\cite{maddpg} (2--6 agents) and SMAC~\cite{samvelyan2019starcraft} (3--27 agents), to large-scale simulations such as Neural MMO~\cite{suarez2023neural} (16--128 agents) and CityFlow~\cite{zhang2019cityflow} (unbounded). As agent populations increase, the degree of system-level coordination typically decreases, as interactions become sparser and more local. This trend also reflects the strengths and limitations of different MARL paradigms: centrally coupled methods offer high coordination fidelity but struggle to scale; agent-wise decoupled methods scale well but exhibit weaker coordination; abstraction-based methods offer a middle ground, supporting moderate scalability while preserving coordination structure.}
  \label{fig:population}
\end{figure}

\subsection{Population Scaling}
\label{sec:population}

A critical dimension of learning adaptability in MARL is the population and topology regime under which a learning paradigm remains trainable. As agent populations grow, the dimensionality of the joint observation-action space increases exponentially, making coordination increasingly difficult and expensive. In practice, environments with large agent populations often exhibit weaker system-level coordination, as the density of agent interactions becomes sparser and more localized. This variability introduces structural tensions between scalability and coordination fidelity, which must be carefully balanced by algorithmic design.

To analyse these trade-offs, we group MARL learning paradigms into three representative strategies based on how they structure agent interactions during training: (i) \textit{centrally coupled optimization}, which enforces strong coordination via centralized critics or shared objectives; (ii) \textit{agent-wise decoupling}, which emphasizes scalability by training agents independently or with minimal coupling; and (iii) \textit{coordination via abstraction}, which approximates or restricts inter-agent dynamics using statistical or topological simplifications. These categories align with different points on the scalability-coordination spectrum, as illustrated in Fig.~\ref{fig:population}.

\textit{Centrally coupled} methods embed inter-agent dependencies directly into the learning objective. Instantiated within the CTDE paradigm, they perform well in small to moderate populations~\cite{samvelyan2019starcraft, kurach2020google}, where coordination is critical. Value Decomposition (VD) methods such as QMIX~\cite{qmix} and QPLEX~\cite{qplex} aggregate individual Q-values via monotonic mixing networks, enabling joint value estimation but incurring a scalability bottleneck. Reward Decomposition (RD) methods like COMA~\cite{coma} or SHAQ~\cite{wang2022shaq} offer fine-grained credit assignment using counterfactual baselines or Shapley values, but scale poorly due to their combinatorial complexity. Centralized Critic (CC) methods such as MADDPG~\cite{maddpg} and MAPPO~\cite{mappo} employ global critics that suffer from input explosion in large teams. Heterogeneous Agent (HA) approaches like HATRPO~\cite{hatrpohappo} reduce gradient interference via sequential updates but sacrifice sample efficiency in large populations.

\textit{Agent-wise decoupled} strategies remove joint training dependencies, enabling scalable and distributed learning. Independent Learning (IL) methods like IQL~\cite{tan1993multi} and IPPO~\cite{de2020independent} optimize policies based solely on local observations, allowing linear scalability. However, they typically struggle to learn coordinated behaviours in tightly coupled environments~\cite{papoudakis1benchmarking}. Offline MARL methods such as ICQ-MA~\cite{yang2021believe} and OMAR~\cite{pan2022plan} train from static datasets with no online interaction, but coordination must be implicitly embedded in the data. These methods scale well to large systems, yet often yield weaker global coordination.

\textit{Abstraction-based} methods offer a middle ground, enabling partial coordination at scale via statistical or graph-based approximations. Mean-field MARL~\cite{yang2018mean, anand2026graphon} represents agent interactions using population-level action statistics, achieving tractable learning in large homogeneous systems~\cite{zheng2018magent}. Networked MARL~\cite{zhang2018networked} introduces graph-structured communication or critic sharing, allowing agents to learn local coordination strategies. These abstractions reduce learning complexity but may limit coordination fidelity if assumptions on homogeneity or topology are violated~\cite{du2021learning}.

In summary, population scaling in MARL reveals a fundamental trade-off: centrally coupled methods promote strong coordination but struggle with scalability; agent-wise decoupled methods scale to large systems but often underperform in cooperative tasks; abstraction-based methods strike a balance, offering scalable learning while maintaining a degree of coordination structure. These trade-offs highlight the need for adaptive designs that can modulate between scalability and coordination demands as agent populations grow.

\subsection{Applicability Across Task Structures}
\label{sec:objectives}

While most MARL algorithms are designed with specific task structures in mind, real-world applications often present diverse or evolving inter-agent reward schemes. These settings span a spectrum from fully cooperative to competitive, collaborative, and mixed interactions. Learning adaptability does not require a single paradigm to perform optimally across all task types. It instead asks which objective assumptions are built into the paradigm and how much redesign is needed when those assumptions no longer match the target setting.

\begin{figure}[t]
  \centering
  \includegraphics[width=0.9\linewidth]{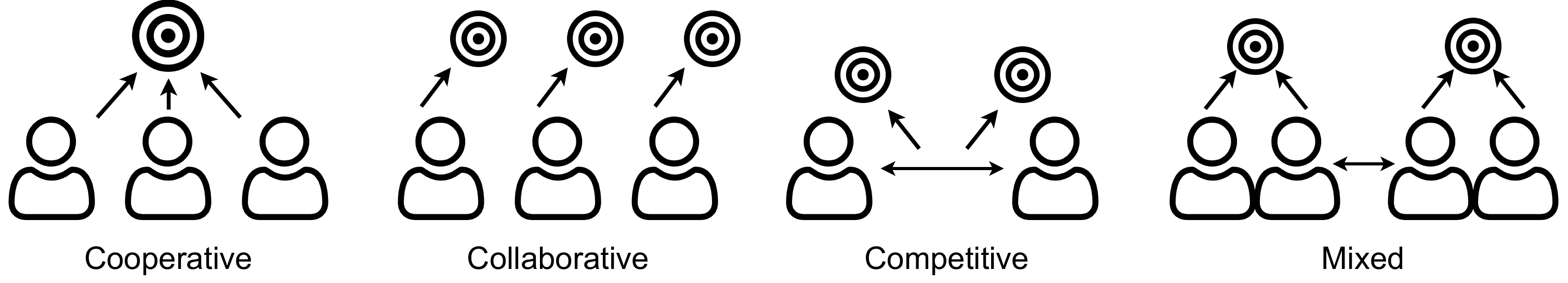} 
\caption{Illustration of four task modes in multi-agent systems categorized by reward structures: (a) Cooperative tasks where all agents share an identical global reward; (b) Collaborative tasks with similar but individual rewards for each agent; (c) Competitive tasks characterized by zero-sum individual rewards; and (d) Mixed tasks combining intra-team shared rewards and inter-team zero-sum competition. These distinctions highlight varying coordination and conflict dynamics fundamental to multi-agent learning.}
\label{fig:4taskmodes}
\end{figure}

To assess this axis of adaptability, we focus on four canonical task modes, summarized in Fig.~\ref{fig:4taskmodes}. Each task mode imposes different coordination and conflict dynamics and highlights distinct inductive biases in algorithm design. We evaluate the alignment of each paradigm by categorizing its suitability as: (i) \textit{natively suitable}, (ii) \textit{broadly applicable}, (iii) \textit{usable with task adjustment}, or (iv) \textit{incompatible}.

\textit{Fully cooperative tasks} require agents to optimize a shared global reward. CTDE paradigms including VD~\cite{qmix, vdn, facmac}, RD~\cite{wang2020shapley, wang2022shaq, li2025nucleolus}, CC~\cite{maddpg, coma, mappo}, and HA~\cite{hatrpohappo, harl, hasac} are natively suitable due to their tight coupling between agent policies and joint objectives~\cite{samvelyan2019starcraft, papoudakis1benchmarking}. Abstraction-based methods such as mean-field MARL~\cite{yang2018mean, ganapathi2021partially, carmona2023model} and networked architectures~\cite{foerster2016learning, jiang2018graph, zhang2018networked} also support effective cooperation in large-scale settings.

\textit{Collaborative tasks} exhibit partially aligned objectives and often decentralized observability. IL methods~\cite{tan1993multi, de2020independent} and networked MARL naturally support such settings, leveraging local policy optimization~\cite{zhang2018fully} and \cite{zhang2018networked} and communication graphs~\cite{sukhbaatar2016learning, du2021learning, sheng2022learning}. CC and mean-field methods are broadly applicable but may require task-specific reward shaping. VD and RD methods, which rely on a globally shared reward signal, are generally unsuitable for collaboration without significant modifications.

\textit{Competitive tasks} involve directly conflicting goals across agents or teams. Model-based MARL~\cite{zhang2021centralized, xu2022mingling, yu2022model} is natively suited to these settings, offering capabilities for planning and opponent modelling. Offline MARL~\cite{yang2021believe, pan2022plan} can be conditionally effective if adversarial interactions are well-represented in the data. IL and mean-field methods are usable only in restricted forms and typically lack mechanisms for anticipating adversarial strategies. CTDE methods are fundamentally misaligned with competitive settings due to their cooperative training assumptions.

\textit{Mixed cooperative-competitive tasks} require intra-team coordination and inter-team competition. Model-based MARL can be applicable when its planning or opponent-modelling assumptions match the task, supporting hierarchical reasoning across team boundaries. CTDE methods can be applied within teams but are limited in generalizing across teams with competing objectives. Networked MARL can support such hybrid scenarios when communication structures are carefully defined. Offline MARL retains conditional applicability, provided task transitions are present in the training data.

Overall, the suitability of a MARL paradigm is closely tied to the task structure. Although no paradigm is universally optimal, a clear account of where structural or algorithmic changes become necessary is one of the key markers of learning adaptability.

\subsection{Execution Constraints on Distribution and Synchronization}
\label{sec:exec_constraints}

In real-world deployments, MARL algorithms must address not only the challenges of coordination and learning, but also the system-level constraints imposed by practical execution environments. Two constraints are particularly prominent. First, \emph{distributed training infrastructure} is often required, where agents are deployed across physically or logically disjoint computational nodes. This arises in settings such as edge-computing robotics, networked sensors, and geographically dispersed systems, necessitating training paradigms that minimize centralized dependencies. Second, many applications involve \emph{asynchronous execution}, where agents operate at different temporal resolutions or update frequencies due to hardware heterogeneity, task allocation, or communication latency. These constraints define whether a learning paradigm is feasible under deployment-like training conditions, independently of whether a final policy later transfers to new tasks.

\textit{Distributed Training Support.}
Distributed training enables agents to learn in parallel across decentralized systems, reducing computational bottlenecks and accommodating bandwidth or privacy constraints. IL algorithms~\cite{tan1993multi, de2020independent} are fully compatible with distributed training: agents optimize local policies using local observations, without parameter sharing or joint critics. Networked MARL~\cite{zhang2018networked, foerster2016learning, shaik2026constrainedcommunication} also supports distributed learning through peer-to-peer message passing or consensus-based updates, requiring no centralized coordination.
In contrast, most CTDE methods, such as VD, CC, and HA, are only partially compatible. These algorithms permit decentralized execution but require centralized training inputs or synchronized gradient updates. This limits their deployment in bandwidth-constrained or privacy-sensitive systems. Model-based MARL~\cite{zhang2021model}, mean-field MARL~\cite{yang2018mean}, and offline MARL~\cite{pan2022plan} generally assume access to global state information or joint trajectories during training, making them incompatible with fully distributed infrastructure.
\begin{table}[t!]
  \caption{Learning adaptability of MARL paradigms across seven dimensions: Pop. Scal. = Population Scaling, Coop. = Cooperative, Collab. = Collaborative, Comp. = Competitive, Dist. Train. = Distributed Training, Async. Exec. = Asynchronous Execution.}
  \label{tab:paradigm-adaptability}
  \centering
  \begingroup
  \scriptsize
  \renewcommand{\arraystretch}{1.18}
  \setlength{\tabcolsep}{1.5pt}
  \begin{tabular*}{\linewidth}{@{\extracolsep{\fill}}lccccccc@{}}
    \toprule
    \textbf{Paradigm} & \shortstack{\textbf{Pop.}\\\textbf{Scal.}} & \textbf{Coop.} & \textbf{Collab.} & \textbf{Comp.} & \textbf{Mixed} & \shortstack{\textbf{Dist.}\\\textbf{Train.}} & \shortstack{\textbf{Async.}\\\textbf{Exec.}} \\
    \midrule
    Value Decomposition (VD)   & \xmark & \cmark & \xmark & \xmark & \xmark & \xmark & \xmark \\
    Reward Decomposition (RD)  & \xmark & \cmark & \xmark & \xmark & \xmark & \xmark & \xmark \\
    Centralized Critic (CC)    & \tmark & \cmark & \cmark & \cmark & \cmark & \tmark & \tmark \\
    Heterogeneous Agent (HA)   & \tmark & \cmark & \xmark & \xmark & \xmark & \tmark & \xmark \\
    Independent Learning (IL)  & \cmark & \cmark & \cmark & \cmark & \cmark & \cmark & \cmark \\
    Offline MARL              & \tmark & \tmark & \tmark & \tmark & \tmark & \xmark & \tmark \\
    Model-Based MARL          & \tmark & \cmark & \cmark & \cmark & \cmark & \xmark & \tmark \\
    Mean-Field MARL           & \cmark & \cmark & \cmark & \xmark & \cmark & \xmark & \xmark \\
    Networked MARL            & \cmark & \xmark & \cmark & \xmark & \tmark & \cmark & \xmark \\
    \bottomrule
  \end{tabular*}
  \endgroup
  \vspace{1.2em}
  {\footnotesize \textbf{Legend:} \cmark — natively suitable, \tmark — partially suitable or task-dependent, \xmark — incompatible.
  }

\end{table}

\textit{Asynchronous Execution Support.}
Asynchronous adaptability is crucial in environments where agents act independently, receive delayed observations, or are triggered by event-driven processes. IL methods are natively asynchronous: agents update independently and require no shared timing, making them well-suited for real-time robotic and sensor network applications~\cite{de2020independent}. CC methods, such as MADDPG~\cite{maddpg}, can be adapted to asynchronous settings by decoupling critic evaluations from agent policy updates. Similarly, model-based MARL may support asynchronous dynamics if agent-specific transition functions are modelled separately.
Offline MARL methods exhibit asynchronous compatibility during deployment, as policy inference does not require synchronized execution. However, their training-phase adaptability depends on the structure of the offline dataset: if trajectories reflect synchronized execution, learned policies may inherit synchrony assumptions~\cite{meng2023offline}. Most other paradigms including VD, HA, mean-field, networked MARL, and safe MARL assume synchrony for value aggregation, communication, or constraint satisfaction, and thus fail in environments with variable update rates or latency~\cite{zhang2019mamps, gu2023safe}.

In summary, execution adaptability varies widely. IL and networked MARL offer the greatest flexibility for distributed, asynchronous deployment. CTDE methods are effective under centralized infrastructure but require careful scheduling. Model-based and offline approaches provide partial compatibility, contingent on problem structure and data assumptions.

\section{Policy Adaptability}
\label{sec:policy}

While learning adaptability focuses on the assumptions of the training procedure, it does not fully capture the behaviour of the \textit{learned policy} after training. In real-world deployments, a policy may encounter altered task specifications, changed agent roles, different population sizes, or unfamiliar partner behaviours. This motivates the second axis of our framework: \textit{policy adaptability}, which concerns the conditions under which a trained policy or policy family can be reused, conditioned, fine-tuned, or updated after the original training phase.

\begin{figure}[t]
  \centering
  \includegraphics[width=0.95\linewidth]{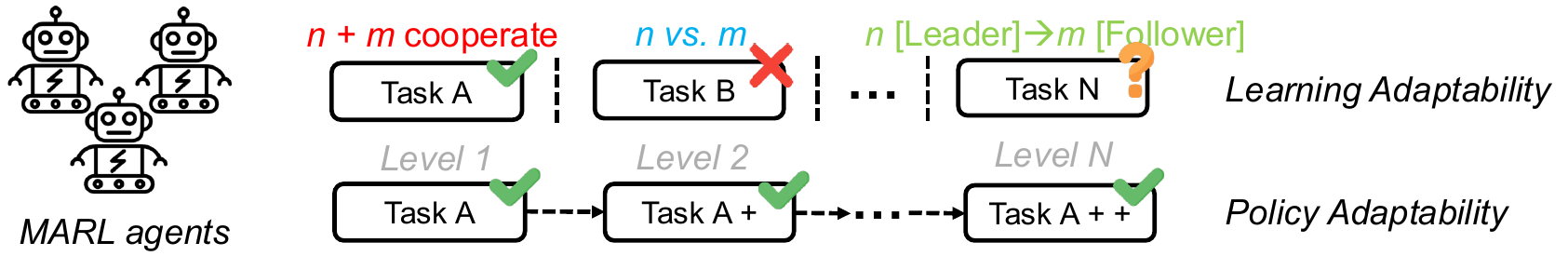}
  \caption{Illustration of the key distinction between learning adaptability and policy adaptability. Learning adaptability evaluates whether a learning paradigm can be instantiated under different training or system assumptions (e.g., Task A, B, C). Policy adaptability instead evaluates whether a trained policy (e.g., from Task A) can handle related deployment-time shifts (e.g., Task A+, A++) under a specified adaptation-cost level.}
  \label{fig:policy_vs_learning}
\end{figure}

Figure~\ref{fig:policy_vs_learning} illustrates the conceptual distinction between learning and policy adaptability. Learning adaptability asks whether a learning paradigm can be instantiated under multiple training or system assumptions. Policy adaptability instead asks whether the resulting policy can handle a specified deployment-time shift, and what adaptation cost is required. This distinction is important because a paradigm may be learning-adaptable but produce task-specific policies, while a policy may transfer effectively even if its original training process used restrictive assumptions.
The following questions are central to understanding policy adaptability:

\begin{tcolorbox}[title=Key Questions for Policy Adaptability]
\begin{enumerate}
\item \textsf{Task and dynamics shift: can a trained policy handle related tasks with changed objectives, layouts, or transition dynamics?}
\item \textsf{Population and role shift: can the policy handle different numbers of agents, altered roles, or changed observation-action interfaces?}
\item \textsf{Partner and convention shift: can the policy coordinate with independently trained or previously unseen agents?}
\item \textsf{Adaptation cost: does success require zero-shot transfer, context conditioning, fine-tuning, online adaptation, or full retraining?}
\end{enumerate}
\end{tcolorbox}

\begin{figure}
    \centering
    \includegraphics[width=0.95\linewidth]{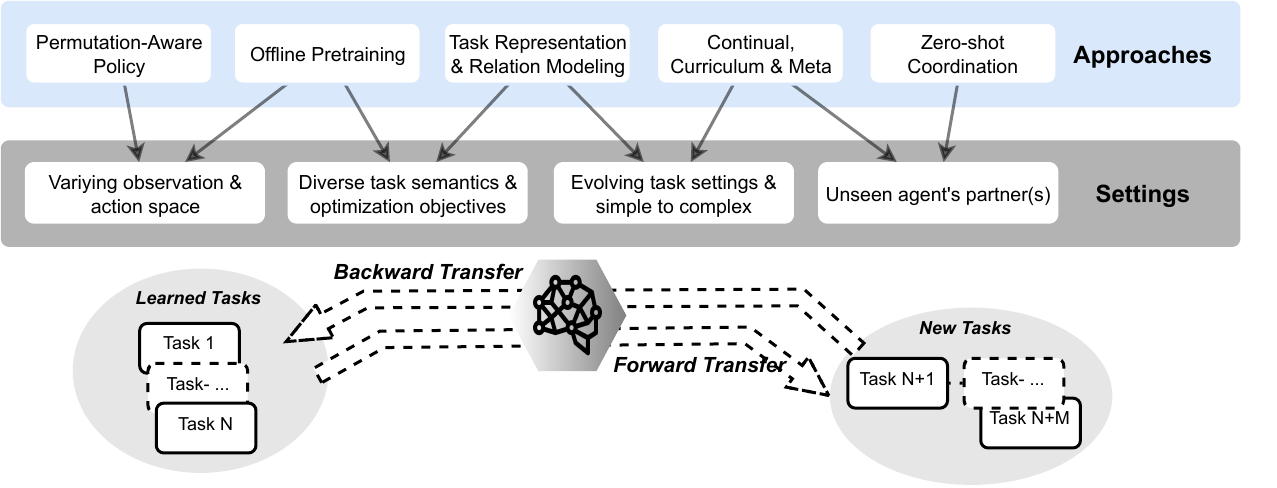}
    \caption{Relationship between policy-level shifts and available approaches. A policy-adaptability evaluation should specify the target gap, such as task, population, role, or partner variation, and the adaptation cost allowed when moving beyond the original training setting.}
    \label{fig:policy-adaptability}
\end{figure}

These questions can also be viewed from the perspective of task gaps. When task semantics lie within a well-defined space, acquired knowledge may be transferred across related settings. However, the evaluation should specify the allowed adaptation mechanism. Table~\ref{tab:policy-adaptation-cost} distinguishes several common levels. Representative gaps include changes in objectives or environmental settings (task generalization), changes in the number or roles of agents (population and interface generalization), sequential or incremental task progressions (lifelong learning), and the presence of novel partners with unfamiliar behaviours (partner generalization). As illustrated in Figure~\ref{fig:policy-adaptability}, different methods are suited to bridging different types of task gaps.

\begin{table}[t]
  \caption{Adaptation-cost levels for policy adaptability.}
  \label{tab:policy-adaptation-cost}
  \centering
  \footnotesize
  \setlength{\tabcolsep}{3pt}
  \renewcommand{\tabularxcolumn}[1]{>{\raggedright\arraybackslash}m{#1}}
  \begin{tabularx}{\linewidth}{@{}>{\raggedright\arraybackslash}m{0.22\linewidth}>{\raggedright\arraybackslash}m{0.32\linewidth}>{\raggedright\arraybackslash}X@{}}
    \toprule
    \textbf{Level} & \textbf{What is allowed} & \textbf{Interpretation} \\
    \midrule
    Zero-shot reuse & Fixed policy parameters and no target-task data & Strongest form of policy adaptability, but only meaningful when the target shift is explicitly specified. \\
    \midrule
    Context conditioning & Inference-time task, agent, history, or partner context without gradient updates & Tests whether the policy has learned a reusable conditional representation. \\
    \midrule
    Fine-tuning or online adaptation & Additional target-environment interaction, gradient updates, or policy updates & Measures adaptation efficiency rather than pure zero-shot transfer. \\
    \midrule
    Full retraining & Training a new policy from the target setting & Defines a boundary case; useful for comparison but not evidence that the original policy was adaptable. \\
    \bottomrule
  \end{tabularx}
\end{table}

To explore these challenges, we organize the literature into five methodological categories that explicitly promote generalization across tasks and agents.

\begin{itemize}
    \item \textit{Permutation-Aware Policy} (Sec.~\ref{sec:arch_foundation}): Investigates invariant and equivariant architectures that decouple policy learning from agent identity and ordering.
    \item \textit{Offline Pretraining} (Sec.~\ref{sec:offline_pretrain}): Leverages static datasets and decision transformers to encode generalized behavioural priors.
    \item \textit{Task Representation and Relation Modelling} (Sec.~\ref{sec:task_representation}): Uses task embeddings and trajectory-based structure discovery to support transfer across task distributions.
    \item \textit{Continual, Curriculum, and Meta-Learning} (Sec.~\ref{sec:curriculum_meta}): Employs progressive training schemes and fast adaptation mechanisms to support lifelong learning in evolving environments.
    \item \textit{Zero-Shot Coordination} (Sec.~\ref{sec:zsc}): Focuses on social generalization, enabling agents to align with unfamiliar teammates without shared training history.
\end{itemize}

Collectively, these approaches provide mechanisms for policy reuse under specified shifts. Their relevance depends on which shift is being tested and whether the evaluation permits zero-shot reuse, context conditioning, fine-tuning, online adaptation, or retraining.

\subsection{Permutation-Aware Policy}
\label{sec:arch_foundation}

A foundational component of policy adaptability is the architectural capacity to represent coordination patterns independent of agent identities or ordering. Recent work has converged on three core principles for enabling permutation-aware generalization in multi-agent policy models: the use of transformer architectures, entity-wise attention mechanisms, and explicit enforcement of permutation invariance or equivariance.

\textit{Transformers for Variable Agent Inputs.}
Transformer-based policy models offer a natural framework for processing variable-length, unordered agent observations. One representative approach is UPDeT~\cite{updet}, which introduces policy decoupling through transformers. By representing agent observations and actions as sets of entities, UPDeT processes these inputs via a shared attention backbone, allowing the same policy model to generalize across tasks with different numbers and types of agents. This design obviates the need for handcrafted task-specific modules and enables multi-task generalization from a unified architecture.

\textit{Entity-Wise Attention and Factorization.}
Entity-centric models extend this paradigm by isolating reusable interaction patterns across agents. For instance, Randomized Entity-wise Factorization~\cite{iqbal2021randomized} partitions input observations into semantic substructures, enabling policies to attend selectively to relevant context while maintaining awareness of task dynamics. Such designs decouple agent-specific variability from shared coordination signals, facilitating transfer across scenarios with heterogeneous inputs and team structures.

\textit{Permutation Invariance and Equivariance.}
Several models go further by explicitly encoding permutation symmetry into the policy architecture. Dynamic Permutation Networks~\cite{jianye2022boosting} and HGAP~\cite{lin2024hgap} construct neural modules whose outputs are invariant or equivariant to reordering of agents. These models combine modular feature extractors with hypernetworks or graph attention layers to enforce symmetry constraints. The resulting policies maintain consistent behaviour under arbitrary agent re-indexing, a crucial property for generalization to unseen team compositions.

Together, these architectural strategies provide the structural groundwork for adaptable policies. By abstracting away agent identity and focusing on relational or set-based reasoning, permutation-aware policies form the basis for downstream generalization in offline learning, transfer learning, and zero-shot coordination scenarios.

\subsection{Offline Pretraining}
\label{sec:offline_pretrain}

Offline pretraining offers a promising pathway to improve policy adaptability by learning from diverse multi-task datasets without requiring interactive environment access~\cite{jeon2026stairsformer}. These approaches aim to extract transferable structure from offline trajectories and deploy a unified model across tasks or agent populations. Recent work has introduced techniques that vary along three key axes: decision-transformer modelling, offline multi-task learning architectures, and mask-based generalization mechanisms.

\textit{Decision Transformer for Sequence modelling.}
A central innovation in this space is the application of decision transformer architectures to multi-agent settings. Multi-Agent Decision Transformers (MADT)~\cite{meng2023offline} extend autoregressive models to encode multi-agent trajectories conditioned on prior state, action, and return sequences. These models are pretrained over offline datasets collected from diverse tasks, allowing them to generalize behaviour without explicit fine-tuning. The sequence modelling framework allows policies to reason over temporal dependencies and supports compositional reuse of learned behaviours across task boundaries.

\textit{Offline Multi-Task Learning with Task Conditioning.}
Beyond raw sequence modelling, offline pretraining methods increasingly leverage modularity to enhance task transfer. M3~\cite{meng2023m3} introduces a task-conditioned architecture that incorporates explicit prompts and agent-invariant embeddings. It uses a vector-quantized variational autoencoder (VQ-VAE)~\cite{van2017neural} to encode heterogeneous agent roles and supports a decoupled representation of shared skills and task-specific behaviours. Similarly, hierarchical frameworks~\cite{liu2025learning} learn decomposable sub-policies, enabling selection of both generic and task-tailored strategies at inference time. These models facilitate generalization to novel tasks by encoding structured latent representations of skill and role.

\textit{Mask-Based Generalization Across Agent Variability.}
To accommodate input and output variability across tasks, recent work introduces masking strategies that condition the policy on different subsets of observation-action dimensions. MaskMA~\cite{liumaskma} combines transformers with mask-based training to enable a single model to handle agents with differing roles, modalities, or interface structures. This approach supports strong zero-shot transfer, as the model implicitly learns a general action representation that is robust to agent-specific permutations and dimensionality shifts.

Collectively, these methods demonstrate that offline pretraining can serve as a foundation for adaptable MARL. By integrating decision sequence modelling, modular task representation, and input masking, they enable broad generalization across agent configurations and task domains with minimal reliance on online fine-tuning.

\subsection{Task Representation and Relation Modelling}
\label{sec:task_representation}

A growing body of work enhances policy adaptability by constructing explicit representations of tasks and inter-agent relationships. Rather than relying solely on architecture or data scale, these methods aim to encode transferable structure across tasks, enabling more efficient reuse and generalization. Key directions include task embedding, trajectory-based clustering, and credit assignment strategies that model inter-task and inter-agent relations.

\textit{Task Embedding for Policy Conditioning.}
Task embedding methods learn compact representations of task identity or context and use them to modulate policy execution. These embeddings may capture global task features~\cite{schafer2023learning} or agent-specific roles~\cite{li2024multi}, and can be inferred from initial observations or support trajectories. By conditioning policy outputs on these learned embeddings, agents can adapt behaviour with minimal retraining. Methods such as subtask encoders~\cite{tian2023decompose} decompose complex objectives into reusable latent components, supporting structured policy reuse across tasks with shared skill dependencies.

\textit{Trajectory Clustering for Transferable Structure.}
Another line of work clusters past trajectories to uncover recurring coordination patterns or latent task variations. This allows agents to identify and transfer strategies from similar prior experiences. For example, unsupervised task discovery via trajectory distributions~\cite{na2025trajectory} enables zero-shot generalization by matching new tasks with similar previously encountered scenarios. These approaches can also be integrated with transformer-based regret modelling~\cite{zhu2024multi}, where clustering guides adaptive loss weighting across tasks of varying difficulty or progression rates.

\textit{Relation-Aware Credit Assignment.}
To further facilitate generalization, several methods incorporate relational reasoning across tasks and agents. By modelling similarity between task objectives or environmental dynamics~\cite{qin2024multi, akula2026asalt}, policies can selectively transfer knowledge between aligned domains while avoiding negative transfer. Relation-aware credit assignment~\cite{yu2024relation} extends this idea to cooperative settings, ensuring reward attribution reflects task-specific dependencies and agent roles. This relational perspective supports more robust policy updates across heterogeneous or evolving task structures.

Together, these methods enhance policy adaptability by enabling agents to extract, represent, and reason over latent structure in multi-task settings. By leveraging task embeddings, trajectory-based clustering, and relation-aware reasoning, they provide scalable mechanisms for transfer and generalization in MARL.

\subsection{Continual, Curriculum, and Meta-Learning}
\label{sec:curriculum_meta}

Another pathway to policy adaptability is through mechanisms that support dynamic policy evolution over time. These methods approach generalization as a process of continual refinement, where agents accumulate and adapt knowledge progressively across changing task distributions. Key strategies include structured curricula, modular continual learning, and meta-learning for fast adaptation.

\textit{Curriculum Learning for Task Staging.}
Curriculum learning organizes the training process into sequences of tasks with increasing complexity, enabling agents to bootstrap performance in difficult scenarios from prior experience with simpler ones. Evolutionary curricula~\cite{long2020evolutionary} gradually introduce harder coordination challenges by modifying team compositions or interaction structures. More adaptive schemes~\cite{wang2023towards} employ contextual bandit models to dynamically construct curricula based on agent progress, supporting stable population-invariant learning under sparse rewards. 
Other approaches propose auto-curricula, where the training sequence is generated automatically. For sparse-reward environments,~\cite{chen2021variational} employs task expansion and entity progression by generating training curricula that adapt both the task configurations and the number of participating entities to promote scalable and adaptable multi-agent learning. In zero-sum games, the study in~\cite{chen2024accelerate} introduces subgame curriculum learning to accelerate the learning process, where agents progressively train on strategically selected subgames to build competence before tackling the full game complexity. PORTAL~\cite{wu2024portal} automatically generates a curriculum by learning a shared feature space across tasks, enabling it to characterize tasks based on their feature distributions and prioritize those most similar to the target task for more effective transfer and adaptation.
Moreover, curriculum learning has been used to construct policies with better scalability. A novel network architecture, Dynamic Agent-number Network (DyAN)~\cite{wang2020few} introduces transfer mechanisms across curricula, which efficiently handles varying numbers of agents through dynamic input sizing. Also, in~\cite{zhang2022automatic}, the auto-curriculum approach begins by training agents in multi-agent scenarios with a small number of agents and progressively increases the number of agents, allowing the learning process to adapt gradually to more complex interactions.
These curricula promote sample-efficient generalization and reduce the risk of unstable learning in large or heterogeneous populations.

\textit{Progressive Heads for Continual Coordination.}
Continual learning strategies often adopt modular architectures to mitigate catastrophic forgetting. Progressive task heads~\cite{yuan2024multiagent} allocate separate output modules for each task while sharing a common feature backbone, allowing new skills to be acquired without overwriting prior ones. Complementary approaches prioritize experience replay or task sampling based on hindsight-derived importance metrics~\cite{yu2023prioritized}, maintaining performance across previously encountered tasks while encouraging exploration. These methods support lifelong adaptation by explicitly managing task-specific knowledge retention and reuse.

\textit{Meta-RL for Fast Generalization.}
Meta-reinforcement learning formulations target fast policy adaptation across tasks by learning how to learn. Collaborative meta-RL~\cite{jia2022crmrl} encodes agent-type relationships to facilitate coordination transfer between heterogeneous teams, even under role or capability shifts. Transformer-based approaches~\cite{zhou2023cooperative} implement credit assignment at the coalition level, allowing agents to generalize coordination strategies under dynamic observation and action spaces. These meta-learning architectures accelerate adaptation to new tasks or partner configurations by internalizing transferable learning priors.

Collectively, curriculum learning, continual modularity, and meta-adaptation provide a framework for progressive coordination development. These methods extend policy adaptability beyond static generalization by enabling agents to grow and adapt their behaviour in response to evolving task structures and team compositions.

\subsection{Zero-Shot Coordination}
\label{sec:zsc}

Beyond task generalization, a distinct facet of policy adaptability in MARL is an agent’s ability to coordinate with unfamiliar partners without additional retraining—referred to as \textit{zero-shot coordination} (ZSC). This setting reflects real-world deployments where agents may be developed independently or trained under different assumptions. Success in ZSC requires policies that are robust to partner variability and capable of aligning with unknown conventions or behaviours.

\textit{Other-Play for Convention Robustness.}
The foundational approach to ZSC is the \textit{Other-Play} (OP) framework~\cite{hu2020other}, which addresses the failure of self-play to produce partner-compatible conventions. In symmetric cooperative settings, self-play policies often rely on arbitrary symmetry-breaking strategies (e.g., always going left), leading to coordination failure when paired with independently trained agents. OP combats this by randomizing symmetric factors during training, guiding policies toward shared, robust conventions. OP demonstrated significant gains in Hanabi, where conventional self-play policies failed to align with diverse partners.

\textit{Trajectory Diversity for Coordination Generality.}
While OP assumes known symmetries, subsequent work focuses on learning partner-compatible behaviour without explicit symmetry modelling. The \textit{Trajectory Diversity} (TrajeDi) framework~\cite{lupu2021trajectory} encourages policies that generate diverse trajectory distributions while preserving coordination potential. Using a generalized Jensen-Shannon divergence objective, TrajeDi increases robustness to unseen policies even in partially observable environments. This approach enables ZSC in more complex or asymmetric domains where coordination structure is latent or unknown.

\textit{Belief Modelling and Open-Ended Coordination.}
Recent advances incorporate explicit reasoning over partner behaviour to improve ZSC performance. The \textit{Any-Play} framework~\cite{lucas2022any} trains agents to maximize cross-play success by optimizing compatibility with policies trained under different algorithms. It combines diversity-driven self-play with policy augmentation, producing strategies resilient to inter-algorithm mismatches. Similarly, the \textit{Cooperative Open-ended Learning} (COLE) framework~\cite{li2023cooperative} models agent compatibility as a graph-theoretic problem, iteratively refining strategies via preference graphs and best-response dynamics. These methods move beyond symmetry-breaking to consider partner belief modelling and distributional alignment in cooperative learning~\cite{ni2026tomzsc}.

Together, these approaches redefine policy adaptability through the lens of social compatibility. Instead of adapting to new tasks, ZSC emphasizes generalization across agent identities, a critical capability for deployment in decentralized or mixed-agent environments.

\section{Scenario-Driven Adaptability}
\label{sec:scenario}

Learning and policy adaptability cannot be assessed by benchmark performance alone. They require evaluation settings that make the relevant assumption shifts observable and comparable. We refer to this third axis as \textit{scenario-driven adaptability}: the capacity of benchmarks, datasets, and evaluation protocols to expose controlled shifts in multi-agent settings and diagnose where learning paradigms or trained policies succeed, fail, or require additional adaptation.

This dimension emphasizes the role of \textit{evaluation design}, not merely environment variety. A benchmark may contain many tasks yet still provide weak evidence of adaptability if the shifts are uncontrolled or if the permitted adaptation mechanism is unclear. Conversely, a smaller environment family can be highly diagnostic if it systematically varies agent population, role heterogeneity, communication structure, reward type, observability, data availability, or execution timing while holding other factors fixed.

Accordingly, this section addresses the following questions central to scenario-driven adaptability:

\begin{tcolorbox}[title=Key Questions for Scenario-Driven Adaptability]
\begin{enumerate}
\item \textsf{Which assumption is intentionally shifted: population, topology, role, reward, observability, data source, partner policy, or execution timing?}
\item \textsf{Are source and target scenarios comparable enough that success or failure can be attributed to the intended shift?}
\item \textsf{What adaptation mechanism is allowed: zero-shot evaluation, context conditioning, fine-tuning, online interaction, or retraining?}
\item \textsf{Which diagnostic metric is reported: degradation curve, transfer gain, adaptation speed, retention, cross-play performance, or failure mode?}
\end{enumerate}
\end{tcolorbox}

\begin{table}[t]
  \caption{Scenario-driven evaluation protocol for adaptability.}
  \label{tab:scenario-protocol}
  \centering
  \footnotesize
  \setlength{\tabcolsep}{3pt}
  \renewcommand{\tabularxcolumn}[1]{>{\raggedright\arraybackslash}m{#1}}
  \begin{tabularx}{\linewidth}{@{}>{\raggedright\arraybackslash}m{0.18\linewidth}>{\raggedright\arraybackslash}m{0.26\linewidth}>{\raggedright\arraybackslash}X@{}}
    \toprule
    \textbf{Protocol element} & \textbf{Question} & \textbf{Example diagnostic use} \\
    \midrule
    Source condition & What was the algorithm or policy designed, trained, or selected for? & Fixed 5-agent cooperative task with centralized training. \\
    \midrule
    Target condition & What condition is evaluated after the shift? & 10-agent task, altered reward structure, unseen partners, or asynchronous execution. \\
    \midrule
    Controlled shift & Which assumption changes while other factors are held as stable as possible? & Isolate population scaling from reward changes, or partner shift from task-layout shift. \\
    \midrule
    Allowed adaptation & What response is permitted before or during evaluation? & Zero-shot reuse, context conditioning, fine-tuning, online adaptation, or retraining. \\
    \midrule
    Diagnostic measure & What evidence identifies success, degradation, or failure? & Performance drop, sample efficiency, transfer gain, retention, cross-play score, or failure taxonomy. \\
    \bottomrule
  \end{tabularx}
\end{table}

Table~\ref{tab:scenario-protocol} summarizes the protocol logic used throughout this section. The goal is not to rank benchmarks by realism alone, but to ask what kinds of assumption shifts each benchmark family can reveal. To explore these questions, we structure our review around three major themes:

\begin{itemize}
    \item \textit{Survey of MARL Benchmarks} (Sec.~\ref{sec:scenario_benchmarks}): Reviews structured games, application-oriented simulators, and emerging LLM-based systems, with attention to their configurability and evaluation fidelity.
    \item \textit{Continual and Curriculum Scenarios} (Sec.~\ref{sec:continual_curriculum}): Discusses benchmark affordances for transfer-compatible design and representational continuity across evolving task sequences.
    \item \textit{Offline Pretraining, Online Transfer, and Zero-Shot Scenarios} (Sec.~\ref{sec:offline_online_scenarios} and Sec.~\ref{sec:zsc_scenarios}): Examines environment support for evaluating offline-to-online transfer, generalization from fixed datasets, and coordination with novel partners.
\end{itemize}

Together, these components provide a foundation for scenario-driven evaluations that complement algorithmic innovation by making assumption boundaries empirically visible.

\subsection{Survey of MARL Benchmarks}
\label{sec:scenario_benchmarks}

We review a broad range of benchmarks commonly used to evaluate MARL. Through summarization and comparison, we highlight the configurability and diversity of environments within each category, while asking which controlled shifts they can support and which assumption boundaries they leave difficult to diagnose.

\paragraph{Structured Games.}
Structured games constitute the foundation of many MARL algorithmic developments, providing minimal yet expressive settings for studying inter-agent coordination, credit assignment, and policy generalization. Benchmarks such as MPE~\cite{maddpg}, SMAC~\cite{samvelyan2019starcraft}, GRF~\cite{kurach2020google}, and RWARE~\cite{RWAREandLBF} offer tractable environments with 2–27 agents, well-defined action-observation structures, and support for cooperative or mixed objective formulations. For example, SMAC enables heterogeneous unit control under cooperative goals with partial observability, while MPE introduces mixed-sum scenarios with optional communication and varying team sizes. More scalable environments such as MAgent~\cite{zheng2018magent}, Neural MMO~\cite{suarez2023neural}, and MAPF~\cite{stern2019multi} support populations exceeding 100 agents, albeit often without full observability or agent heterogeneity.

The diversity of structured games extends further across dimensions such as communication modality (e.g., Overcooked~\cite{carroll2019utility}, MACO~\cite{wang2022context, chen2026fivews}), asynchronous interactions (e.g., Matrix Games~\cite{claus1998dynamics}, MARL"O~\cite{perez2019multi}), and the degree of environment customizability. Environments like Hide-and-Seek~\cite{baker2019emergent} and Hallway~\cite{wang2020learning} provide sparse or emergent coordination structures, while Hanabi~\cite{hanabi} uniquely emphasizes belief modelling and implicit coordination. Collectively, structured games support controlled experimentation and comparative analysis, though they often abstract away complexities inherent to real-world deployments.

\paragraph{Application-Oriented Simulators.}
Application simulators advance beyond abstract coordination to incorporate real-world constraints, offering a higher-fidelity testbed for evaluating MARL algorithms under deployment-oriented assumptions. These include robotic manipulation domains (e.g., MAMuJoCo~\cite{facmac}, Bi-DexHands~\cite{chen2022towards}), smart city and mobility simulators (e.g., CityFlow~\cite{zhang2019cityflow}, SUMO~\cite{krajzewicz2002sumo}), and emerging space and sustainability applications (e.g., BSK-RL~\cite{stephenson2024bsk}, SustainDC~\cite{naug2024sustaindc}). These simulators typically support larger populations (e.g., MAPDN~\cite{wang2021multi}, MetaDrive~\cite{li2022metadrive}), asynchronous execution (e.g., SMARTS~\cite{SMARTS}, WFCRL~\cite{monroc2025wfcrl}), and diverse agent roles or morphologies (e.g., MARBLER~\cite{torbati2023marbler}, MaMo~\cite{xue2022multi}).

Many application simulators prioritize task realism and configurability. MetaDrive includes procedural generation of traffic and road topology, facilitating generalization to novel driving conditions. BSK-RL and Flatland~\cite{mohanty2020flatland} support mission composability and dynamic task scaling. Meanwhile, LAG~\cite{liu2022light}, MABIM~\cite{yang2023versatile}, and SMARTS offer fine-grained agent-environment interactions with mixed-motive tasks and temporal constraints. Although such environments introduce higher variance and computational demands, they are valuable for stress-testing policy robustness.

\paragraph{LLM-Based Multi-Agent Systems.}
Recent work has introduced language-enabled multi-agent environments that leverage large language models (LLMs) to support open-ended, natural language-driven interactions~\cite{yao2026langmarl}. These settings emphasize high-level reasoning, task modularity, and human-aligned communication protocols. Environments such as Welfare~\cite{mukobi2023welfare}, AgentVerse~\cite{chen2023agentverse}, and Collab-Overcooked~\cite{sun2025collab} focus on cooperation, negotiation, and task decomposition via structured prompts. Llmarena~\cite{chen2024llmarena} and BattleAgentBench~\cite{wang2024battleagentbench} extend this paradigm to adversarial and mixed-motive interactions, incorporating multi-round strategy evolution and language-based negotiation. These environments typically support 2--8 agents under partial observability and rely on prompt-based control interfaces, offering a unique lens into emergent coordination and role specialization. However, they currently lack standardized protocols and evaluation metrics, limiting their utility for systematic comparison.

\paragraph{Benchmark Characterization.}
To enable a systematic comparison, Table~\ref{tab:env_learning_adaptability} provides an overview of representative MARL benchmarks across seven key dimensions. Although Structured Games, Application Simulators, and LLM-based environments differ in their underlying assumptions, task abstractions, and complexity, they exhibit considerable overlap in core features. Across all categories, one can find environments that support large-scale agent populations (often exceeding 100 agents), partial observability, inter-agent communication, diverse reward structures, and asynchronous execution or agent heterogeneity. Notably, heterogeneity and task customizability are not confined to any single category but instead manifest differently depending on the simulation context and design emphasis.
Rather than implying a strict progression or superiority among these categories, this taxonomy highlights the importance of aligning benchmark selection with the specific goals of a study. Structured games provide controlled, interpretable environments well-suited for analysing coordination strategies and algorithmic components in isolation. Application-oriented simulators introduce realistic dynamics, environmental stochasticity, and mission-driven variability, thereby enabling evaluation under more deployment-relevant conditions. In contrast, LLM-based environments foreground natural language interfaces, emergent role specialization, and high-level reasoning, offering a unique lens into communication and generalization in language-mediated multi-agent systems. Collectively, these benchmarks form a complementary suite, and their utility should be assessed in terms of research intent: whether to isolate algorithmic contributions, test robustness in realistic domains, or explore language-grounded agent interaction.

\begin{sidewaystable}[p]
  \centering
  \caption{Environment diversity and configurability. Each benchmark is assessed across seven dimensions: population range, communication and observability structures, learning objectives, support for asynchronous execution, agent heterogeneity, task customizability, and the number of available tasks. Note: if an environment does not explicitly provide predefined scenarios but allows them to be generated within a continual learning setup, we assign a value of $1$ for its task count.}
  \vspace{-4pt}

  \label{tab:env_learning_adaptability}
  \begingroup
  \tiny
  \renewcommand{\arraystretch}{1.06}
  \setlength{\tabcolsep}{1pt}
  \begin{tabular}{llccccccc}
    \toprule
     & \textbf{Environment} 
     & \textbf{Pop. Scale} 
     & \textbf{Comm./Obs.} 
     & \textbf{Objective} 
     & \textbf{Async.} 
     & \textbf{Hetero.} 
     & \textbf{Customize.} 
     & \textbf{Tasks} \\
    \midrule
    \textbf{Structured Games} 
      & Matrix Game~\cite{claus1998dynamics}        & 2         & No / Full         & Mixed     & \cmark    & \cmark   & \cmark     & 1 \\
      & RWARE~\cite{RWAREandLBF}                    & 2--4      & No / Partial      & Coop      & \xmark    & \xmark   & \cmark     & 1 \\
      & MPE~\cite{maddpg}                           & 2--6      & Yes / Partial     & Mixed     & \cmark    & \cmark   & \cmark     & 6 \\
      & SMAC~\cite{samvelyan2019starcraft}          & 2--27     & No / Partial      & Coop      & \xmark    & \cmark   & \cmark     & 23 \\
      & GRF~\cite{kurach2020google}                 & 2--22     & No / Full         & Mixed     & \xmark    & \cmark   & \xmark     & 7 \\
      & MAgent~\cite{zheng2018magent}               & $>$100    & No / Partial      & Mixed     & \cmark    & \xmark   & \cmark     & 6 \\
      & GoBigger~\cite{zhang2023gobigger}           & $>$100    & No / Partial      & Mixed     & \xmark    & \xmark   & \cmark     & 1 \\
      & Overcooked~\cite{carroll2019utility}        & 2--4      & No / Full         & Coop      & \cmark    & \cmark   & \cmark     & 5 \\
      & Pommerman~\cite{resnick2018pommerman}       & 2--4      & No / Full         & Mixed     & \xmark    & \xmark   & \xmark     & 3 \\
      & SISL~\cite{gupta2017cooperative}            & 3--8      & No / Full         & Coop      & \xmark    & \xmark   & \xmark     & 3 \\
      & Hanabi~\cite{hanabi}                        & 2--5      & No / Partial      & Coop      & \cmark    & \xmark   & \xmark     & 4 \\
      & MACO~\cite{wang2022context}                 & 5--15     & Yes / Partial     & Mixed     & \xmark    & \xmark   & \cmark     & 6 \\
      & MARL\"O~\cite{perez2019multi}               & 2--8      & No / Partial      & Mixed     & \cmark    & \xmark   & \cmark     & 14 \\
      & Hallway~\cite{wang2020learning}             & 2         & Yes / Partial     & Coop      & \xmark    & \xmark   & \cmark     & 1 \\
      & Hide-and-Seek~\cite{baker2019emergent}      & 2--6      & No / Full         & Mixed     & \xmark    & \cmark   &  \xmark    & 1 \\
      & Gathering~\cite{leibo2017multi}             & 2         & No / Full         & Coop      & \xmark    & \xmark   & \cmark     & 1 \\
      & MAPF~\cite{stern2019multi}                  & $>$100    & No / Partial      & Coop      & \xmark    & \cmark   & \cmark     & 27 \\
      & DCA~\cite{fu2022concentration}              & 2--30     & No / Partial      & Mixed     & \xmark    & \xmark   & \cmark     & 1 \\
      & Neural MMO~\cite{suarez2023neural}          & $>$100    & No / Partial      & Mixed     & \cmark    & \xmark   & \cmark     & 1 \\
      & MEAL~\cite{tomilin2026meal}                 & 2         & No / Full         & Coop      & \xmark    & \cmark   & \cmark     & 100 \\
    \midrule
    \textbf{Application Simulators}
      & Bi-DexHands~\cite{chen2022towards}          & 2         & No / Full         & Coop      & \xmark    & \cmark  & \cmark      & 17 \\
      & MATE~\cite{pan2022mate}                     & 2--16     & Yes / Partial     & Mixed     & \xmark    & \cmark  & \cmark      & 5  \\
      & MAMuJoCo~\cite{facmac}                      & 2--6      & No / Full         & Coop      & \xmark    & \cmark  & \xmark      & 10 \\
      & SustainDC~\cite{naug2024sustaindc}          & 3         & No / Partial      & Coop      & \xmark    & \cmark  & \cmark      & 1 \\
      & MAPDN~\cite{wang2021multi}                  & $>$100    & No / Partial      & Coop      & \xmark    & \cmark  & \xmark      & 3 \\
      & LAG~\cite{liu2022light}                     & 2--8      & No / Partial      & Mixed     & \xmark    & \xmark  & \cmark      & 3 \\
      & BSK-RL~\cite{stephenson2024bsk}             & $>$100    & Yes / Partial     & Coop      & \xmark    & \xmark  & \cmark      & 2 \\
      & WFCRL~\cite{monroc2025wfcrl}                & 7--92     & No / Partial      & Mixed     & \cmark    & \xmark  & \xmark      & 2  \\
      & CityFlow~\cite{zhang2019cityflow}           & $>$100    & No / Partial      & Coop      & \cmark    & \xmark  & \xmark      & 1 \\
      & MetaDrive~\cite{li2022metadrive}            & 20--40    & No / Partial      & Mixed     & \cmark    & \xmark  & \cmark      & 7 \\
      & Flatland~\cite{mohanty2020flatland}         & $>$100    & No / Partial      & Coop      & \cmark    & \xmark  & \cmark      & 1 \\
      & SUMO~\cite{krajzewicz2002sumo}              & 2--6      & No / Full         & Mixed     & \xmark    & \xmark  & \cmark      & 1 \\
      & MARBLER~\cite{torbati2023marbler}           & 4--6      & Yes / Partial     & Mixed     & \xmark    & \cmark  & \cmark      & 5 \\
      & MaMo~\cite{xue2022multi}                    & 2--4      & No / Partial      & Coop      & \xmark    & \cmark  & \cmark      & 8 \\
      & MABIM~\cite{yang2023versatile}              & $>$100    & No / Partial      & Mixed     & \xmark    & \xmark  & \cmark      & 2 \\
      & SMARTS~\cite{SMARTS}                        & 3--5      & No / Partial      & Mixed     & \cmark    & \cmark  & \cmark      & 1 \\
    \midrule
    \textbf{LLM-based Benchmark}
      & Welfare~\cite{mukobi2023welfare}            & 2-7       & Yes/Full          & Coop      & \xmark    & \cmark  & \cmark      & 1 \\
      & Magic~\cite{xu2023magic}                    & 3         & Yes/Partial       & Coop      & \xmark    & \xmark  & \xmark      & 5 \\
      & Agentverse~\cite{chen2023agentverse}        & 2-3       & Yes/Partial       & Coop      & \cmark    & \cmark  & \cmark      & 3 \\
      & Avalonbench~\cite{light2023avalonbench}     & 5         & Yes/Partial       & Mixed     & \xmark    & \cmark  & \xmark      & 1 \\
      & Villageragent~\cite{dong2024villageragent}  & 2-8       & No/Partial        & Coop      & \xmark    & \cmark  & \cmark      & 3 \\
      & Llmarena~\cite{chen2024llmarena}            & 2-5       & Yes/Partial       & Mixed     & \cmark    & \cmark  & \cmark      & 7 \\
      & Battleagentbench~\cite{wang2024battleagentbench} & 1-6  & Yes/Partial       & Mixed     & \xmark    & \xmark  & \cmark      & 3 \\
      & PokerBench~\cite{zhuang2025pokerbench}      & 6         & No/Partial        & Comp      & \xmark    & \xmark  & \xmark      & 1 \\
      & Multiagentbench~\cite{zhu2025multiagentbench}  & 2-7    & Yes/Partial       & Mixed     & \xmark    & \cmark  & \cmark      & 6 \\
      & Collab-Overcooked~\cite{sun2025collab}      &  2        & No/Full           & Coop      & \xmark    & \cmark  & \cmark      & 6 \\
    \bottomrule
  \end{tabular}
  \endgroup
\end{sidewaystable}

\subsection{Continual and Curriculum Scenarios}
\label{sec:continual_curriculum}

Scenario-driven adaptability is not only about diversity in environment configurations, but also about how task transitions are constructed. Continual and curriculum settings are useful because they define ordered source-to-target gaps: the evaluator can specify what changes between tasks, what is preserved, and whether progress on earlier tasks helps or harms later performance~\cite{long2020evolutionary, wang2023towards, chen2021variational, chen2024accelerate, zhang2022automatic, tomilin2026meal}. These gaps, whether arising from increasing population size, added agent heterogeneity, or modified reward semantics, provide a structured way to test \textit{reuse}, \textit{retention}, and \textit{extension}.

Specifically, \textit{reuse} refers to the agent's capacity to apply learned behaviours to similar but incrementally more complex settings; \textit{retention} concerns the stability of earlier skills when new learning occurs; and \textit{extension} captures the ability to build on previous knowledge to solve qualitatively novel tasks. The structure and semantics of task gaps determine whether these dimensions of adaptability can be meaningfully assessed. As such, continual and curriculum settings provide a principled framework for designing multi-agent scenarios that gradually introduce complexity while preserving meaningful continuity in learning signals and task structure.

\paragraph{Task Gaps and Principles for Transfer-Compatible Design.}
Designing transfer-compatible task gaps requires more than simply increasing difficulty. Effective curricula must preserve semantic alignment across tasks to ensure that policy improvements reflect generalization rather than task-specific tuning. Several principles have emerged as critical for structuring meaningful transitions in multi-agent settings:

\begin{enumerate}
\item \textit{Incremental population scaling:} Gradually increasing the number of agents (e.g., 3 $\rightarrow$ 5 $\rightarrow$ 8) supports the reuse of coordination strategies and exposes scalability constraints.
\item \textit{Progressive role diversification:} Introducing new agent types or abilities in modular stages (e.g., Zealot $\rightarrow$ Zealot+Stalker $\rightarrow$ Zealot+Stalker+Colossus~\cite{samvelyan2019starcraft}) enables compositional policy learning and generalization to heterogeneous teams.
\item \textit{Reward structure consistency:} Maintaining coherent reward objectives across tasks (e.g., always cooperative) reduces the need for strategy re-invention and promotes cumulative skill development.
\end{enumerate}

Despite the promise of curriculum-based training, it remains underexplored in MARL. Most benchmarks lack support for structured task graphs~\cite{claus1998dynamics, RWAREandLBF, maddpg}, scaffolded skill progression~\cite{facmac, wang2021multi, stephenson2024bsk}, or staged evaluation protocols~\cite{xue2022multi, yang2023versatile, SMARTS} that enable quantitative assessment of \emph{forward transfer} (performance gains on future tasks due to earlier learning) or \emph{backward transfer} (retention of earlier capabilities). These metrics are essential for characterizing the trajectory of learning across task gaps.

\paragraph{Representational Continuity Across Tasks.}
In addition to behavioural alignment, effective curricula must ensure representational consistency across tasks to facilitate transfer. Several constraints support this goal:

\begin{enumerate}
\item \textit{Feature alignment:} Observation and action spaces should maintain consistent semantics. For example, stable slots for agent-local, opponent-specific, and global inputs to support shared encoders~\cite{hanabi, facmac, maddpg}.
\item \textit{Predictable dimensional scaling:} As task complexity increases (e.g., more agents or abilities), changes in the input/output space should follow structured patterns to avoid frequent architectural redesign~\cite{samvelyan2019starcraft, li2022metadrive, SMARTS}.
\item \textit{Scenario modularity:} Tasks should include reusable behavioural components (e.g., navigation, foraging, cooperation) to encourage the emergence of transferable sub-skills~\cite{RWAREandLBF, pan2022mate, liu2022light}.
\end{enumerate}

By adhering to these principles, scenario designers can structure task gaps that meaningfully evaluate reuse, retention, and extension. This not only supports rigorous benchmarking of continual MARL methods, but also clarifies which assumption shift caused success or failure: population growth, role diversification, reward modification, interface change, or loss of previously acquired skills.

\subsection{Offline Pretraining and Online Transfer Scenarios}
\label{sec:offline_online_scenarios}

While curriculum learning emphasizes structured environment interaction, many practical MARL deployments operate under strict limitations on real-time data collection~\cite{monroc2025wfcrl, stephenson2024bsk, wang2021multi}. In such cases, agents must be trained from static datasets, often with no access to online rollouts. Offline~\cite{kostrikov2021offline, kumar2020conservative, chen2021decision, li2026logo} and offline-to-online~\cite{song2022hybrid, xie2021policy, ball2023efficient} scenarios are therefore diagnostic because they explicitly shift the data-collection assumption: from fixed historical trajectories to either zero-shot deployment or continued online interaction.
These settings lie at the intersection of learning adaptability and policy adaptability. The learning process must cope with fixed and potentially biased datasets, while the resulting policies should be evaluated under a stated adaptation-cost level, such as zero-shot transfer or online fine-tuning.

\paragraph{Offline Scenario.}
Offline MARL entails learning policies entirely from pre-collected trajectories, with no further environment interaction during training. This paradigm is particularly relevant for domains where online sampling is impractical due to cost, safety, or logistical constraints. In the absence of interactive feedback, generalization depends heavily on dataset quality and diversity. Desirable properties include:

\begin{enumerate}
\item Coverage of diverse coordination behaviours, agent roles, and reward signals;
\item Inclusion of supervision levels ranging from expert to exploratory or random policies;
\item Scenario heterogeneity across different environment configurations and population structures.
\end{enumerate}

The OG-MARL benchmark suite~\cite{formanek2023ogmarl} exemplifies this setting, providing stratified datasets across canonical environments such as SMAC~\cite{samvelyan2019starcraft}, SMACv2~\cite{ellis2023smacv2}, MAMuJoCo~\cite{facmac}, Flatland~\cite{mohanty2020flatland}, RWARE~\cite{RWAREandLBF}, and MPE~\cite{maddpg}.
In the meantime, many existing datasets are collected using hand-designed or static policies and lack mechanisms to promote behavioural diversity or balanced task coverage~\cite{wang2023offline, pan2022plan, shao2023counterfactual}. This can lead to under-representation of critical coordination scenarios and hinder robust evaluation.

\paragraph{Offline-to-Online Transfer.}
The offline-to-online setting bridges the gap between static offline training and fully interactive learning~\cite{song2022hybrid, xie2021policy, ball2023efficient}. It evaluates whether policies pretrained on fixed datasets—or supplemented with additional offline data—can accelerate learning in the online version of the same or similar tasks. This setting centres on continued policy optimization using a combination of offline and online experience.

In contrast to curriculum learning, where task complexity typically increases in a structured progression (e.g., 3 $\rightarrow$ 5 $\rightarrow$ 8 agents), offline-to-online transfer often involves transitions across tasks that are unordered or non-monotonic in complexity. For example, an agent trained offline on SMAC tasks such as \texttt{3m} and \texttt{8m} may be evaluated online on \texttt{5m}, requiring compositional generalization rather than stepwise skill accumulation. This makes offline-to-online transfer a more flexible—yet also more challenging—test of policy adaptability.
From the perspective of task gaps, offline tasks should be designed to complement each other in ways that support generalization. Specifically, knowledge acquired from offline behavioural datasets A and B should be composable or synergistic, enabling more effective adaptation when fine-tuning on a distinct but related online task C.
To design a good offline-to-online MARL setting, the following principles are essential:

\begin{enumerate}
\item \textit{Complementary offline tasks:} Offline datasets should span distinct but related tasks to support compositional generalization during online adaptation.
\item \textit{Representational consistency:} Observation and action spaces must be aligned across offline and online tasks to enable direct policy reuse without architectural changes.
\item \textit{Behavioural diversity:} Offline data should include a mix of expert and suboptimal behaviours to encourage robust initialization and flexible fine-tuning.
\end{enumerate}

Overall, Offline MARL highlights a unique angle in the adaptability landscape: its success hinges on whether agents can generalize from fixed data distributions, and on whether the benchmark separates dataset coverage from policy transfer. When equipped with sufficiently rich offline data, a flexible learning algorithm may implicitly absorb a wide spectrum of interaction dynamics and task structures. However, scenario-driven evaluation should still report the source dataset, target scenario, permitted online interaction, and failure modes, because offline-to-online transfer remains underexplored in multi-agent settings compared with its single-agent counterpart~\cite{song2022hybrid, xie2021policy, ball2023efficient}.

\subsection{Zero-Shot Coordination Scenarios}
\label{sec:zsc_scenarios}

Zero-shot coordination (ZSC) scenarios assess an agent’s capacity to collaborate with unfamiliar partners at test time without prior joint training, shared parameters, or explicit coordination protocols. In the scenario-driven taxonomy, ZSC isolates a partner-policy or convention shift: the task may remain fixed, but the co-player distribution changes. Rather than optimizing for task performance alone, these scenarios foreground \emph{social adaptability}: the agent's ability to interpret novel conventions, align with diverse behaviours, and act coherently in uncertain multi-agent settings.

In practical deployments, such as modular robotics, autonomous driving, or simulation-based multi-agent platforms, agents frequently encounter partners developed independently under different assumptions, inductive biases, or design paradigms. This variability challenges agents to operate not within a fixed policy ecosystem, but across a dynamic \emph{social landscape} composed of heterogeneous conventions and strategies.

Empirical studies show that independently trained agents often converge on idiosyncratic or brittle conventions that fail under cross-play~\cite{hu2020other, lupu2021trajectory}. Recent approaches mitigate this by explicitly injecting \emph{partner diversity} during training, encouraging robustness to novel coordination styles via latent belief modelling~\cite{lucas2022any, li2023cooperative, powell2026zscshaping}. These findings motivate the need for benchmarks that expose agents to a wide range of conventions and promote generalization across unseen social configurations.

Effective ZSC benchmarks should therefore be designed with three core criteria:

\begin{enumerate}
\item \textit{Multiple viable conventions:} Tasks must permit a spectrum of valid coordination equilibria, avoiding overfitting to a single dominant solution.
\item \textit{Held-out partner diversity:} Evaluation should involve interaction with previously unseen agents exhibiting distinct learning trajectories, inductive priors, or reward shaping.
\item \textit{Implicit interaction mechanisms:} Partial observability and restricted communication ensure that agents must infer partner intent from behaviour, rather than rely on explicit signalling.
\end{enumerate}

Prominent examples include Hanabi~\cite{bard2020hanabi}, which enforces theory-of-mind reasoning under strict communication limits, and Overcooked-AI~\cite{carroll2019utility}, where agents must disambiguate spatial-temporal plans in environments with multiple emergent conventions. These domains illustrate how structured randomness in the social landscape can reveal coordination failures or successes under zero-shot constraints.

\section{Future Directions}
\label{sec:future}

We conclude by outlining open challenges and research opportunities for improving adaptability in MARL. To reflect the structure of this survey, we organize this discussion along three axes corresponding to learning adaptability, policy adaptability, and scenario-driven adaptability.

For learning adaptability, a key challenge lies in balancing scalability with coordination fidelity. Current methods such as value decomposition and centralized critics offer effective coordination but scale poorly to large or asynchronous systems due to communication overhead and growing input dimensionality. Future work may explore partially coupled architectures that enable localized coordination within agent subsets, while maintaining global coherence through compositional critics, graph-based learning, or dynamic communication structures. Additionally, real-world systems often involve transitions between cooperative, competitive, or mixed-motive dynamics. This calls for adaptable objective representations, reward disentanglement, and regularization techniques that support generalization across task interaction types. Supporting distributed and asynchronous execution remains another open problem. Hybrid approaches that distil centralized critics into decentralized policies, or apply minimal synchronization protocols in networked settings, present promising directions.

For policy adaptability, a central challenge is to design policies that generalize across tasks, roles, and teammates under clearly stated adaptation costs. Transformer-based models and permutation-invariant networks provide useful inductive biases but often lack mechanisms for semantic modularity or task-aware decomposition. Future designs should incorporate subtask abstractions, interpretable attention patterns, and modular heads aligned with agent roles or latent strategies. Offline pretraining can further support policy transfer, yet current approaches are constrained by dataset diversity and limited reuse. Developing shared policy libraries through distillation, policy interpolation, or embedding-based indexing may help enable fast adaptation and in-context learning. Lifelong learning also remains underexplored. More flexible adaptation strategies such as online embedding updates, latent-conditioned policy generation, and regret-aware updates could improve robustness across evolving task sequences. Zero-shot coordination with unfamiliar partners introduces additional challenges, requiring mechanisms for belief modelling, convention inference, and partner profiling to align behaviour under uncertainty.

Advancing these capabilities depends on structured and diagnostic evaluation environments. Many existing benchmarks consist of static tasks with limited configurability, offering coarse-grained assessments of generalization. To enable fine-grained analysis, future environments should support parametrized variations in agent count, role heterogeneity, reward structure, and communication topology, while reporting which factors are held fixed. Such continuous scenario spaces would facilitate the study of curriculum learning, continual adaptation, and transfer across structured complexity gradients. There is also a pressing need for standardized offline MARL datasets that span diverse tasks and agent types, together with explicit offline-to-online protocols that report source data, target scenarios, permitted interaction, and adaptation cost. To evaluate social adaptability, benchmarks should include multiple viable conventions, diverse held-out partners, and limited observability or communication. Evaluation protocols should prioritize cross-play with independently trained agents using different architectures or learning objectives. Finally, diagnostic toolkits such as transferability matrices, skill composition graphs, degradation curves, and failure mode taxonomies are essential for understanding where and how current algorithms succeed or fail.

To address these gaps, we advocate for future research on more assumption-aware learning paradigms, more explicit policy-transfer mechanisms, and more structured testing environments. By aligning algorithmic development with diagnostic benchmarks, MARL research can provide clearer evidence about where methods are applicable before deployment in dynamic real-world settings.

\section{Conclusion}
\label{sec:conclusion}

This survey has presented a structured analysis of adaptability in multi-agent reinforcement learning, organized along three core axes: learning adaptability, policy adaptability, and scenario-driven adaptability. We examined how existing methods address population scaling, task variation, and execution constraints during training; how policies generalize across tasks, roles, and partners under different adaptation-cost levels; and how scenario design can expose controlled shifts for diagnostic evaluation.
Despite recent progress, our review highlights significant gaps in both algorithmic approaches and benchmark support. Key challenges include specifying which training assumptions a paradigm requires, measuring policy reuse without conflating zero-shot transfer with fine-tuning or retraining, and designing source-target protocols that isolate continual learning, offline-to-online transfer, and zero-shot coordination shifts.
We encourage future work to develop scalable training paradigms, modular and transferable policy architectures, and structured evaluation environments that report the assumptions being shifted, the adaptation mechanisms allowed, and the failure modes observed. Addressing these challenges can move MARL research beyond isolated benchmark performance toward clearer evidence of applicability under changing multi-agent conditions.

\bibliography{references}


\begin{thebibliography}{181}
\ifx \bisbn   \undefined \def \bisbn  #1{ISBN #1}\fi
\ifx \binits  \undefined \def \binits#1{#1}\fi
\ifx \bauthor  \undefined \def \bauthor#1{#1}\fi
\ifx \batitle  \undefined \def \batitle#1{#1}\fi
\ifx \bjtitle  \undefined \def \bjtitle#1{#1}\fi
\ifx \bvolume  \undefined \def \bvolume#1{\textbf{#1}}\fi
\ifx \byear  \undefined \def \byear#1{#1}\fi
\ifx \bissue  \undefined \def \bissue#1{#1}\fi
\ifx \bfpage  \undefined \def \bfpage#1{#1}\fi
\ifx \blpage  \undefined \def \blpage #1{#1}\fi
\ifx \burl  \undefined \def \burl#1{\textsf{#1}}\fi
\ifx \doiurl  \undefined \def \doiurl#1{\url{https://doi.org/#1}}\fi
\ifx \betal  \undefined \def \betal{\textit{et al.}}\fi
\ifx \binstitute  \undefined \def \binstitute#1{#1}\fi
\ifx \binstitutionaled  \undefined \def \binstitutionaled#1{#1}\fi
\ifx \bctitle  \undefined \def \bctitle#1{#1}\fi
\ifx \beditor  \undefined \def \beditor#1{#1}\fi
\ifx \bpublisher  \undefined \def \bpublisher#1{#1}\fi
\ifx \bbtitle  \undefined \def \bbtitle#1{#1}\fi
\ifx \bedition  \undefined \def \bedition#1{#1}\fi
\ifx \bseriesno  \undefined \def \bseriesno#1{#1}\fi
\ifx \blocation  \undefined \def \blocation#1{#1}\fi
\ifx \bsertitle  \undefined \def \bsertitle#1{#1}\fi
\ifx \bsnm \undefined \def \bsnm#1{#1}\fi
\ifx \bsuffix \undefined \def \bsuffix#1{#1}\fi
\ifx \bparticle \undefined \def \bparticle#1{#1}\fi
\ifx \barticle \undefined \def \barticle#1{#1}\fi
\bibcommenthead
\ifx \bconfdate \undefined \def \bconfdate #1{#1}\fi
\ifx \botherref \undefined \def \botherref #1{#1}\fi
\ifx \url \undefined \def \url#1{\textsf{#1}}\fi
\ifx \bchapter \undefined \def \bchapter#1{#1}\fi
\ifx \bbook \undefined \def \bbook#1{#1}\fi
\ifx \bcomment \undefined \def \bcomment#1{#1}\fi
\ifx \oauthor \undefined \def \oauthor#1{#1}\fi
\ifx \citeauthoryear \undefined \def \citeauthoryear#1{#1}\fi
\ifx \endbibitem  \undefined \def \endbibitem {}\fi
\ifx \bconflocation  \undefined \def \bconflocation#1{#1}\fi
\ifx \arxivurl  \undefined \def \arxivurl#1{\textsf{#1}}\fi
\csname PreBibitemsHook\endcsname

%%% 1
\bibitem[\protect\citeauthoryear{Arulkumaran
  et~al.}{2017}]{arulkumaran2017deep}
\begin{barticle}
\bauthor{\bsnm{Arulkumaran}, \binits{K.}},
\bauthor{\bsnm{Deisenroth}, \binits{M.P.}},
\bauthor{\bsnm{Brundage}, \binits{M.}},
\bauthor{\bsnm{Bharath}, \binits{A.A.}}:
\batitle{Deep reinforcement learning: A brief survey}.
\bjtitle{IEEE Signal Processing Magazine}
\bvolume{34}(\bissue{6}),
\bfpage{26}--\blpage{38}
(\byear{2017})
\end{barticle}
\endbibitem

%%% 2
\bibitem[\protect\citeauthoryear{Anand et~al.}{2026}]{anand2026graphon}
\begin{barticle}
\bauthor{\bsnm{Anand}, \binits{E.}},
\bauthor{\bsnm{Hoffmann}, \binits{R.}},
\bauthor{\bsnm{Liaw}, \binits{S.}},
\bauthor{\bsnm{Wierman}, \binits{A.}}:
\batitle{Graphon mean-field subsampling for cooperative heterogeneous
  multi-agent reinforcement learning}.
\bjtitle{arXiv preprint arXiv:2602.16196}
(\byear{2026})
\doiurl{10.48550/arXiv.2602.16196}
\end{barticle}
\endbibitem

%%% 3
\bibitem[\protect\citeauthoryear{Aoki}{1965}]{aoki1965optimal}
\begin{barticle}
\bauthor{\bsnm{Aoki}, \binits{M.}}:
\batitle{Optimal control of partially observable markovian systems}.
\bjtitle{Journal of The Franklin Institute}
\bvolume{280}(\bissue{5}),
\bfpage{367}--\blpage{386}
(\byear{1965})
\end{barticle}
\endbibitem

%%% 4
\bibitem[\protect\citeauthoryear{Akula et~al.}{2026}]{akula2026asalt}
\begin{barticle}
\bauthor{\bsnm{Akula}, \binits{A.}},
\bauthor{\bsnm{Perepu}, \binits{S.K.}},
\bauthor{\bsnm{Sarkar}, \binits{A.}},
\bauthor{\bsnm{Dey}, \binits{K.}}:
\batitle{{ASALT}: Adaptive state alignment for lateral transfer in multi-agent
  reinforcement learning}.
\bjtitle{arXiv preprint arXiv:2606.24601}
(\byear{2026})
\doiurl{10.48550/arXiv.2606.24601}
\end{barticle}
\endbibitem

%%% 5
\bibitem[\protect\citeauthoryear{Barab{\'a}si et~al.}{1999}]{barabasi1999mean}
\begin{barticle}
\bauthor{\bsnm{Barab{\'a}si}, \binits{A.-L.}},
\bauthor{\bsnm{Albert}, \binits{R.}},
\bauthor{\bsnm{Jeong}, \binits{H.}}:
\batitle{Mean-field theory for scale-free random networks}.
\bjtitle{Physica A: Statistical Mechanics and its Applications}
\bvolume{272}(\bissue{1-2}),
\bfpage{173}--\blpage{187}
(\byear{1999})
\end{barticle}
\endbibitem

%%% 6
\bibitem[\protect\citeauthoryear{Busoniu
  et~al.}{2008}]{busoniu2008comprehensive}
\begin{barticle}
\bauthor{\bsnm{Busoniu}, \binits{L.}},
\bauthor{\bsnm{Babuska}, \binits{R.}},
\bauthor{\bsnm{De~Schutter}, \binits{B.}}:
\batitle{A comprehensive survey of multiagent reinforcement learning}.
\bjtitle{IEEE Transactions on Systems, Man, and Cybernetics, Part C
  (Applications and Reviews)}
\bvolume{38}(\bissue{2}),
\bfpage{156}--\blpage{172}
(\byear{2008})
\end{barticle}
\endbibitem

%%% 7
\bibitem[\protect\citeauthoryear{Brambilla et~al.}{2013}]{brambilla2013swarm}
\begin{barticle}
\bauthor{\bsnm{Brambilla}, \binits{M.}},
\bauthor{\bsnm{Ferrante}, \binits{E.}},
\bauthor{\bsnm{Birattari}, \binits{M.}},
\bauthor{\bsnm{Dorigo}, \binits{M.}}:
\batitle{Swarm robotics: a review from the swarm engineering perspective}.
\bjtitle{Swarm Intelligence}
\bvolume{7},
\bfpage{1}--\blpage{41}
(\byear{2013})
\end{barticle}
\endbibitem

%%% 8
\bibitem[\protect\citeauthoryear{Bard et~al.}{2020a}]{hanabi}
\begin{barticle}
\bauthor{\bsnm{Bard}, \binits{N.}},
\bauthor{\bsnm{Foerster}, \binits{J.N.}},
\bauthor{\bsnm{Chandar}, \binits{S.}},
\bauthor{\bsnm{Burch}, \binits{N.}},
\bauthor{\bsnm{Lanctot}, \binits{M.}},
\bauthor{\bsnm{Song}, \binits{H.F.}},
\bauthor{\bsnm{Parisotto}, \binits{E.}},
\bauthor{\bsnm{Dumoulin}, \binits{V.}},
\bauthor{\bsnm{Moitra}, \binits{S.}},
\bauthor{\bsnm{Hughes}, \binits{E.}}, \betal:
\batitle{The hanabi challenge: A new frontier for ai research}.
\bjtitle{Artificial Intelligence}
\bvolume{280},
\bfpage{103216}
(\byear{2020})
\end{barticle}
\endbibitem

%%% 9
\bibitem[\protect\citeauthoryear{Bard et~al.}{2020b}]{bard2020hanabi}
\begin{barticle}
\bauthor{\bsnm{Bard}, \binits{N.}},
\bauthor{\bsnm{Foerster}, \binits{J.N.}},
\bauthor{\bsnm{Chandar}, \binits{S.}},
\bauthor{\bsnm{Burch}, \binits{N.}},
\bauthor{\bsnm{Lanctot}, \binits{M.}},
\bauthor{\bsnm{Song}, \binits{H.F.}},
\bauthor{\bsnm{Parisotto}, \binits{E.}},
\bauthor{\bsnm{Dumoulin}, \binits{V.}},
\bauthor{\bsnm{Moitra}, \binits{S.}},
\bauthor{\bsnm{Hughes}, \binits{E.}}, \betal:
\batitle{The hanabi challenge: A new frontier for ai research}.
\bjtitle{Artificial Intelligence}
\bvolume{280},
\bfpage{103216}
(\byear{2020})
\end{barticle}
\endbibitem

%%% 10
\bibitem[\protect\citeauthoryear{Bensoussan et~al.}{2013}]{bensoussan2013mean}
\begin{bbook}
\bauthor{\bsnm{Bensoussan}, \binits{A.}},
\bauthor{\bsnm{Frehse}, \binits{J.}},
\bauthor{\bsnm{Yam}, \binits{P.}}, \betal:
\bbtitle{Mean Field Games and Mean Field Type Control Theory}
vol. \bseriesno{101}.
\bpublisher{Springer}, \blocation{???}
(\byear{2013})
\end{bbook}
\endbibitem

%%% 11
\bibitem[\protect\citeauthoryear{Bai et~al.}{2021}]{bai2021value}
\begin{botherref}
\oauthor{\bsnm{Bai}, \binits{Y.}},
\oauthor{\bsnm{Gong}, \binits{C.}},
\oauthor{\bsnm{Zhang}, \binits{B.}},
\oauthor{\bsnm{Fan}, \binits{G.}},
\oauthor{\bsnm{Hou}, \binits{X.}}:
Value function factorisation with hypergraph convolution for cooperative
  multi-agent reinforcement learning.
arXiv:2112.06771
(2021)
\end{botherref}
\endbibitem

%%% 12
\bibitem[\protect\citeauthoryear{Baker et~al.}{2019}]{baker2019emergent}
\begin{bchapter}
\bauthor{\bsnm{Baker}, \binits{B.}},
\bauthor{\bsnm{Kanitscheider}, \binits{I.}},
\bauthor{\bsnm{Markov}, \binits{T.}},
\bauthor{\bsnm{Wu}, \binits{Y.}},
\bauthor{\bsnm{Powell}, \binits{G.}},
\bauthor{\bsnm{McGrew}, \binits{B.}},
\bauthor{\bsnm{Mordatch}, \binits{I.}}:
\bctitle{Emergent tool use from multi-agent autocurricula}.
In: \bbtitle{International Conference on Learning Representations}
(\byear{2019})
\end{bchapter}
\endbibitem

%%% 13
\bibitem[\protect\citeauthoryear{Ball et~al.}{2023}]{ball2023efficient}
\begin{bchapter}
\bauthor{\bsnm{Ball}, \binits{P.J.}},
\bauthor{\bsnm{Smith}, \binits{L.}},
\bauthor{\bsnm{Kostrikov}, \binits{I.}},
\bauthor{\bsnm{Levine}, \binits{S.}}:
\bctitle{Efficient online reinforcement learning with offline data}.
In: \bbtitle{International Conference on Machine Learning},
pp. \bfpage{1577}--\blpage{1594}
(\byear{2023}).
\bcomment{PMLR}
\end{bchapter}
\endbibitem

%%% 14
\bibitem[\protect\citeauthoryear{Bloembergen
  et~al.}{2015}]{bloembergen2015evolutionary}
\begin{barticle}
\bauthor{\bsnm{Bloembergen}, \binits{D.}},
\bauthor{\bsnm{Tuyls}, \binits{K.}},
\bauthor{\bsnm{Hennes}, \binits{D.}},
\bauthor{\bsnm{Kaisers}, \binits{M.}}:
\batitle{Evolutionary dynamics of multi-agent learning: A survey}.
\bjtitle{Journal of Artificial Intelligence Research}
\bvolume{53},
\bfpage{659}--\blpage{697}
(\byear{2015})
\end{barticle}
\endbibitem

%%% 15
\bibitem[\protect\citeauthoryear{Claus and Boutilier}{1998}]{claus1998dynamics}
\begin{barticle}
\bauthor{\bsnm{Claus}, \binits{C.}},
\bauthor{\bsnm{Boutilier}, \binits{C.}}:
\batitle{The dynamics of reinforcement learning in cooperative multiagent
  systems}.
\bjtitle{AAAI/IAAI}
\bvolume{1998}(\bissue{746-752}),
\bfpage{2}
(\byear{1998})
\end{barticle}
\endbibitem

%%% 16
\bibitem[\protect\citeauthoryear{Chen et~al.}{2024}]{chen2024llmarena}
\begin{botherref}
\oauthor{\bsnm{Chen}, \binits{J.}},
\oauthor{\bsnm{Hu}, \binits{X.}},
\oauthor{\bsnm{Liu}, \binits{S.}},
\oauthor{\bsnm{Huang}, \binits{S.}},
\oauthor{\bsnm{Tu}, \binits{W.-W.}},
\oauthor{\bsnm{He}, \binits{Z.}},
\oauthor{\bsnm{Wen}, \binits{L.}}:
Llmarena: Assessing capabilities of large language models in dynamic
  multi-agent environments.
arXiv preprint arXiv:2402.16499
(2024)
\end{botherref}
\endbibitem

%%% 17
\bibitem[\protect\citeauthoryear{Chen et~al.}{2021}]{chen2021decision}
\begin{barticle}
\bauthor{\bsnm{Chen}, \binits{L.}},
\bauthor{\bsnm{Lu}, \binits{K.}},
\bauthor{\bsnm{Rajeswaran}, \binits{A.}},
\bauthor{\bsnm{Lee}, \binits{K.}},
\bauthor{\bsnm{Grover}, \binits{A.}},
\bauthor{\bsnm{Laskin}, \binits{M.}},
\bauthor{\bsnm{Abbeel}, \binits{P.}},
\bauthor{\bsnm{Srinivas}, \binits{A.}},
\bauthor{\bsnm{Mordatch}, \binits{I.}}:
\batitle{Decision transformer: Reinforcement learning via sequence modeling}.
\bjtitle{Advances in neural information processing systems}
\bvolume{34},
\bfpage{15084}--\blpage{15097}
(\byear{2021})
\end{barticle}
\endbibitem

%%% 18
\bibitem[\protect\citeauthoryear{Carmona et~al.}{2019}]{carmona2019model}
\begin{botherref}
\oauthor{\bsnm{Carmona}, \binits{R.}},
\oauthor{\bsnm{Lauri{\`e}re}, \binits{M.}},
\oauthor{\bsnm{Tan}, \binits{Z.}}:
Model-free mean-field reinforcement learning: mean-field {MDP} and mean-field
  {Q-learning}.
arXiv:1910.12802
(2019)
\end{botherref}
\endbibitem

%%% 19
\bibitem[\protect\citeauthoryear{Carmona et~al.}{2023}]{carmona2023model}
\begin{barticle}
\bauthor{\bsnm{Carmona}, \binits{R.}},
\bauthor{\bsnm{Lauri{\`e}re}, \binits{M.}},
\bauthor{\bsnm{Tan}, \binits{Z.}}:
\batitle{Model-free mean-field reinforcement learning: mean-field mdp and
  mean-field q-learning}.
\bjtitle{The Annals of Applied Probability}
\bvolume{33}(\bissue{6B}),
\bfpage{5334}--\blpage{5381}
(\byear{2023})
\end{barticle}
\endbibitem

%%% 20
\bibitem[\protect\citeauthoryear{Carroll et~al.}{2019}]{carroll2019utility}
\begin{botherref}
\oauthor{\bsnm{Carroll}, \binits{M.}},
\oauthor{\bsnm{Shah}, \binits{R.}},
\oauthor{\bsnm{Ho}, \binits{M.K.}},
\oauthor{\bsnm{Griffiths}, \binits{T.}},
\oauthor{\bsnm{Seshia}, \binits{S.}},
\oauthor{\bsnm{Abbeel}, \binits{P.}},
\oauthor{\bsnm{Dragan}, \binits{A.}}:
On the utility of learning about humans for human-ai coordination.
Advances in neural information processing systems
\textbf{32}
(2019)
\end{botherref}
\endbibitem

%%% 21
\bibitem[\protect\citeauthoryear{Chen et~al.}{2023}]{chen2023agentverse}
\begin{botherref}
\oauthor{\bsnm{Chen}, \binits{W.}},
\oauthor{\bsnm{Su}, \binits{Y.}},
\oauthor{\bsnm{Zuo}, \binits{J.}},
\oauthor{\bsnm{Yang}, \binits{C.}},
\oauthor{\bsnm{Yuan}, \binits{C.}},
\oauthor{\bsnm{Qian}, \binits{C.}},
\oauthor{\bsnm{Chan}, \binits{C.-M.}},
\oauthor{\bsnm{Qin}, \binits{Y.}},
\oauthor{\bsnm{Lu}, \binits{Y.}},
\oauthor{\bsnm{Xie}, \binits{R.}}, et al.:
Agentverse: Facilitating multi-agent collaboration and exploring emergent
  behaviors in agents.
arXiv preprint arXiv:2308.10848
(2023)
\end{botherref}
\endbibitem

%%% 22
\bibitem[\protect\citeauthoryear{Cui et~al.}{2022}]{cui2022survey}
\begin{botherref}
\oauthor{\bsnm{Cui}, \binits{K.}},
\oauthor{\bsnm{Tahir}, \binits{A.}},
\oauthor{\bsnm{Ekinci}, \binits{G.}},
\oauthor{\bsnm{Elshamanhory}, \binits{A.}},
\oauthor{\bsnm{Eich}, \binits{Y.}},
\oauthor{\bsnm{Li}, \binits{M.}},
\oauthor{\bsnm{Koeppl}, \binits{H.}}:
A survey on large-population systems and scalable multi-agent reinforcement
  learning.
arXiv preprint arXiv:2209.03859
(2022)
\end{botherref}
\endbibitem

%%% 23
\bibitem[\protect\citeauthoryear{Chen et~al.}{2024}]{chen2024accelerate}
\begin{bchapter}
\bauthor{\bsnm{Chen}, \binits{J.}},
\bauthor{\bsnm{Xu}, \binits{Z.}},
\bauthor{\bsnm{Li}, \binits{Y.}},
\bauthor{\bsnm{Yu}, \binits{C.}},
\bauthor{\bsnm{Song}, \binits{J.}},
\bauthor{\bsnm{Yang}, \binits{H.}},
\bauthor{\bsnm{Fang}, \binits{F.}},
\bauthor{\bsnm{Wang}, \binits{Y.}},
\bauthor{\bsnm{Wu}, \binits{Y.}}:
\bctitle{Accelerate multi-agent reinforcement learning in zero-sum games with
  subgame curriculum learning}.
In: \bbtitle{Proceedings of the AAAI Conference on Artificial Intelligence},
vol. \bseriesno{38},
pp. \bfpage{11320}--\blpage{11328}
(\byear{2024})
\end{bchapter}
\endbibitem

%%% 24
\bibitem[\protect\citeauthoryear{Chen et~al.}{2026}]{chen2026fivews}
\begin{barticle}
\bauthor{\bsnm{Chen}, \binits{J.}},
\bauthor{\bsnm{Yang}, \binits{H.}},
\bauthor{\bsnm{Liu}, \binits{Z.}},
\bauthor{\bsnm{Joe-Wong}, \binits{C.}}:
\batitle{The five {W}s of multi-agent communication: Who talks to whom, when,
  what, and why -- a survey from {MARL} to emergent language and {LLM}s}.
\bjtitle{arXiv preprint arXiv:2602.11583}
(\byear{2026})
\doiurl{10.48550/arXiv.2602.11583}
\end{barticle}
\endbibitem

%%% 25
\bibitem[\protect\citeauthoryear{Chen et~al.}{2022}]{chen2022towards}
\begin{bchapter}
\bauthor{\bsnm{Chen}, \binits{Y.}},
\bauthor{\bsnm{Yang}, \binits{Y.}},
\bauthor{\bsnm{Wu}, \binits{T.}},
\bauthor{\bsnm{Wang}, \binits{S.}},
\bauthor{\bsnm{Feng}, \binits{X.}},
\bauthor{\bsnm{Jiang}, \binits{J.}},
\bauthor{\bsnm{Lu}, \binits{Z.}},
\bauthor{\bsnm{McAleer}, \binits{S.M.}},
\bauthor{\bsnm{Dong}, \binits{H.}},
\bauthor{\bsnm{Zhu}, \binits{S.-C.}}:
\bctitle{Towards human-level bimanual dexterous manipulation with reinforcement
  learning}.
In: \bbtitle{Thirty-sixth Conference on Neural Information Processing Systems
  Datasets and Benchmarks Track}
(\byear{2022}).
\burl{https://openreview.net/forum?id=D29JbExncTP}
\end{bchapter}
\endbibitem

%%% 26
\bibitem[\protect\citeauthoryear{Chen et~al.}{2021}]{chen2021variational}
\begin{barticle}
\bauthor{\bsnm{Chen}, \binits{J.}},
\bauthor{\bsnm{Zhang}, \binits{Y.}},
\bauthor{\bsnm{Xu}, \binits{Y.}},
\bauthor{\bsnm{Ma}, \binits{H.}},
\bauthor{\bsnm{Yang}, \binits{H.}},
\bauthor{\bsnm{Song}, \binits{J.}},
\bauthor{\bsnm{Wang}, \binits{Y.}},
\bauthor{\bsnm{Wu}, \binits{Y.}}:
\batitle{Variational automatic curriculum learning for sparse-reward
  cooperative multi-agent problems}.
\bjtitle{Advances in Neural Information Processing Systems}
\bvolume{34},
\bfpage{9681}--\blpage{9693}
(\byear{2021})
\end{barticle}
\endbibitem

%%% 27
\bibitem[\protect\citeauthoryear{Du et~al.}{2021}]{du2021learning}
\begin{bchapter}
\bauthor{\bsnm{Du}, \binits{Y.}},
\bauthor{\bsnm{Liu}, \binits{B.}},
\bauthor{\bsnm{Moens}, \binits{V.}},
\bauthor{\bsnm{Liu}, \binits{Z.}},
\bauthor{\bsnm{Ren}, \binits{Z.}},
\bauthor{\bsnm{Wang}, \binits{J.}},
\bauthor{\bsnm{Chen}, \binits{X.}},
\bauthor{\bsnm{Zhang}, \binits{H.}}:
\bctitle{Learning correlated communication topology in multi-agent
  reinforcement learning}.
In: \bbtitle{Proceedings of the 20th International Conference on Autonomous
  Agents and MultiAgent Systems},
pp. \bfpage{456}--\blpage{464}
(\byear{2021})
\end{bchapter}
\endbibitem

%%% 28
\bibitem[\protect\citeauthoryear{Da~Silva and Costa}{2019}]{da2019survey}
\begin{barticle}
\bauthor{\bsnm{Da~Silva}, \binits{F.L.}},
\bauthor{\bsnm{Costa}, \binits{A.H.R.}}:
\batitle{A survey on transfer learning for multiagent reinforcement learning
  systems}.
\bjtitle{Journal of Artificial Intelligence Research}
\bvolume{64},
\bfpage{645}--\blpage{703}
(\byear{2019})
\end{barticle}
\endbibitem

%%% 29
\bibitem[\protect\citeauthoryear{De~Witt et~al.}{2020}]{de2020independent}
\begin{botherref}
\oauthor{\bsnm{De~Witt}, \binits{C.S.}},
\oauthor{\bsnm{Gupta}, \binits{T.}},
\oauthor{\bsnm{Makoviichuk}, \binits{D.}},
\oauthor{\bsnm{Makoviychuk}, \binits{V.}},
\oauthor{\bsnm{Torr}, \binits{P.H.}},
\oauthor{\bsnm{Sun}, \binits{M.}},
\oauthor{\bsnm{Whiteson}, \binits{S.}}:
Is independent learning all you need in the starcraft multi-agent challenge?
arXiv preprint arXiv:2011.09533
(2020)
\end{botherref}
\endbibitem

%%% 30
\bibitem[\protect\citeauthoryear{Dong et~al.}{2024}]{dong2024villageragent}
\begin{botherref}
\oauthor{\bsnm{Dong}, \binits{Y.}},
\oauthor{\bsnm{Zhu}, \binits{X.}},
\oauthor{\bsnm{Pan}, \binits{Z.}},
\oauthor{\bsnm{Zhu}, \binits{L.}},
\oauthor{\bsnm{Yang}, \binits{Y.}}:
Villageragent: A graph-based multi-agent framework for coordinating complex
  task dependencies in minecraft.
arXiv preprint arXiv:2406.05720
(2024)
\end{botherref}
\endbibitem

%%% 31
\bibitem[\protect\citeauthoryear{Ellis et~al.}{2023}]{ellis2023smacv2}
\begin{bchapter}
\bauthor{\bsnm{Ellis}, \binits{B.}},
\bauthor{\bsnm{Cook}, \binits{J.}},
\bauthor{\bsnm{Moalla}, \binits{S.}},
\bauthor{\bsnm{Samvelyan}, \binits{M.}},
\bauthor{\bsnm{Sun}, \binits{M.}},
\bauthor{\bsnm{Mahajan}, \binits{A.}},
\bauthor{\bsnm{Foerster}, \binits{J.N.}},
\bauthor{\bsnm{Whiteson}, \binits{S.}}:
\bctitle{{SMAC}v2: An improved benchmark for cooperative multi-agent
  reinforcement learning}.
In: \bbtitle{Thirty-seventh Conference on Neural Information Processing Systems
  Datasets and Benchmarks Track}
(\byear{2023}).
\burl{https://openreview.net/forum?id=5OjLGiJW3u}
\end{bchapter}
\endbibitem

%%% 32
\bibitem[\protect\citeauthoryear{Egorov and
  Shpilman}{2022}]{egorov2022scalable}
\begin{botherref}
\oauthor{\bsnm{Egorov}, \binits{V.}},
\oauthor{\bsnm{Shpilman}, \binits{A.}}:
Scalable multi-agent model-based reinforcement learning.
arXiv preprint arXiv:2205.15023
(2022)
\end{botherref}
\endbibitem

%%% 33
\bibitem[\protect\citeauthoryear{Foerster et~al.}{2016}]{foerster2016learning}
\begin{botherref}
\oauthor{\bsnm{Foerster}, \binits{J.}},
\oauthor{\bsnm{Assael}, \binits{I.A.}},
\oauthor{\bsnm{De~Freitas}, \binits{N.}},
\oauthor{\bsnm{Whiteson}, \binits{S.}}:
Learning to communicate with deep multi-agent reinforcement learning.
Advances in neural information processing systems
\textbf{29}
(2016)
\end{botherref}
\endbibitem

%%% 34
\bibitem[\protect\citeauthoryear{Foerster et~al.}{2018}]{coma}
\begin{bchapter}
\bauthor{\bsnm{Foerster}, \binits{J.}},
\bauthor{\bsnm{Farquhar}, \binits{G.}},
\bauthor{\bsnm{Afouras}, \binits{T.}},
\bauthor{\bsnm{Nardelli}, \binits{N.}},
\bauthor{\bsnm{Whiteson}, \binits{S.}}:
\bctitle{Counterfactual multi-agent policy gradients}.
In: \bbtitle{Proceedings of the {AAAI} Conference on Artificial Intelligence},
pp. \bfpage{2974}--\blpage{2982}
(\byear{2018})
\end{bchapter}
\endbibitem

%%% 35
\bibitem[\protect\citeauthoryear{Feriani and Hossain}{2021}]{feriani2021single}
\begin{barticle}
\bauthor{\bsnm{Feriani}, \binits{A.}},
\bauthor{\bsnm{Hossain}, \binits{E.}}:
\batitle{Single and multi-agent deep reinforcement learning for ai-enabled
  wireless networks: A tutorial}.
\bjtitle{IEEE Communications Surveys \& Tutorials}
\bvolume{23}(\bissue{2}),
\bfpage{1226}--\blpage{1252}
(\byear{2021})
\end{barticle}
\endbibitem

%%% 36
\bibitem[\protect\citeauthoryear{Formanek et~al.}{2023}]{formanek2023ogmarl}
\begin{bchapter}
\bauthor{\bsnm{Formanek}, \binits{C.}},
\bauthor{\bsnm{Jeewa}, \binits{A.}},
\bauthor{\bsnm{Shock}, \binits{J.}},
\bauthor{\bsnm{Pretorius}, \binits{A.}}:
\bctitle{Off-the-grid marl: Datasets and baselines for offline multi-agent
  reinforcement learning}.
In: \bbtitle{Extended Abstract at the 2023 International Conference on
  Autonomous Agents and Multiagent Systems}.
\bpublisher{AAMAS}, \blocation{???}
(\byear{2023})
\end{bchapter}
\endbibitem

%%% 37
\bibitem[\protect\citeauthoryear{Fujimoto et~al.}{2019}]{fujimoto2019off}
\begin{bchapter}
\bauthor{\bsnm{Fujimoto}, \binits{S.}},
\bauthor{\bsnm{Meger}, \binits{D.}},
\bauthor{\bsnm{Precup}, \binits{D.}}:
\bctitle{Off-policy deep reinforcement learning without exploration}.
In: \bbtitle{International Conference on Machine Learning},
pp. \bfpage{2052}--\blpage{2062}
(\byear{2019}).
\bcomment{PMLR}
\end{bchapter}
\endbibitem

%%% 38
\bibitem[\protect\citeauthoryear{Fu et~al.}{2022}]{fu2022concentration}
\begin{bchapter}
\bauthor{\bsnm{Fu}, \binits{Q.}},
\bauthor{\bsnm{Qiu}, \binits{T.}},
\bauthor{\bsnm{Yi}, \binits{J.}},
\bauthor{\bsnm{Pu}, \binits{Z.}},
\bauthor{\bsnm{Wu}, \binits{S.}}:
\bctitle{Concentration network for reinforcement learning of large-scale
  multi-agent systems}.
In: \bbtitle{Proceedings of the AAAI Conference on Artificial Intelligence}
(\byear{2022})
\end{bchapter}
\endbibitem

%%% 39
\bibitem[\protect\citeauthoryear{Gronauer and
  Diepold}{2022}]{gronauer2022multi}
\begin{barticle}
\bauthor{\bsnm{Gronauer}, \binits{S.}},
\bauthor{\bsnm{Diepold}, \binits{K.}}:
\batitle{Multi-agent deep reinforcement learning: a survey}.
\bjtitle{Artificial Intelligence Review}
\bvolume{55}(\bissue{2}),
\bfpage{895}--\blpage{943}
(\byear{2022})
\end{barticle}
\endbibitem

%%% 40
\bibitem[\protect\citeauthoryear{Gupta et~al.}{2017}]{gupta2017cooperative}
\begin{bchapter}
\bauthor{\bsnm{Gupta}, \binits{J.K.}},
\bauthor{\bsnm{Egorov}, \binits{M.}},
\bauthor{\bsnm{Kochenderfer}, \binits{M.}}:
\bctitle{Cooperative multi-agent control using deep reinforcement learning}.
In: \bbtitle{International Conference on Autonomous Agents and Multiagent
  Systems},
pp. \bfpage{66}--\blpage{83}
(\byear{2017}).
\bcomment{Springer}
\end{bchapter}
\endbibitem

%%% 41
\bibitem[\protect\citeauthoryear{Gu et~al.}{2021}]{gu2021mean}
\begin{botherref}
\oauthor{\bsnm{Gu}, \binits{H.}},
\oauthor{\bsnm{Guo}, \binits{X.}},
\oauthor{\bsnm{Wei}, \binits{X.}},
\oauthor{\bsnm{Xu}, \binits{R.}}:
Mean-field multi-agent reinforcement learning: A decentralized network
  approach.
arXiv:2108.02731
(2021)
\end{botherref}
\endbibitem

%%% 42
\bibitem[\protect\citeauthoryear{Guo et~al.}{2019}]{guo2019learning}
\begin{botherref}
\oauthor{\bsnm{Guo}, \binits{X.}},
\oauthor{\bsnm{Hu}, \binits{A.}},
\oauthor{\bsnm{Xu}, \binits{R.}},
\oauthor{\bsnm{Zhang}, \binits{J.}}:
Learning mean-field games.
Advances in neural information processing systems
\textbf{32}
(2019)
\end{botherref}
\endbibitem

%%% 43
\bibitem[\protect\citeauthoryear{Gu et~al.}{2023}]{gu2023safe}
\begin{barticle}
\bauthor{\bsnm{Gu}, \binits{S.}},
\bauthor{\bsnm{Kuba}, \binits{J.G.}},
\bauthor{\bsnm{Chen}, \binits{Y.}},
\bauthor{\bsnm{Du}, \binits{Y.}},
\bauthor{\bsnm{Yang}, \binits{L.}},
\bauthor{\bsnm{Knoll}, \binits{A.}},
\bauthor{\bsnm{Yang}, \binits{Y.}}:
\batitle{Safe multi-agent reinforcement learning for multi-robot control}.
\bjtitle{Artificial Intelligence}
\bvolume{319},
\bfpage{103905}
(\byear{2023})
\end{barticle}
\endbibitem

%%% 44
\bibitem[\protect\citeauthoryear{Guestrin et~al.}{2001}]{guestrin2001}
\begin{bchapter}
\bauthor{\bsnm{Guestrin}, \binits{C.}},
\bauthor{\bsnm{Koller}, \binits{D.}},
\bauthor{\bsnm{Parr}, \binits{R.}}:
\bctitle{Multiagent planning with factored {MDP}s}.
In: \bbtitle{Proc. NIPS},
vol. \bseriesno{14},
pp. \bfpage{1523}--\blpage{1530}.
\bpublisher{MIT Press}, \blocation{???}
(\byear{2001})
\end{bchapter}
\endbibitem

%%% 45
\bibitem[\protect\citeauthoryear{Guestrin et~al.}{2003}]{guestrin2003efficient}
\begin{barticle}
\bauthor{\bsnm{Guestrin}, \binits{C.}},
\bauthor{\bsnm{Koller}, \binits{D.}},
\bauthor{\bsnm{Parr}, \binits{R.}},
\bauthor{\bsnm{Venkataraman}, \binits{S.}}:
\batitle{Efficient solution algorithms for factored mdps}.
\bjtitle{Journal of Artificial Intelligence Research}
\bvolume{19},
\bfpage{399}--\blpage{468}
(\byear{2003})
\end{barticle}
\endbibitem

%%% 46
\bibitem[\protect\citeauthoryear{Guestrin
  et~al.}{2002}]{guestrin2002coordinated}
\begin{bchapter}
\bauthor{\bsnm{Guestrin}, \binits{C.}},
\bauthor{\bsnm{Lagoudakis}, \binits{M.}},
\bauthor{\bsnm{Parr}, \binits{R.}}:
\bctitle{Coordinated reinforcement learning}.
In: \bbtitle{Proc. ICML},
vol. \bseriesno{2},
pp. \bfpage{227}--\blpage{234}
(\byear{2002})
\end{bchapter}
\endbibitem

%%% 47
\bibitem[\protect\citeauthoryear{Geng et~al.}{2026}]{geng2026scaling}
\begin{barticle}
\bauthor{\bsnm{Geng}, \binits{M.}},
\bauthor{\bsnm{Pateria}, \binits{S.}},
\bauthor{\bsnm{Subagdja}, \binits{B.}},
\bauthor{\bsnm{Tan}, \binits{A.-H.}}:
\batitle{Scaling up multi-agent reinforcement learning for large agent teams
  and long-horizon tasks: A survey}.
\bjtitle{ACM Computing Surveys}
(\byear{2026})
\doiurl{10.1145/3817113}
\end{barticle}
\endbibitem

%%% 48
\bibitem[\protect\citeauthoryear{Ganapathi~Subramanian
  et~al.}{2020}]{ganapathi2020multi}
\begin{bchapter}
\bauthor{\bsnm{Ganapathi~Subramanian}, \binits{S.}},
\bauthor{\bsnm{Poupart}, \binits{P.}},
\bauthor{\bsnm{Taylor}, \binits{M.E.}},
\bauthor{\bsnm{Hegde}, \binits{N.}}:
\bctitle{Multi type mean field reinforcement learning}.
In: \bbtitle{Proceedings of the 19th International Conference on Autonomous
  Agents and MultiAgent Systems},
pp. \bfpage{411}--\blpage{419}
(\byear{2020})
\end{bchapter}
\endbibitem

%%% 49
\bibitem[\protect\citeauthoryear{Ganapathi~Subramanian
  et~al.}{2021a}]{ganapathi2021partially}
\begin{bchapter}
\bauthor{\bsnm{Ganapathi~Subramanian}, \binits{S.}},
\bauthor{\bsnm{Taylor}, \binits{M.E.}},
\bauthor{\bsnm{Crowley}, \binits{M.}},
\bauthor{\bsnm{Poupart}, \binits{P.}}:
\bctitle{Partially observable mean field reinforcement learning}.
In: \bbtitle{Proceedings of the 20th International Conference on Autonomous
  Agents and MultiAgent Systems},
pp. \bfpage{537}--\blpage{545}
(\byear{2021})
\end{bchapter}
\endbibitem

%%% 50
\bibitem[\protect\citeauthoryear{Ganapathi~Subramanian
  et~al.}{2021b}]{subramanian2020partially}
\begin{bchapter}
\bauthor{\bsnm{Ganapathi~Subramanian}, \binits{S.}},
\bauthor{\bsnm{Taylor}, \binits{M.E.}},
\bauthor{\bsnm{Crowley}, \binits{M.}},
\bauthor{\bsnm{Poupart}, \binits{P.}}:
\bctitle{Partially observable mean field reinforcement learning}.
In: \bbtitle{Proc. AAMAS},
vol. \bseriesno{20},
pp. \bfpage{537}--\blpage{545}
(\byear{2021})
\end{bchapter}
\endbibitem

%%% 51
\bibitem[\protect\citeauthoryear{Gu et~al.}{2022}]{gu2022review}
\begin{botherref}
\oauthor{\bsnm{Gu}, \binits{S.}},
\oauthor{\bsnm{Yang}, \binits{L.}},
\oauthor{\bsnm{Du}, \binits{Y.}},
\oauthor{\bsnm{Chen}, \binits{G.}},
\oauthor{\bsnm{Walter}, \binits{F.}},
\oauthor{\bsnm{Wang}, \binits{J.}},
\oauthor{\bsnm{Knoll}, \binits{A.}}:
A review of safe reinforcement learning: Methods, theory and applications.
arXiv preprint arXiv:2205.10330
(2022)
\end{botherref}
\endbibitem

%%% 52
\bibitem[\protect\citeauthoryear{Hernandez-Leal
  et~al.}{2017}]{hernandez2017survey}
\begin{botherref}
\oauthor{\bsnm{Hernandez-Leal}, \binits{P.}},
\oauthor{\bsnm{Kaisers}, \binits{M.}},
\oauthor{\bsnm{Baarslag}, \binits{T.}},
\oauthor{\bsnm{De~Cote}, \binits{E.M.}}:
A survey of learning in multiagent environments: Dealing with non-stationarity.
arXiv preprint arXiv:1707.09183
(2017)
\end{botherref}
\endbibitem

%%% 53
\bibitem[\protect\citeauthoryear{Hu et~al.}{2020}]{hu2020other}
\begin{bchapter}
\bauthor{\bsnm{Hu}, \binits{H.}},
\bauthor{\bsnm{Lerer}, \binits{A.}},
\bauthor{\bsnm{Peysakhovich}, \binits{A.}},
\bauthor{\bsnm{Foerster}, \binits{J.}}:
\bctitle{“other-play” for zero-shot coordination}.
In: \bbtitle{International Conference on Machine Learning},
pp. \bfpage{4399}--\blpage{4410}
(\byear{2020}).
\bcomment{PMLR}
\end{bchapter}
\endbibitem

%%% 54
\bibitem[\protect\citeauthoryear{Hu et~al.}{2021}]{updet}
\begin{bchapter}
\bauthor{\bsnm{Hu}, \binits{S.}},
\bauthor{\bsnm{Zhu}, \binits{F.}},
\bauthor{\bsnm{Chang}, \binits{X.}},
\bauthor{\bsnm{Liang}, \binits{X.}}:
\bctitle{Updet: Universal multi-agent rl via policy decoupling with
  transformers}.
In: \bbtitle{International Conference on Learning Representations}
(\byear{2021})
\end{bchapter}
\endbibitem

%%% 55
\bibitem[\protect\citeauthoryear{Iqbal et~al.}{2021}]{iqbal2021randomized}
\begin{bchapter}
\bauthor{\bsnm{Iqbal}, \binits{S.}},
\bauthor{\bsnm{De~Witt}, \binits{C.A.S.}},
\bauthor{\bsnm{Peng}, \binits{B.}},
\bauthor{\bsnm{B{\"o}hmer}, \binits{W.}},
\bauthor{\bsnm{Whiteson}, \binits{S.}},
\bauthor{\bsnm{Sha}, \binits{F.}}:
\bctitle{Randomized entity-wise factorization for multi-agent reinforcement
  learning}.
In: \bbtitle{International Conference on Machine Learning},
pp. \bfpage{4596}--\blpage{4606}
(\byear{2021}).
\bcomment{PMLR}
\end{bchapter}
\endbibitem

%%% 56
\bibitem[\protect\citeauthoryear{Jeon et~al.}{2026}]{jeon2026stairsformer}
\begin{barticle}
\bauthor{\bsnm{Jeon}, \binits{J.}},
\bauthor{\bsnm{Cho}, \binits{M.}},
\bauthor{\bsnm{Sung}, \binits{Y.}}:
\batitle{{STAIRS}-former: Spatio-temporal attention with interleaved recursive
  structure transformer for offline multi-task multi-agent reinforcement
  learning}.
\bjtitle{arXiv preprint arXiv:2603.11691}
(\byear{2026})
\doiurl{10.48550/arXiv.2603.11691}
\end{barticle}
\endbibitem

%%% 57
\bibitem[\protect\citeauthoryear{Jiang et~al.}{2018}]{jiang2018graph}
\begin{botherref}
\oauthor{\bsnm{Jiang}, \binits{J.}},
\oauthor{\bsnm{Dun}, \binits{C.}},
\oauthor{\bsnm{Huang}, \binits{T.}},
\oauthor{\bsnm{Lu}, \binits{Z.}}:
Graph convolutional reinforcement learning.
arXiv preprint arXiv:1810.09202
(2018)
\end{botherref}
\endbibitem

%%% 58
\bibitem[\protect\citeauthoryear{Jianye et~al.}{2022}]{jianye2022boosting}
\begin{bchapter}
\bauthor{\bsnm{Jianye}, \binits{H.}},
\bauthor{\bsnm{Hao}, \binits{X.}},
\bauthor{\bsnm{Mao}, \binits{H.}},
\bauthor{\bsnm{Wang}, \binits{W.}},
\bauthor{\bsnm{Yang}, \binits{Y.}},
\bauthor{\bsnm{Li}, \binits{D.}},
\bauthor{\bsnm{Zheng}, \binits{Y.}},
\bauthor{\bsnm{Wang}, \binits{Z.}}:
\bctitle{Boosting multiagent reinforcement learning via permutation invariant
  and permutation equivariant networks}.
In: \bbtitle{The Eleventh International Conference on Learning Representations}
(\byear{2022})
\end{bchapter}
\endbibitem

%%% 59
\bibitem[\protect\citeauthoryear{Jia et~al.}{2022}]{jia2022crmrl}
\begin{botherref}
\oauthor{\bsnm{Jia}, \binits{H.}},
\oauthor{\bsnm{Zhao}, \binits{Y.}},
\oauthor{\bsnm{Zhai}, \binits{Y.}},
\oauthor{\bsnm{Ding}, \binits{B.}},
\oauthor{\bsnm{Wang}, \binits{H.}},
\oauthor{\bsnm{Wu}, \binits{Q.}}:
Crmrl: Collaborative relationship meta reinforcement learning for effectively
  adapting to type changes in multi-robotic system.
IEEE Robotics and Automation Letters
(2022)
\end{botherref}
\endbibitem

%%% 60
\bibitem[\protect\citeauthoryear{Kuba et~al.}{2022}]{hatrpohappo}
\begin{bchapter}
\bauthor{\bsnm{Kuba}, \binits{J.}},
\bauthor{\bsnm{Chen}, \binits{R.}},
\bauthor{\bsnm{Wen}, \binits{M.}},
\bauthor{\bsnm{Wen}, \binits{Y.}},
\bauthor{\bsnm{Sun}, \binits{F.}},
\bauthor{\bsnm{Wang}, \binits{J.}},
\bauthor{\bsnm{Yang}, \binits{Y.}}:
\bctitle{Trust region policy optimisation in multi-agent reinforcement
  learning}.
In: \bbtitle{ICLR 2022-10th International Conference on Learning
  Representations},
p. \bfpage{1046}
(\byear{2022}).
\bcomment{The International Conference on Learning Representations (ICLR)}
\end{bchapter}
\endbibitem

%%% 61
\bibitem[\protect\citeauthoryear{Kuba et~al.}{2022}]{kuba2022heterogeneous}
\begin{botherref}
\oauthor{\bsnm{Kuba}, \binits{J.G.}},
\oauthor{\bsnm{Feng}, \binits{X.}},
\oauthor{\bsnm{Ding}, \binits{S.}},
\oauthor{\bsnm{Dong}, \binits{H.}},
\oauthor{\bsnm{Wang}, \binits{J.}},
\oauthor{\bsnm{Yang}, \binits{Y.}}:
Heterogeneous-agent mirror learning: A continuum of solutions to cooperative
  marl.
arXiv preprint arXiv:2208.01682
(2022)
\end{botherref}
\endbibitem

%%% 62
\bibitem[\protect\citeauthoryear{Krajzewicz et~al.}{2002}]{krajzewicz2002sumo}
\begin{bchapter}
\bauthor{\bsnm{Krajzewicz}, \binits{D.}},
\bauthor{\bsnm{Hertkorn}, \binits{G.}},
\bauthor{\bsnm{R{\"o}ssel}, \binits{C.}},
\bauthor{\bsnm{Wagner}, \binits{P.}}:
\bctitle{Sumo (simulation of urban mobility)-an open-source traffic
  simulation}.
In: \bbtitle{Proceedings of the 4th Middle East Symposium on Simulation and
  Modelling (MESM20002)},
pp. \bfpage{183}--\blpage{187}
(\byear{2002})
\end{bchapter}
\endbibitem

%%% 63
\bibitem[\protect\citeauthoryear{Kostrikov et~al.}{2021}]{kostrikov2021offline}
\begin{botherref}
\oauthor{\bsnm{Kostrikov}, \binits{I.}},
\oauthor{\bsnm{Nair}, \binits{A.}},
\oauthor{\bsnm{Levine}, \binits{S.}}:
Offline reinforcement learning with implicit q-learning.
arXiv preprint arXiv:2110.06169
(2021)
\end{botherref}
\endbibitem

%%% 64
\bibitem[\protect\citeauthoryear{Kurach et~al.}{2020}]{kurach2020google}
\begin{bchapter}
\bauthor{\bsnm{Kurach}, \binits{K.}},
\bauthor{\bsnm{Raichuk}, \binits{A.}},
\bauthor{\bsnm{Sta{\'n}czyk}, \binits{P.}},
\bauthor{\bsnm{Zaj{\k{a}}c}, \binits{M.}},
\bauthor{\bsnm{Bachem}, \binits{O.}},
\bauthor{\bsnm{Espeholt}, \binits{L.}},
\bauthor{\bsnm{Riquelme}, \binits{C.}},
\bauthor{\bsnm{Vincent}, \binits{D.}},
\bauthor{\bsnm{Michalski}, \binits{M.}},
\bauthor{\bsnm{Bousquet}, \binits{O.}}, \betal:
\bctitle{Google research football: A novel reinforcement learning environment}.
In: \bbtitle{Proceedings of the AAAI Conference on Artificial Intelligence},
vol. \bseriesno{34},
pp. \bfpage{4501}--\blpage{4510}
(\byear{2020})
\end{bchapter}
\endbibitem

%%% 65
\bibitem[\protect\citeauthoryear{Kok and Vlassis}{2004}]{kok2004sparse}
\begin{bchapter}
\bauthor{\bsnm{Kok}, \binits{J.R.}},
\bauthor{\bsnm{Vlassis}, \binits{N.}}:
\bctitle{Sparse cooperative q-learning}.
In: \bbtitle{Proceedings of the Twenty-first International Conference on
  Machine Learning},
p. \bfpage{61}
(\byear{2004})
\end{bchapter}
\endbibitem

%%% 66
\bibitem[\protect\citeauthoryear{Kok and Vlassis}{2006}]{kok2006}
\begin{barticle}
\bauthor{\bsnm{Kok}, \binits{J.R.}},
\bauthor{\bsnm{Vlassis}, \binits{N.}}:
\batitle{Collaborative multiagent reinforcement learning by payoff
  propagation}.
\bjtitle{J. Mach. Learn. Res.}
\bvolume{7}(\bissue{65}),
\bfpage{1789}--\blpage{1828}
(\byear{2006})
\end{barticle}
\endbibitem

%%% 67
\bibitem[\protect\citeauthoryear{Kumar et~al.}{2020}]{kumar2020conservative}
\begin{barticle}
\bauthor{\bsnm{Kumar}, \binits{A.}},
\bauthor{\bsnm{Zhou}, \binits{A.}},
\bauthor{\bsnm{Tucker}, \binits{G.}},
\bauthor{\bsnm{Levine}, \binits{S.}}:
\batitle{Conservative q-learning for offline reinforcement learning}.
\bjtitle{Advances in neural information processing systems}
\bvolume{33},
\bfpage{1179}--\blpage{1191}
(\byear{2020})
\end{barticle}
\endbibitem

%%% 68
\bibitem[\protect\citeauthoryear{Lucas and Allen}{2022}]{lucas2022any}
\begin{bchapter}
\bauthor{\bsnm{Lucas}, \binits{K.}},
\bauthor{\bsnm{Allen}, \binits{R.E.}}:
\bctitle{Any-play: An intrinsic augmentation for zero-shot coordination}.
In: \bbtitle{Proceedings of the 21st International Conference on Autonomous
  Agents and Multiagent Systems},
pp. \bfpage{853}--\blpage{861}
(\byear{2022})
\end{bchapter}
\endbibitem

%%% 69
\bibitem[\protect\citeauthoryear{Lupu et~al.}{2021}]{lupu2021trajectory}
\begin{bchapter}
\bauthor{\bsnm{Lupu}, \binits{A.}},
\bauthor{\bsnm{Cui}, \binits{B.}},
\bauthor{\bsnm{Hu}, \binits{H.}},
\bauthor{\bsnm{Foerster}, \binits{J.}}:
\bctitle{Trajectory diversity for zero-shot coordination}.
In: \bbtitle{International Conference on Machine Learning},
pp. \bfpage{7204}--\blpage{7213}
(\byear{2021}).
\bcomment{PMLR}
\end{bchapter}
\endbibitem

%%% 70
\bibitem[\protect\citeauthoryear{Li et~al.}{2025}]{li2025nucleolus}
\begin{botherref}
\oauthor{\bsnm{Li}, \binits{Y.}},
\oauthor{\bsnm{Cao}, \binits{Z.}},
\oauthor{\bsnm{Qiao}, \binits{J.}},
\oauthor{\bsnm{Hu}, \binits{S.}}:
Nucleolus credit assignment for effective coalitions in multi-agent
  reinforcement learning.
arXiv preprint arXiv:2503.00372
(2025)
\end{botherref}
\endbibitem

%%% 71
\bibitem[\protect\citeauthoryear{Light et~al.}{2023}]{light2023avalonbench}
\begin{botherref}
\oauthor{\bsnm{Light}, \binits{J.}},
\oauthor{\bsnm{Cai}, \binits{M.}},
\oauthor{\bsnm{Shen}, \binits{S.}},
\oauthor{\bsnm{Hu}, \binits{Z.}}:
Avalonbench: Evaluating llms playing the game of avalon.
URL https://arxiv. org/abs/2310.05036
(2023)
\end{botherref}
\endbibitem

%%% 72
\bibitem[\protect\citeauthoryear{Li et~al.}{2024}]{li2024multi}
\begin{botherref}
\oauthor{\bsnm{Li}, \binits{C.}},
\oauthor{\bsnm{Dong}, \binits{S.}},
\oauthor{\bsnm{Yang}, \binits{S.}},
\oauthor{\bsnm{Hu}, \binits{Y.}},
\oauthor{\bsnm{Ding}, \binits{T.}},
\oauthor{\bsnm{Li}, \binits{W.}},
\oauthor{\bsnm{Gao}, \binits{Y.}}:
Multi-task multi-agent reinforcement learning with interaction and task
  representations.
IEEE Transactions on Neural Networks and Learning Systems
(2024)
\end{botherref}
\endbibitem

%%% 73
\bibitem[\protect\citeauthoryear{Liu et~al.}{2021}]{liu2021cmix}
\begin{bchapter}
\bauthor{\bsnm{Liu}, \binits{C.}},
\bauthor{\bsnm{Geng}, \binits{N.}},
\bauthor{\bsnm{Aggarwal}, \binits{V.}},
\bauthor{\bsnm{Lan}, \binits{T.}},
\bauthor{\bsnm{Yang}, \binits{Y.}},
\bauthor{\bsnm{Xu}, \binits{M.}}:
\bctitle{Cmix: Deep multi-agent reinforcement learning with peak and average
  constraints}.
In: \bbtitle{Machine Learning and Knowledge Discovery in Databases. Research
  Track: European Conference, ECML PKDD 2021, Bilbao, Spain, September 13--17,
  2021, Proceedings, Part I 21},
pp. \bfpage{157}--\blpage{173}
(\byear{2021}).
\bcomment{Springer}
\end{bchapter}
\endbibitem

%%% 74
\bibitem[\protect\citeauthoryear{Liu et~al.}{2022}]{liu2022light}
\begin{botherref}
\oauthor{\bsnm{Liu}, \binits{Q.}},
\oauthor{\bsnm{Jiang}, \binits{Y.}},
\oauthor{\bsnm{Ma}, \binits{X.}}:
Light Aircraft Game: A lightweight, scalable, gym-wrapped aircraft competitive
  environment with baseline reinforcement learning algorithms.
GitHub
(2022)
\end{botherref}
\endbibitem

%%% 75
\bibitem[\protect\citeauthoryear{Lin and Lee}{2024}]{lin2024hgap}
\begin{bchapter}
\bauthor{\bsnm{Lin}, \binits{B.-J.}},
\bauthor{\bsnm{Lee}, \binits{C.-Y.}}:
\bctitle{Hgap: boosting permutation invariant and permutation equivariant in
  multi-agent reinforcement learning via graph attention network}.
In: \bbtitle{Forty-first International Conference on Machine Learning}
(\byear{2024})
\end{bchapter}
\endbibitem

%%% 76
\bibitem[\protect\citeauthoryear{Li et~al.}{2026}]{li2026logo}
\begin{barticle}
\bauthor{\bsnm{Li}, \binits{S.}},
\bauthor{\bsnm{Li}, \binits{X.}},
\bauthor{\bsnm{Chen}, \binits{S.}},
\bauthor{\bsnm{Zhang}, \binits{J.}}:
\batitle{Puzzle it out: Local-to-global world model for offline multi-agent
  reinforcement learning}.
\bjtitle{arXiv preprint arXiv:2601.07463}
(\byear{2026})
\doiurl{10.48550/arXiv.2601.07463}
\end{barticle}
\endbibitem

%%% 77
\bibitem[\protect\citeauthoryear{Li et~al.}{2022}]{li2022metadrive}
\begin{barticle}
\bauthor{\bsnm{Li}, \binits{Q.}},
\bauthor{\bsnm{Peng}, \binits{Z.}},
\bauthor{\bsnm{Feng}, \binits{L.}},
\bauthor{\bsnm{Zhang}, \binits{Q.}},
\bauthor{\bsnm{Xue}, \binits{Z.}},
\bauthor{\bsnm{Zhou}, \binits{B.}}:
\batitle{Metadrive: Composing diverse driving scenarios for generalizable
  reinforcement learning}.
\bjtitle{IEEE transactions on pattern analysis and machine intelligence}
\bvolume{45}(\bissue{3}),
\bfpage{3461}--\blpage{3475}
(\byear{2022})
\end{barticle}
\endbibitem

%%% 78
\bibitem[\protect\citeauthoryear{Lin et~al.}{2021}]{lin2021multi}
\begin{barticle}
\bauthor{\bsnm{Lin}, \binits{Y.}},
\bauthor{\bsnm{Qu}, \binits{G.}},
\bauthor{\bsnm{Huang}, \binits{L.}},
\bauthor{\bsnm{Wierman}, \binits{A.}}:
\batitle{Multi-agent reinforcement learning in stochastic networked systems}.
\bjtitle{Advances in neural information processing systems}
\bvolume{34},
\bfpage{7825}--\blpage{7837}
(\byear{2021})
\end{barticle}
\endbibitem

%%% 79
\bibitem[\protect\citeauthoryear{Liu et~al.}{2025}]{liu2025learning}
\begin{bchapter}
\bauthor{\bsnm{Liu}, \binits{S.}},
\bauthor{\bsnm{Shu}, \binits{Y.}},
\bauthor{\bsnm{Guo}, \binits{C.}},
\bauthor{\bsnm{Yang}, \binits{B.}}:
\bctitle{Learning generalizable skills from offline multi-task data for
  multi-agent cooperation}.
In: \bbtitle{International Conference on Learning Representations}
(\byear{2025})
\end{bchapter}
\endbibitem

%%% 80
\bibitem[\protect\citeauthoryear{Lowe et~al.}{2017}]{maddpg}
\begin{bchapter}
\bauthor{\bsnm{Lowe}, \binits{R.}},
\bauthor{\bsnm{Wu}, \binits{Y.}},
\bauthor{\bsnm{Tamar}, \binits{A.}},
\bauthor{\bsnm{Harb}, \binits{J.}},
\bauthor{\bsnm{Abbeel}, \binits{P.}},
\bauthor{\bsnm{Mordatch}, \binits{I.}}:
\bctitle{Multi-agent actor-critic for mixed cooperative-competitive
  environments}.
In: \bbtitle{Advances in Neural Information Processing Systems},
pp. \bfpage{6379}--\blpage{6390}
(\byear{2017})
\end{bchapter}
\endbibitem

%%% 81
\bibitem[\protect\citeauthoryear{Lu et~al.}{2021}]{lu2021decentralized}
\begin{bchapter}
\bauthor{\bsnm{Lu}, \binits{S.}},
\bauthor{\bsnm{Zhang}, \binits{K.}},
\bauthor{\bsnm{Chen}, \binits{T.}},
\bauthor{\bsnm{Ba{\c{s}}ar}, \binits{T.}},
\bauthor{\bsnm{Horesh}, \binits{L.}}:
\bctitle{Decentralized policy gradient descent ascent for safe multi-agent
  reinforcement learning}.
In: \bbtitle{Proceedings of the AAAI Conference on Artificial Intelligence},
vol. \bseriesno{35},
pp. \bfpage{8767}--\blpage{8775}
(\byear{2021})
\end{bchapter}
\endbibitem

%%% 82
\bibitem[\protect\citeauthoryear{Long et~al.}{2020}]{long2020evolutionary}
\begin{bchapter}
\bauthor{\bsnm{Long}, \binits{Q.}},
\bauthor{\bsnm{Zhou}, \binits{Z.}},
\bauthor{\bsnm{Gupta}, \binits{A.}},
\bauthor{\bsnm{Fang}, \binits{F.}},
\bauthor{\bsnm{Wu}, \binits{Y.}},
\bauthor{\bsnm{Wang}, \binits{X.}}:
\bctitle{Evolutionary population curriculum for scaling multi-agent
  reinforcement learning}.
In: \bbtitle{International Conference on Learning Representations}
(\byear{2020})
\end{bchapter}
\endbibitem

%%% 83
\bibitem[\protect\citeauthoryear{Liu et~al.}{2024}]{hasac}
\begin{bchapter}
\bauthor{\bsnm{Liu}, \binits{J.}},
\bauthor{\bsnm{Zhong}, \binits{Y.}},
\bauthor{\bsnm{Hu}, \binits{S.}},
\bauthor{\bsnm{Fu}, \binits{H.}},
\bauthor{\bsnm{FU}, \binits{Q.}},
\bauthor{\bsnm{Chang}, \binits{X.}},
\bauthor{\bsnm{Yang}, \binits{Y.}}:
\bctitle{Maximum entropy heterogeneous-agent reinforcement learning}.
In: \bbtitle{The Twelfth International Conference on Learning Representations}
(\byear{2024}).
\burl{https://openreview.net/forum?id=tmqOhBC4a5}
\end{bchapter}
\endbibitem

%%% 84
\bibitem[\protect\citeauthoryear{Leibo et~al.}{2017}]{leibo2017multi}
\begin{bchapter}
\bauthor{\bsnm{Leibo}, \binits{J.Z.}},
\bauthor{\bsnm{Zambaldi}, \binits{V.}},
\bauthor{\bsnm{Lanctot}, \binits{M.}},
\bauthor{\bsnm{Marecki}, \binits{J.}},
\bauthor{\bsnm{Graepel}, \binits{T.}}:
\bctitle{Multi-agent reinforcement learning in sequential social dilemmas}.
In: \bbtitle{International Conference on Autonomous Agents and Multiagent
  Systems}
(\byear{2017})
\end{bchapter}
\endbibitem

%%% 85
\bibitem[\protect\citeauthoryear{Liu et~al.}{2023}]{liumaskma}
\begin{botherref}
\oauthor{\bsnm{Liu}, \binits{J.}},
\oauthor{\bsnm{Zhang}, \binits{Y.}},
\oauthor{\bsnm{Li}, \binits{C.}},
\oauthor{\bsnm{You}, \binits{Z.}},
\oauthor{\bsnm{Zhou}, \binits{Z.}},
\oauthor{\bsnm{Yang}, \binits{C.}},
\oauthor{\bsnm{Yang}, \binits{Y.}},
\oauthor{\bsnm{Liu}, \binits{Y.}},
\oauthor{\bsnm{Ouyang}, \binits{W.}}:
Maskma: Towards zero-shot multi-agent decision making with mask-based
  collaborative learning.
Transactions on Machine Learning Research
(2023)
\end{botherref}
\endbibitem

%%% 86
\bibitem[\protect\citeauthoryear{Li et~al.}{2023}]{li2023cooperative}
\begin{bchapter}
\bauthor{\bsnm{Li}, \binits{Y.}},
\bauthor{\bsnm{Zhang}, \binits{S.}},
\bauthor{\bsnm{Sun}, \binits{J.}},
\bauthor{\bsnm{Du}, \binits{Y.}},
\bauthor{\bsnm{Wen}, \binits{Y.}},
\bauthor{\bsnm{Wang}, \binits{X.}},
\bauthor{\bsnm{Pan}, \binits{W.}}:
\bctitle{Cooperative open-ended learning framework for zero-shot coordination}.
In: \bbtitle{International Conference on Machine Learning},
pp. \bfpage{20470}--\blpage{20484}
(\byear{2023}).
\bcomment{PMLR}
\end{bchapter}
\endbibitem

%%% 87
\bibitem[\protect\citeauthoryear{Mondal et~al.}{2022}]{mondal2022approximation}
\begin{barticle}
\bauthor{\bsnm{Mondal}, \binits{W.U.}},
\bauthor{\bsnm{Agarwal}, \binits{M.}},
\bauthor{\bsnm{Aggarwal}, \binits{V.}},
\bauthor{\bsnm{Ukkusuri}, \binits{S.V.}}:
\batitle{On the approximation of cooperative heterogeneous multi-agent
  reinforcement learning (marl) using mean field control (mfc)}.
\bjtitle{Journal of Machine Learning Research}
\bvolume{23}(\bissue{129}),
\bfpage{1}--\blpage{46}
(\byear{2022})
\end{barticle}
\endbibitem

%%% 88
\bibitem[\protect\citeauthoryear{Melcer et~al.}{2022}]{melcer2022shield}
\begin{barticle}
\bauthor{\bsnm{Melcer}, \binits{D.}},
\bauthor{\bsnm{Amato}, \binits{C.}},
\bauthor{\bsnm{Tripakis}, \binits{S.}}:
\batitle{Shield decentralization for safe multi-agent reinforcement learning}.
\bjtitle{Advances in Neural Information Processing Systems}
\bvolume{35},
\bfpage{13367}--\blpage{13379}
(\byear{2022})
\end{barticle}
\endbibitem

%%% 89
\bibitem[\protect\citeauthoryear{Monroc et~al.}{2025}]{monroc2025wfcrl}
\begin{botherref}
\oauthor{\bsnm{Monroc}, \binits{C.B.}},
\oauthor{\bsnm{Bu{\v{s}}i{\'c}}, \binits{A.}},
\oauthor{\bsnm{Dubuc}, \binits{D.}},
\oauthor{\bsnm{Zhu}, \binits{J.}}:
Wfcrl: A multi-agent reinforcement learning benchmark for wind farm control.
arXiv preprint arXiv:2501.13592
(2025)
\end{botherref}
\endbibitem

%%% 90
\bibitem[\protect\citeauthoryear{Mukobi et~al.}{2023}]{mukobi2023welfare}
\begin{botherref}
\oauthor{\bsnm{Mukobi}, \binits{G.}},
\oauthor{\bsnm{Erlebach}, \binits{H.}},
\oauthor{\bsnm{Lauffer}, \binits{N.}},
\oauthor{\bsnm{Hammond}, \binits{L.}},
\oauthor{\bsnm{Chan}, \binits{A.}},
\oauthor{\bsnm{Clifton}, \binits{J.}}:
Welfare diplomacy: Benchmarking language model cooperation.
arXiv preprint arXiv:2310.08901
(2023)
\end{botherref}
\endbibitem

%%% 91
\bibitem[\protect\citeauthoryear{Mnih et~al.}{2015}]{mnih2015human}
\begin{barticle}
\bauthor{\bsnm{Mnih}, \binits{V.}},
\bauthor{\bsnm{Kavukcuoglu}, \binits{K.}},
\bauthor{\bsnm{Silver}, \binits{D.}},
\bauthor{\bsnm{Rusu}, \binits{A.A.}},
\bauthor{\bsnm{Veness}, \binits{J.}},
\bauthor{\bsnm{Bellemare}, \binits{M.G.}},
\bauthor{\bsnm{Graves}, \binits{A.}},
\bauthor{\bsnm{Riedmiller}, \binits{M.}},
\bauthor{\bsnm{Fidjeland}, \binits{A.K.}},
\bauthor{\bsnm{Ostrovski}, \binits{G.}}, \betal:
\batitle{Human-level control through deep reinforcement learning}.
\bjtitle{nature}
\bvolume{518}(\bissue{7540}),
\bfpage{529}--\blpage{533}
(\byear{2015})
\end{barticle}
\endbibitem

%%% 92
\bibitem[\protect\citeauthoryear{Mohanty et~al.}{2020}]{mohanty2020flatland}
\begin{botherref}
\oauthor{\bsnm{Mohanty}, \binits{S.}},
\oauthor{\bsnm{Nygren}, \binits{E.}},
\oauthor{\bsnm{Laurent}, \binits{F.}},
\oauthor{\bsnm{Schneider}, \binits{M.}},
\oauthor{\bsnm{Scheller}, \binits{C.}},
\oauthor{\bsnm{Bhattacharya}, \binits{N.}},
\oauthor{\bsnm{Watson}, \binits{J.}},
\oauthor{\bsnm{Egli}, \binits{A.}},
\oauthor{\bsnm{Eichenberger}, \binits{C.}},
\oauthor{\bsnm{Baumberger}, \binits{C.}}, et al.:
Flatland-rl: Multi-agent reinforcement learning on trains.
arXiv preprint arXiv:2012.05893
(2020)
\end{botherref}
\endbibitem

%%% 93
\bibitem[\protect\citeauthoryear{Meng et~al.}{2023}]{meng2023m3}
\begin{bchapter}
\bauthor{\bsnm{Meng}, \binits{L.}},
\bauthor{\bsnm{Ruan}, \binits{J.}},
\bauthor{\bsnm{Xiong}, \binits{X.}},
\bauthor{\bsnm{Li}, \binits{X.}},
\bauthor{\bsnm{Zhang}, \binits{X.}},
\bauthor{\bsnm{Xing}, \binits{D.}},
\bauthor{\bsnm{Xu}, \binits{B.}}:
\bctitle{M3: Modularization for multi-task and multi-agent offline
  pre-training}.
In: \bbtitle{Proceedings of the 2023 International Conference on Autonomous
  Agents and Multiagent Systems}
(\byear{2023})
\end{bchapter}
\endbibitem

%%% 94
\bibitem[\protect\citeauthoryear{Meng et~al.}{2023}]{meng2023offline}
\begin{botherref}
\oauthor{\bsnm{Meng}, \binits{L.}},
\oauthor{\bsnm{Wen}, \binits{M.}},
\oauthor{\bsnm{Le}, \binits{C.}},
\oauthor{\bsnm{Li}, \binits{X.}},
\oauthor{\bsnm{Xing}, \binits{D.}},
\oauthor{\bsnm{Zhang}, \binits{W.}},
\oauthor{\bsnm{Wen}, \binits{Y.}},
\oauthor{\bsnm{Zhang}, \binits{H.}},
\oauthor{\bsnm{Wang}, \binits{J.}},
\oauthor{\bsnm{Yang}, \binits{Y.}}, et al.:
Offline pre-trained multi-agent decision transformer.
Machine Intelligence Research
(2023)
\end{botherref}
\endbibitem

%%% 95
\bibitem[\protect\citeauthoryear{Naug et~al.}{2024}]{naug2024sustaindc}
\begin{barticle}
\bauthor{\bsnm{Naug}, \binits{A.}},
\bauthor{\bsnm{Guillen-Perez}, \binits{A.}},
\bauthor{\bsnm{Luna~Gutierrez}, \binits{R.}},
\bauthor{\bsnm{Gundecha}, \binits{V.}},
\bauthor{\bsnm{Bash}, \binits{C.}},
\bauthor{\bsnm{Ghorbanpour}, \binits{S.}},
\bauthor{\bsnm{Mousavi}, \binits{S.}},
\bauthor{\bsnm{Ramesh~Babu}, \binits{A.}},
\bauthor{\bsnm{Markovikj}, \binits{D.}},
\bauthor{\bsnm{Dheeraj~Kashyap}, \binits{L.}}, \betal:
\batitle{Sustaindc: Benchmarking for sustainable data center control}.
\bjtitle{Advances in Neural Information Processing Systems}
\bvolume{37},
\bfpage{100630}--\blpage{100669}
(\byear{2024})
\end{barticle}
\endbibitem

%%% 96
\bibitem[\protect\citeauthoryear{Na et~al.}{2025}]{na2025trajectory}
\begin{bchapter}
\bauthor{\bsnm{Na}, \binits{H.}},
\bauthor{\bsnm{Lee}, \binits{K.}},
\bauthor{\bsnm{Lee}, \binits{S.}},
\bauthor{\bsnm{Moon}, \binits{I.-C.}}:
\bctitle{Trajectory-class-aware multi-agent reinforcement learning}.
In: \bbtitle{The Thirteenth International Conference on Learning
  Representations}
(\byear{2025})
\end{bchapter}
\endbibitem

%%% 97
\bibitem[\protect\citeauthoryear{Ni et~al.}{2026}]{ni2026tomzsc}
\begin{barticle}
\bauthor{\bsnm{Ni}, \binits{A.}},
\bauthor{\bsnm{Stepputtis}, \binits{S.}},
\bauthor{\bsnm{Nikolaidis}, \binits{S.}},
\bauthor{\bsnm{Lewis}, \binits{M.}},
\bauthor{\bsnm{Sycara}, \binits{K.P.}},
\bauthor{\bsnm{Kim}, \binits{W.}}:
\batitle{Theory of mind guided strategy adaptation for zero-shot coordination}.
\bjtitle{arXiv preprint arXiv:2602.12458}
(\byear{2026})
\doiurl{10.48550/arXiv.2602.12458}
\end{barticle}
\endbibitem

%%% 98
\bibitem[\protect\citeauthoryear{Nair et~al.}{2005}]{nair2005networked}
\begin{bchapter}
\bauthor{\bsnm{Nair}, \binits{R.}},
\bauthor{\bsnm{Varakantham}, \binits{P.}},
\bauthor{\bsnm{Tambe}, \binits{M.}},
\bauthor{\bsnm{Yokoo}, \binits{M.}}:
\bctitle{Networked distributed pomdps: A synthesis of distributed constraint
  optimization and pomdps}.
In: \bbtitle{AAAI},
vol. \bseriesno{5},
pp. \bfpage{133}--\blpage{139}
(\byear{2005})
\end{bchapter}
\endbibitem

%%% 99
\bibitem[\protect\citeauthoryear{Orr and Dutta}{2023}]{orr2023multi}
\begin{barticle}
\bauthor{\bsnm{Orr}, \binits{J.}},
\bauthor{\bsnm{Dutta}, \binits{A.}}:
\batitle{Multi-agent deep reinforcement learning for multi-robot applications:
  A survey}.
\bjtitle{Sensors}
\bvolume{23}(\bissue{7}),
\bfpage{3625}
(\byear{2023})
\end{barticle}
\endbibitem

%%% 100
\bibitem[\protect\citeauthoryear{Oroojlooy and
  Hajinezhad}{2023}]{oroojlooy2023review}
\begin{barticle}
\bauthor{\bsnm{Oroojlooy}, \binits{A.}},
\bauthor{\bsnm{Hajinezhad}, \binits{D.}}:
\batitle{A review of cooperative multi-agent deep reinforcement learning}.
\bjtitle{Applied Intelligence}
\bvolume{53}(\bissue{11}),
\bfpage{13677}--\blpage{13722}
(\byear{2023})
\end{barticle}
\endbibitem

%%% 101
\bibitem[\protect\citeauthoryear{Oliehoek
  et~al.}{2013}]{oliehoek2013approximate}
\begin{bchapter}
\bauthor{\bsnm{Oliehoek}, \binits{F.A.}},
\bauthor{\bsnm{Whiteson}, \binits{S.}},
\bauthor{\bsnm{Spaan}, \binits{M.T.}}, \betal:
\bctitle{Approximate solutions for factored dec-pomdps with many agents.}
In: \bbtitle{AAMAS},
pp. \bfpage{563}--\blpage{570}
(\byear{2013})
\end{bchapter}
\endbibitem

%%% 102
\bibitem[\protect\citeauthoryear{Papoudakis et~al.}{2021}]{RWAREandLBF}
\begin{bchapter}
\bauthor{\bsnm{Papoudakis}, \binits{G.}},
\bauthor{\bsnm{Christianos}, \binits{F.}},
\bauthor{\bsnm{Sch{\"a}fer}, \binits{L.}},
\bauthor{\bsnm{Albrecht}, \binits{S.V.}}:
\bctitle{Benchmarking multi-agent deep reinforcement learning algorithms in
  cooperative tasks}.
In: \bbtitle{Thirty-fifth Conference on Neural Information Processing Systems
  Datasets and Benchmarks Track (Round 1)}
(\byear{2021})
\end{bchapter}
\endbibitem

%%% 103
\bibitem[\protect\citeauthoryear{Papoudakis
  et~al.}{2022}]{papoudakis1benchmarking}
\begin{bchapter}
\bauthor{\bsnm{Papoudakis}, \binits{G.}},
\bauthor{\bsnm{Christianos}, \binits{F.}},
\bauthor{\bsnm{Sch{\"a}fer}, \binits{L.}},
\bauthor{\bsnm{Albrecht}, \binits{S.V.}}:
\bctitle{Benchmarking multi-agent deep reinforcement learning algorithms in
  cooperative tasks}.
In: \bbtitle{Thirty-fifth Conference on Neural Information Processing Systems
  Datasets and Benchmarks Track (Round 1)}
(\byear{2022})
\end{bchapter}
\endbibitem

%%% 104
\bibitem[\protect\citeauthoryear{Pan et~al.}{2022}]{pan2022plan}
\begin{bchapter}
\bauthor{\bsnm{Pan}, \binits{L.}},
\bauthor{\bsnm{Huang}, \binits{L.}},
\bauthor{\bsnm{Ma}, \binits{T.}},
\bauthor{\bsnm{Xu}, \binits{H.}}:
\bctitle{Plan better amid conservatism: Offline multi-agent reinforcement
  learning with actor rectification}.
In: \bbtitle{International Conference on Machine Learning},
pp. \bfpage{17221}--\blpage{17237}
(\byear{2022}).
\bcomment{PMLR}
\end{bchapter}
\endbibitem

%%% 105
\bibitem[\protect\citeauthoryear{P{\'a}sztor et~al.}{2021}]{pasztorefficient}
\begin{botherref}
\oauthor{\bsnm{P{\'a}sztor}, \binits{B.}},
\oauthor{\bsnm{Krause}, \binits{A.}},
\oauthor{\bsnm{Bogunovic}, \binits{I.}}:
Efficient model-based multi-agent mean-field reinforcement learning.
Transactions on Machine Learning Research
(2021)
\end{botherref}
\endbibitem

%%% 106
\bibitem[\protect\citeauthoryear{Perez-Liebana et~al.}{2019}]{perez2019multi}
\begin{botherref}
\oauthor{\bsnm{Perez-Liebana}, \binits{D.}},
\oauthor{\bsnm{Hofmann}, \binits{K.}},
\oauthor{\bsnm{Mohanty}, \binits{S.P.}},
\oauthor{\bsnm{Kuno}, \binits{N.}},
\oauthor{\bsnm{Kramer}, \binits{A.}},
\oauthor{\bsnm{Devlin}, \binits{S.}},
\oauthor{\bsnm{Gaina}, \binits{R.D.}},
\oauthor{\bsnm{Ionita}, \binits{D.}}:
The multi-agent reinforcement learning in malm$\backslash$" o
  (marl$\backslash$" o) competition.
arXiv preprint arXiv:1901.08129
(2019)
\end{botherref}
\endbibitem

%%% 107
\bibitem[\protect\citeauthoryear{Pan et~al.}{2022}]{pan2022mate}
\begin{barticle}
\bauthor{\bsnm{Pan}, \binits{X.}},
\bauthor{\bsnm{Liu}, \binits{M.}},
\bauthor{\bsnm{Zhong}, \binits{F.}},
\bauthor{\bsnm{Yang}, \binits{Y.}},
\bauthor{\bsnm{Zhu}, \binits{S.-C.}},
\bauthor{\bsnm{Wang}, \binits{Y.}}:
\batitle{Mate: Benchmarking multi-agent reinforcement learning in distributed
  target coverage control}.
\bjtitle{Advances in Neural Information Processing Systems}
\bvolume{35},
\bfpage{27862}--\blpage{27879}
(\byear{2022})
\end{barticle}
\endbibitem

%%% 108
\bibitem[\protect\citeauthoryear{Peng et~al.}{2021}]{facmac}
\begin{bchapter}
\bauthor{\bsnm{Peng}, \binits{B.}},
\bauthor{\bsnm{Rashid}, \binits{T.}},
\bauthor{\bsnm{Witt}, \binits{C.}},
\bauthor{\bsnm{Kamienny}, \binits{P.-A.}},
\bauthor{\bsnm{Torr}, \binits{P.}},
\bauthor{\bsnm{B{\"o}hmer}, \binits{W.}},
\bauthor{\bsnm{Whiteson}, \binits{S.}}:
\bctitle{Facmac: Factored multi-agent centralised policy gradients}.
In: \bbtitle{Advances in Neural Information Processing Systems},
pp. \bfpage{12208}--\blpage{12221}
(\byear{2021})
\end{bchapter}
\endbibitem

%%% 109
\bibitem[\protect\citeauthoryear{Powell et~al.}{2026}]{powell2026zscshaping}
\begin{barticle}
\bauthor{\bsnm{Powell}, \binits{K.}},
\bauthor{\bsnm{Yu}, \binits{P.}},
\bauthor{\bsnm{Tokekar}, \binits{P.}}:
\batitle{Zero shot coordination for sparse reward tasks with diverse reward
  shapings}.
\bjtitle{arXiv preprint arXiv:2604.25076}
(\byear{2026})
\doiurl{10.48550/arXiv.2604.25076}
\end{barticle}
\endbibitem

%%% 110
\bibitem[\protect\citeauthoryear{Qin et~al.}{2024}]{qin2024multi}
\begin{botherref}
\oauthor{\bsnm{Qin}, \binits{R.}},
\oauthor{\bsnm{Chen}, \binits{F.}},
\oauthor{\bsnm{Wang}, \binits{T.}},
\oauthor{\bsnm{Yuan}, \binits{L.}},
\oauthor{\bsnm{Wu}, \binits{X.}},
\oauthor{\bsnm{Kang}, \binits{Y.}},
\oauthor{\bsnm{Zhang}, \binits{Z.}},
\oauthor{\bsnm{Zhang}, \binits{C.}},
\oauthor{\bsnm{Yu}, \binits{Y.}}:
Multi-agent policy transfer via task relationship modeling.
Science China Information Sciences
(2024)
\end{botherref}
\endbibitem

%%% 111
\bibitem[\protect\citeauthoryear{Qu et~al.}{2020}]{qu2020scalable}
\begin{bchapter}
\bauthor{\bsnm{Qu}, \binits{G.}},
\bauthor{\bsnm{Wierman}, \binits{A.}},
\bauthor{\bsnm{Li}, \binits{N.}}:
\bctitle{Scalable reinforcement learning of localized policies for multi-agent
  networked systems}.
In: \bbtitle{Learning for Dynamics and Control},
pp. \bfpage{256}--\blpage{266}
(\byear{2020}).
\bcomment{PMLR}
\end{bchapter}
\endbibitem

%%% 112
\bibitem[\protect\citeauthoryear{Resnick et~al.}{2018}]{resnick2018pommerman}
\begin{botherref}
\oauthor{\bsnm{Resnick}, \binits{C.}},
\oauthor{\bsnm{Eldridge}, \binits{W.}},
\oauthor{\bsnm{Ha}, \binits{D.}},
\oauthor{\bsnm{Britz}, \binits{D.}},
\oauthor{\bsnm{Foerster}, \binits{J.}},
\oauthor{\bsnm{Togelius}, \binits{J.}},
\oauthor{\bsnm{Cho}, \binits{K.}},
\oauthor{\bsnm{Bruna}, \binits{J.}}:
Pommerman: A multi-agent playground.
arXiv preprint arXiv:1809.07124
(2018)
\end{botherref}
\endbibitem

%%% 113
\bibitem[\protect\citeauthoryear{Rashid et~al.}{2020}]{wqmix}
\begin{barticle}
\bauthor{\bsnm{Rashid}, \binits{T.}},
\bauthor{\bsnm{Farquhar}, \binits{G.}},
\bauthor{\bsnm{Peng}, \binits{B.}},
\bauthor{\bsnm{Whiteson}, \binits{S.}}:
\batitle{Weighted qmix: Expanding monotonic value function factorisation for
  deep multi-agent reinforcement learning}.
\bjtitle{Advances in neural information processing systems}
\bvolume{33},
\bfpage{10199}--\blpage{10210}
(\byear{2020})
\end{barticle}
\endbibitem

%%% 114
\bibitem[\protect\citeauthoryear{Rashid et~al.}{2018}]{qmix}
\begin{bchapter}
\bauthor{\bsnm{Rashid}, \binits{T.}},
\bauthor{\bsnm{Samvelyan}, \binits{M.}},
\bauthor{\bsnm{Schroeder}, \binits{C.}},
\bauthor{\bsnm{Farquhar}, \binits{G.}},
\bauthor{\bsnm{Foerster}, \binits{J.}},
\bauthor{\bsnm{Whiteson}, \binits{S.}}:
\bctitle{Qmix: Monotonic value function factorisation for deep multi-agent
  reinforcement learning}.
In: \bbtitle{Proceedings of the International Conference on Machine Learning},
pp. \bfpage{4295}--\blpage{4304}
(\byear{2018})
\end{bchapter}
\endbibitem

%%% 115
\bibitem[\protect\citeauthoryear{Sutton et~al.}{1998}]{sutton1998reinforcement}
\begin{bbook}
\bauthor{\bsnm{Sutton}, \binits{R.S.}},
\bauthor{\bsnm{Barto}, \binits{A.G.}}, \betal:
\bbtitle{Reinforcement Learning: An Introduction}
vol. \bseriesno{1}.
\bpublisher{MIT press Cambridge}, \blocation{???}
(\byear{1998})
\end{bbook}
\endbibitem

%%% 116
\bibitem[\protect\citeauthoryear{Suarez et~al.}{2023}]{suarez2023neural}
\begin{barticle}
\bauthor{\bsnm{Suarez}, \binits{J.}},
\bauthor{\bsnm{Bloomin}, \binits{D.}},
\bauthor{\bsnm{Choe}, \binits{K.W.}},
\bauthor{\bsnm{Li}, \binits{H.X.}},
\bauthor{\bsnm{Sullivan}, \binits{R.}},
\bauthor{\bsnm{Kanna}, \binits{N.}},
\bauthor{\bsnm{Scott}, \binits{D.}},
\bauthor{\bsnm{Shuman}, \binits{R.}},
\bauthor{\bsnm{Bradley}, \binits{H.}},
\bauthor{\bsnm{Castricato}, \binits{L.}}, \betal:
\batitle{Neural mmo 2.0: A massively multi-task addition to massively
  multi-agent learning}.
\bjtitle{Advances in Neural Information Processing Systems}
\bvolume{36},
\bfpage{50094}--\blpage{50104}
(\byear{2023})
\end{barticle}
\endbibitem

%%% 117
\bibitem[\protect\citeauthoryear{Schafer et~al.}{2023}]{schafer2023learning}
\begin{bchapter}
\bauthor{\bsnm{Schafer}, \binits{L.}},
\bauthor{\bsnm{Christianos}, \binits{F.}},
\bauthor{\bsnm{Storkey}, \binits{A.}},
\bauthor{\bsnm{Albrecht}, \binits{S.}}:
\bctitle{Learning task embeddings for teamwork adaptation in multi-agent
  reinforcement learning}.
In: \bbtitle{NeurIPS 2023 Workshop on Generalization in Planning}
(\byear{2023})
\end{bchapter}
\endbibitem

%%% 118
\bibitem[\protect\citeauthoryear{Sukhbaatar
  et~al.}{2016}]{sukhbaatar2016learning}
\begin{botherref}
\oauthor{\bsnm{Sukhbaatar}, \binits{S.}},
\oauthor{\bsnm{Fergus}, \binits{R.}}, et al.:
Learning multiagent communication with backpropagation.
Advances in neural information processing systems
\textbf{29}
(2016)
\end{botherref}
\endbibitem

%%% 119
\bibitem[\protect\citeauthoryear{Son et~al.}{2019}]{son2019qtran}
\begin{bchapter}
\bauthor{\bsnm{Son}, \binits{K.}},
\bauthor{\bsnm{Kim}, \binits{D.}},
\bauthor{\bsnm{Kang}, \binits{W.J.}},
\bauthor{\bsnm{Hostallero}, \binits{D.E.}},
\bauthor{\bsnm{Yi}, \binits{Y.}}:
\bctitle{Qtran: Learning to factorize with transformation for cooperative
  multi-agent reinforcement learning}.
In: \bbtitle{International Conference on Machine Learning},
pp. \bfpage{5887}--\blpage{5896}
(\byear{2019}).
\bcomment{PMLR}
\end{bchapter}
\endbibitem

%%% 120
\bibitem[\protect\citeauthoryear{Sunehag et~al.}{2018}]{vdn}
\begin{bchapter}
\bauthor{\bsnm{Sunehag}, \binits{P.}},
\bauthor{\bsnm{Lever}, \binits{G.}},
\bauthor{\bsnm{Gruslys}, \binits{A.}},
\bauthor{\bsnm{Czarnecki}, \binits{W.M.}},
\bauthor{\bsnm{Zambaldi}, \binits{V.F.}},
\bauthor{\bsnm{Jaderberg}, \binits{M.}},
\bauthor{\bsnm{Lanctot}, \binits{M.}},
\bauthor{\bsnm{Sonnerat}, \binits{N.}},
\bauthor{\bsnm{Leibo}, \binits{J.Z.}},
\bauthor{\bsnm{Tuyls}, \binits{K.}},
\bauthor{\bsnm{Graepel}, \binits{T.}}:
\bctitle{Value-decomposition networks for cooperative multi-agent learning
  based on team reward}.
In: \bbtitle{Proceedings of the International Conference on Autonomous Agents
  and MultiAgent Systems},
pp. \bfpage{2085}--\blpage{2087}
(\byear{2018})
\end{bchapter}
\endbibitem

%%% 121
\bibitem[\protect\citeauthoryear{Shao et~al.}{2023}]{shao2023counterfactual}
\begin{barticle}
\bauthor{\bsnm{Shao}, \binits{J.}},
\bauthor{\bsnm{Qu}, \binits{Y.}},
\bauthor{\bsnm{Chen}, \binits{C.}},
\bauthor{\bsnm{Zhang}, \binits{H.}},
\bauthor{\bsnm{Ji}, \binits{X.}}:
\batitle{Counterfactual conservative q learning for offline multi-agent
  reinforcement learning}.
\bjtitle{Advances in Neural Information Processing Systems}
\bvolume{36},
\bfpage{77290}--\blpage{77312}
(\byear{2023})
\end{barticle}
\endbibitem

%%% 122
\bibitem[\protect\citeauthoryear{Samvelyan
  et~al.}{2019}]{samvelyan2019starcraft}
\begin{botherref}
\oauthor{\bsnm{Samvelyan}, \binits{M.}},
\oauthor{\bsnm{Rashid}, \binits{T.}},
\oauthor{\bsnm{De~Witt}, \binits{C.S.}},
\oauthor{\bsnm{Farquhar}, \binits{G.}},
\oauthor{\bsnm{Nardelli}, \binits{N.}},
\oauthor{\bsnm{Rudner}, \binits{T.G.}},
\oauthor{\bsnm{Hung}, \binits{C.-M.}},
\oauthor{\bsnm{Torr}, \binits{P.H.}},
\oauthor{\bsnm{Foerster}, \binits{J.}},
\oauthor{\bsnm{Whiteson}, \binits{S.}}:
The starcraft multi-agent challenge.
arXiv preprint arXiv:1902.04043
(2019)
\end{botherref}
\endbibitem

%%% 123
\bibitem[\protect\citeauthoryear{Stephenson and
  Schaub}{2024}]{stephenson2024bsk}
\begin{bchapter}
\bauthor{\bsnm{Stephenson}, \binits{M.A.}},
\bauthor{\bsnm{Schaub}, \binits{H.}}:
\bctitle{Bsk-rl: Modular, high-fidelity reinforcement learning environments for
  spacecraft tasking}.
In: \bbtitle{75th International Astronautical Congress, Milan, Italy, IAF}
(\byear{2024})
\end{bchapter}
\endbibitem

%%% 124
\bibitem[\protect\citeauthoryear{Stern et~al.}{2019}]{stern2019multi}
\begin{bchapter}
\bauthor{\bsnm{Stern}, \binits{R.}},
\bauthor{\bsnm{Sturtevant}, \binits{N.}},
\bauthor{\bsnm{Felner}, \binits{A.}},
\bauthor{\bsnm{Koenig}, \binits{S.}},
\bauthor{\bsnm{Ma}, \binits{H.}},
\bauthor{\bsnm{Walker}, \binits{T.}},
\bauthor{\bsnm{Li}, \binits{J.}},
\bauthor{\bsnm{Atzmon}, \binits{D.}},
\bauthor{\bsnm{Cohen}, \binits{L.}},
\bauthor{\bsnm{Kumar}, \binits{T.}}, \betal:
\bctitle{Multi-agent pathfinding: Definitions, variants, and benchmarks}.
In: \bbtitle{Proceedings of the International Symposium on Combinatorial
  Search}
(\byear{2019})
\end{bchapter}
\endbibitem

%%% 125
\bibitem[\protect\citeauthoryear{Shaik
  et~al.}{2026}]{shaik2026constrainedcommunication}
\begin{barticle}
\bauthor{\bsnm{Shaik}, \binits{S.}},
\bauthor{\bsnm{Smereka}, \binits{J.M.}},
\bauthor{\bsnm{Wang}, \binits{Y.}}:
\batitle{Multi-agent deep reinforcement learning under constrained
  communications}.
\bjtitle{arXiv preprint arXiv:2601.17069}
(\byear{2026})
\doiurl{10.48550/arXiv.2601.17069}
\end{barticle}
\endbibitem

%%% 126
\bibitem[\protect\citeauthoryear{Steinberger}{2019}]{Poker}
\begin{botherref}
\oauthor{\bsnm{Steinberger}, \binits{E.}}:
PokerRL.
GitHub
(2019)
\end{botherref}
\endbibitem

%%% 127
\bibitem[\protect\citeauthoryear{Sheng et~al.}{2022}]{sheng2022learning}
\begin{barticle}
\bauthor{\bsnm{Sheng}, \binits{J.}},
\bauthor{\bsnm{Wang}, \binits{X.}},
\bauthor{\bsnm{Jin}, \binits{B.}},
\bauthor{\bsnm{Yan}, \binits{J.}},
\bauthor{\bsnm{Li}, \binits{W.}},
\bauthor{\bsnm{Chang}, \binits{T.-H.}},
\bauthor{\bsnm{Wang}, \binits{J.}},
\bauthor{\bsnm{Zha}, \binits{H.}}:
\batitle{Learning structured communication for multi-agent reinforcement
  learning}.
\bjtitle{Autonomous Agents and Multi-Agent Systems}
\bvolume{36}(\bissue{2}),
\bfpage{50}
(\byear{2022})
\end{barticle}
\endbibitem

%%% 128
\bibitem[\protect\citeauthoryear{Sun et~al.}{2025}]{sun2025collab}
\begin{botherref}
\oauthor{\bsnm{Sun}, \binits{H.}},
\oauthor{\bsnm{Zhang}, \binits{S.}},
\oauthor{\bsnm{Niu}, \binits{L.}},
\oauthor{\bsnm{Ren}, \binits{L.}},
\oauthor{\bsnm{Xu}, \binits{H.}},
\oauthor{\bsnm{Fu}, \binits{H.}},
\oauthor{\bsnm{Zhao}, \binits{F.}},
\oauthor{\bsnm{Yuan}, \binits{C.}},
\oauthor{\bsnm{Wang}, \binits{X.}}:
Collab-overcooked: Benchmarking and evaluating large language models as
  collaborative agents.
arXiv preprint arXiv:2502.20073
(2025)
\end{botherref}
\endbibitem

%%% 129
\bibitem[\protect\citeauthoryear{Song et~al.}{2022}]{song2022hybrid}
\begin{botherref}
\oauthor{\bsnm{Song}, \binits{Y.}},
\oauthor{\bsnm{Zhou}, \binits{Y.}},
\oauthor{\bsnm{Sekhari}, \binits{A.}},
\oauthor{\bsnm{Bagnell}, \binits{J.A.}},
\oauthor{\bsnm{Krishnamurthy}, \binits{A.}},
\oauthor{\bsnm{Sun}, \binits{W.}}:
Hybrid rl: Using both offline and online data can make rl efficient.
arXiv preprint arXiv:2210.06718
(2022)
\end{botherref}
\endbibitem

%%% 130
\bibitem[\protect\citeauthoryear{Tan}{1993}]{tan1993multi}
\begin{bchapter}
\bauthor{\bsnm{Tan}, \binits{M.}}:
\bctitle{Multi-agent reinforcement learning: Independent vs. cooperative
  agents}.
In: \bbtitle{Proceedings of the International Conference on Machine Learning},
pp. \bfpage{330}--\blpage{337}
(\byear{1993})
\end{bchapter}
\endbibitem

%%% 131
\bibitem[\protect\citeauthoryear{Tian et~al.}{2023}]{tian2023decompose}
\begin{botherref}
\oauthor{\bsnm{Tian}, \binits{Z.}},
\oauthor{\bsnm{Chen}, \binits{R.}},
\oauthor{\bsnm{Hu}, \binits{X.}},
\oauthor{\bsnm{Li}, \binits{L.}},
\oauthor{\bsnm{Zhang}, \binits{R.}},
\oauthor{\bsnm{Wu}, \binits{F.}},
\oauthor{\bsnm{Peng}, \binits{S.}},
\oauthor{\bsnm{Guo}, \binits{J.}},
\oauthor{\bsnm{Du}, \binits{Z.}},
\oauthor{\bsnm{Guo}, \binits{Q.}}, et al.:
Decompose a task into generalizable subtasks in multi-agent reinforcement
  learning.
Advances in Neural Information Processing Systems
(2023)
\end{botherref}
\endbibitem

%%% 132
\bibitem[\protect\citeauthoryear{Torbati et~al.}{2023}]{torbati2023marbler}
\begin{bchapter}
\bauthor{\bsnm{Torbati}, \binits{R.J.}},
\bauthor{\bsnm{Lohiya}, \binits{S.}},
\bauthor{\bsnm{Singh}, \binits{S.}},
\bauthor{\bsnm{Nigam}, \binits{M.S.}},
\bauthor{\bsnm{Ravichandar}, \binits{H.}}:
\bctitle{Marbler: An open platform for standardized evaluation of multi-robot
  reinforcement learning algorithms}.
In: \bbtitle{2023 International Symposium on Multi-Robot and Multi-Agent
  Systems (MRS)},
pp. \bfpage{57}--\blpage{63}
(\byear{2023}).
\bcomment{IEEE}
\end{bchapter}
\endbibitem

%%% 133
\bibitem[\protect\citeauthoryear{Tomilin et~al.}{2026}]{tomilin2026meal}
\begin{barticle}
\bauthor{\bsnm{Tomilin}, \binits{T.}},
\bauthor{\bsnm{Boogaard}, \binits{L.}},
\bauthor{\bsnm{Garcin}, \binits{S.}},
\bauthor{\bsnm{Ruhdorfer}, \binits{C.}},
\bauthor{\bsnm{Grooten}, \binits{B.}},
\bauthor{\bsnm{Kusters}, \binits{F.}},
\bauthor{\bsnm{Du}, \binits{Y.}},
\bauthor{\bsnm{Bulling}, \binits{A.}},
\bauthor{\bsnm{Pechenizkiy}, \binits{M.}},
\bauthor{\bsnm{Fang}, \binits{M.}}:
\batitle{{MEAL}: A benchmark for continual multi-agent reinforcement learning}.
\bjtitle{arXiv preprint arXiv:2506.14990}
(\byear{2026})
\doiurl{10.48550/arXiv.2506.14990}
\end{barticle}
\endbibitem

%%% 134
\bibitem[\protect\citeauthoryear{Van Den~Oord et~al.}{2017}]{van2017neural}
\begin{botherref}
\oauthor{\bsnm{Van Den~Oord}, \binits{A.}},
\oauthor{\bsnm{Vinyals}, \binits{O.}}, et al.:
Neural discrete representation learning.
Advances in neural information processing systems
\textbf{30}
(2017)
\end{botherref}
\endbibitem

%%% 135
\bibitem[\protect\citeauthoryear{Willemsen et~al.}{2021}]{willemsen2021mambpo}
\begin{bchapter}
\bauthor{\bsnm{Willemsen}, \binits{D.}},
\bauthor{\bsnm{Coppola}, \binits{M.}},
\bauthor{\bsnm{Croon}, \binits{G.C.}}:
\bctitle{Mambpo: Sample-efficient multi-robot reinforcement learning using
  learned world models}.
In: \bbtitle{2021 IEEE/RSJ International Conference on Intelligent Robots and
  Systems (IROS)},
pp. \bfpage{5635}--\blpage{5640}
(\byear{2021}).
\bcomment{IEEE}
\end{bchapter}
\endbibitem

%%% 136
\bibitem[\protect\citeauthoryear{Wang et~al.}{2021}]{wangrode}
\begin{bchapter}
\bauthor{\bsnm{Wang}, \binits{T.}},
\bauthor{\bsnm{Gupta}, \binits{T.}},
\bauthor{\bsnm{Mahajan}, \binits{A.}},
\bauthor{\bsnm{Peng}, \binits{B.}},
\bauthor{\bsnm{Whiteson}, \binits{S.}},
\bauthor{\bsnm{Zhang}, \binits{C.}}:
\bctitle{Rode: Learning roles to decompose multi-agent tasks}.
In: \bbtitle{International Conference on Learning Representations}
(\byear{2021})
\end{bchapter}
\endbibitem

%%% 137
\bibitem[\protect\citeauthoryear{Wu et~al.}{2024}]{wu2024portal}
\begin{bchapter}
\bauthor{\bsnm{Wu}, \binits{J.}},
\bauthor{\bsnm{Hao}, \binits{J.}},
\bauthor{\bsnm{Yang}, \binits{T.}},
\bauthor{\bsnm{Hao}, \binits{X.}},
\bauthor{\bsnm{Zheng}, \binits{Y.}},
\bauthor{\bsnm{Wang}, \binits{W.}},
\bauthor{\bsnm{Taylor}, \binits{M.E.}}:
\bctitle{Portal: Automatic curricula generation for multiagent reinforcement
  learning}.
In: \bbtitle{Proceedings of the AAAI Conference on Artificial Intelligence},
vol. \bseriesno{38},
pp. \bfpage{15934}--\blpage{15942}
(\byear{2024})
\end{bchapter}
\endbibitem

%%% 138
\bibitem[\protect\citeauthoryear{Wang et~al.}{2021}]{qplex}
\begin{bchapter}
\bauthor{\bsnm{Wang}, \binits{J.}},
\bauthor{\bsnm{Ren}, \binits{Z.}},
\bauthor{\bsnm{Liu}, \binits{T.}},
\bauthor{\bsnm{Yu}, \binits{Y.}},
\bauthor{\bsnm{Zhang}, \binits{C.}}:
\bctitle{{\{}QPLEX{\}}: Duplex dueling multi-agent q-learning}.
In: \bbtitle{International Conference on Learning Representations}
(\byear{2021}).
\burl{https://openreview.net/forum?id=Rcmk0xxIQV}
\end{bchapter}
\endbibitem

%%% 139
\bibitem[\protect\citeauthoryear{Wang et~al.}{2020}]{wang2020learning}
\begin{bchapter}
\bauthor{\bsnm{Wang}, \binits{T.}},
\bauthor{\bsnm{Wang}, \binits{J.}},
\bauthor{\bsnm{Zheng}, \binits{C.}},
\bauthor{\bsnm{Zhang}, \binits{C.}}:
\bctitle{Learning nearly decomposable value functions via communication
  minimization}.
In: \bbtitle{International Conference on Learning Representations}
(\byear{2020})
\end{bchapter}
\endbibitem

%%% 140
\bibitem[\protect\citeauthoryear{Wang et~al.}{2021}]{wang2021multi}
\begin{barticle}
\bauthor{\bsnm{Wang}, \binits{J.}},
\bauthor{\bsnm{Xu}, \binits{W.}},
\bauthor{\bsnm{Gu}, \binits{Y.}},
\bauthor{\bsnm{Song}, \binits{W.}},
\bauthor{\bsnm{Green}, \binits{T.C.}}:
\batitle{Multi-agent reinforcement learning for active voltage control on power
  distribution networks}.
\bjtitle{Advances in Neural Information Processing Systems}
\bvolume{34},
\bfpage{3271}--\blpage{3284}
(\byear{2021})
\end{barticle}
\endbibitem

%%% 141
\bibitem[\protect\citeauthoryear{Wang et~al.}{2023}]{wang2023offline}
\begin{barticle}
\bauthor{\bsnm{Wang}, \binits{X.}},
\bauthor{\bsnm{Xu}, \binits{H.}},
\bauthor{\bsnm{Zheng}, \binits{Y.}},
\bauthor{\bsnm{Zhan}, \binits{X.}}:
\batitle{Offline multi-agent reinforcement learning with implicit
  global-to-local value regularization}.
\bjtitle{Advances in Neural Information Processing Systems}
\bvolume{36},
\bfpage{52413}--\blpage{52429}
(\byear{2023})
\end{barticle}
\endbibitem

%%% 142
\bibitem[\protect\citeauthoryear{Wang et~al.}{2020}]{wang2020few}
\begin{bchapter}
\bauthor{\bsnm{Wang}, \binits{W.}},
\bauthor{\bsnm{Yang}, \binits{T.}},
\bauthor{\bsnm{Liu}, \binits{Y.}},
\bauthor{\bsnm{Hao}, \binits{J.}},
\bauthor{\bsnm{Hao}, \binits{X.}},
\bauthor{\bsnm{Hu}, \binits{Y.}},
\bauthor{\bsnm{Chen}, \binits{Y.}},
\bauthor{\bsnm{Fan}, \binits{C.}},
\bauthor{\bsnm{Gao}, \binits{Y.}}:
\bctitle{From few to more: Large-scale dynamic multiagent curriculum learning}.
In: \bbtitle{Proceedings of the AAAI Conference on Artificial Intelligence},
vol. \bseriesno{34},
pp. \bfpage{7293}--\blpage{7300}
(\byear{2020})
\end{bchapter}
\endbibitem

%%% 143
\bibitem[\protect\citeauthoryear{Wang et~al.}{2022}]{wang2022context}
\begin{bchapter}
\bauthor{\bsnm{Wang}, \binits{T.}},
\bauthor{\bsnm{Zeng}, \binits{L.}},
\bauthor{\bsnm{Dong}, \binits{W.}},
\bauthor{\bsnm{Yang}, \binits{Q.}},
\bauthor{\bsnm{Yu}, \binits{Y.}},
\bauthor{\bsnm{Zhang}, \binits{C.}}:
\bctitle{Context-aware sparse deep coordination graphs}.
In: \bbtitle{International Conference on Learning Representations}
(\byear{2022})
\end{bchapter}
\endbibitem

%%% 144
\bibitem[\protect\citeauthoryear{Wang et~al.}{2024}]{wang2024battleagentbench}
\begin{botherref}
\oauthor{\bsnm{Wang}, \binits{W.}},
\oauthor{\bsnm{Zhang}, \binits{D.}},
\oauthor{\bsnm{Feng}, \binits{T.}},
\oauthor{\bsnm{Wang}, \binits{B.}},
\oauthor{\bsnm{Tang}, \binits{J.}}:
Battleagentbench: A benchmark for evaluating cooperation and competition
  capabilities of language models in multi-agent systems.
arXiv preprint arXiv:2408.15971
(2024)
\end{botherref}
\endbibitem

%%% 145
\bibitem[\protect\citeauthoryear{Wang et~al.}{2022}]{wang2022shaq}
\begin{bchapter}
\bauthor{\bsnm{Wang}, \binits{J.}},
\bauthor{\bsnm{Zhang}, \binits{Y.}},
\bauthor{\bsnm{Gu}, \binits{Y.}},
\bauthor{\bsnm{Kim}, \binits{T.-K.}}:
\bctitle{Shaq: Incorporating shapley value theory into multi-agent q-learning}.
In: \bbtitle{Advances in Neural Information Processing Systems},
vol. \bseriesno{35},
pp. \bfpage{5941}--\blpage{5954}
(\byear{2022})
\end{bchapter}
\endbibitem

%%% 146
\bibitem[\protect\citeauthoryear{Wang et~al.}{2020}]{wang2020shapley}
\begin{bchapter}
\bauthor{\bsnm{Wang}, \binits{J.}},
\bauthor{\bsnm{Zhang}, \binits{Y.}},
\bauthor{\bsnm{Kim}, \binits{T.-K.}},
\bauthor{\bsnm{Gu}, \binits{Y.}}:
\bctitle{Shapley q-value: A local reward approach to solve global reward
  games}.
In: \bbtitle{Proceedings of the AAAI Conference on Artificial Intelligence},
vol. \bseriesno{34},
pp. \bfpage{7285}--\blpage{7292}
(\byear{2020})
\end{bchapter}
\endbibitem

%%% 147
\bibitem[\protect\citeauthoryear{Wang et~al.}{2023}]{wang2023towards}
\begin{botherref}
\oauthor{\bsnm{Wang}, \binits{R.}},
\oauthor{\bsnm{Zheng}, \binits{L.}},
\oauthor{\bsnm{Qiu}, \binits{W.}},
\oauthor{\bsnm{He}, \binits{B.}},
\oauthor{\bsnm{An}, \binits{B.}},
\oauthor{\bsnm{Rabinovich}, \binits{Z.}},
\oauthor{\bsnm{Hu}, \binits{Y.}},
\oauthor{\bsnm{Chen}, \binits{Y.}},
\oauthor{\bsnm{Lv}, \binits{T.}},
\oauthor{\bsnm{Fan}, \binits{C.}}:
Towards skilled population curriculum for multi-agent reinforcement learning.
arXiv preprint arXiv:2302.03429
(2023)
\end{botherref}
\endbibitem

%%% 148
\bibitem[\protect\citeauthoryear{Xia et~al.}{2021}]{xia2021multi}
\begin{barticle}
\bauthor{\bsnm{Xia}, \binits{Z.}},
\bauthor{\bsnm{Du}, \binits{J.}},
\bauthor{\bsnm{Wang}, \binits{J.}},
\bauthor{\bsnm{Jiang}, \binits{C.}},
\bauthor{\bsnm{Ren}, \binits{Y.}},
\bauthor{\bsnm{Li}, \binits{G.}},
\bauthor{\bsnm{Han}, \binits{Z.}}:
\batitle{Multi-agent reinforcement learning aided intelligent uav swarm for
  target tracking}.
\bjtitle{IEEE Transactions on Vehicular Technology}
\bvolume{71}(\bissue{1}),
\bfpage{931}--\blpage{945}
(\byear{2021})
\end{barticle}
\endbibitem

%%% 149
\bibitem[\protect\citeauthoryear{Xu et~al.}{2023}]{xu2023magic}
\begin{botherref}
\oauthor{\bsnm{Xu}, \binits{L.}},
\oauthor{\bsnm{Hu}, \binits{Z.}},
\oauthor{\bsnm{Zhou}, \binits{D.}},
\oauthor{\bsnm{Ren}, \binits{H.}},
\oauthor{\bsnm{Dong}, \binits{Z.}},
\oauthor{\bsnm{Keutzer}, \binits{K.}},
\oauthor{\bsnm{Ng}, \binits{S.K.}},
\oauthor{\bsnm{Feng}, \binits{J.}}:
Magic: Investigation of large language model powered multi-agent in cognition,
  adaptability, rationality and collaboration.
arXiv preprint arXiv:2311.08562
(2023)
\end{botherref}
\endbibitem

%%% 150
\bibitem[\protect\citeauthoryear{Xie et~al.}{2021}]{xie2021policy}
\begin{barticle}
\bauthor{\bsnm{Xie}, \binits{T.}},
\bauthor{\bsnm{Jiang}, \binits{N.}},
\bauthor{\bsnm{Wang}, \binits{H.}},
\bauthor{\bsnm{Xiong}, \binits{C.}},
\bauthor{\bsnm{Bai}, \binits{Y.}}:
\batitle{Policy finetuning: Bridging sample-efficient offline and online
  reinforcement learning}.
\bjtitle{Advances in neural information processing systems}
\bvolume{34},
\bfpage{27395}--\blpage{27407}
(\byear{2021})
\end{barticle}
\endbibitem

%%% 151
\bibitem[\protect\citeauthoryear{Xue et~al.}{2022}]{xue2022multi}
\begin{barticle}
\bauthor{\bsnm{Xue}, \binits{K.}},
\bauthor{\bsnm{Xu}, \binits{J.}},
\bauthor{\bsnm{Yuan}, \binits{L.}},
\bauthor{\bsnm{Li}, \binits{M.}},
\bauthor{\bsnm{Qian}, \binits{C.}},
\bauthor{\bsnm{Zhang}, \binits{Z.}},
\bauthor{\bsnm{Yu}, \binits{Y.}}:
\batitle{Multi-agent dynamic algorithm configuration}.
\bjtitle{Advances in Neural Information Processing Systems}
\bvolume{35},
\bfpage{20147}--\blpage{20161}
(\byear{2022})
\end{barticle}
\endbibitem

%%% 152
\bibitem[\protect\citeauthoryear{Xu et~al.}{2022}]{xu2022mingling}
\begin{barticle}
\bauthor{\bsnm{Xu}, \binits{Z.}},
\bauthor{\bsnm{Zhang}, \binits{B.}},
\bauthor{\bsnm{Zhan}, \binits{Y.}},
\bauthor{\bsnm{Baiia}, \binits{Y.}},
\bauthor{\bsnm{Fan}, \binits{G.}}, \betal:
\batitle{Mingling foresight with imagination: Model-based cooperative
  multi-agent reinforcement learning}.
\bjtitle{Advances in Neural Information Processing Systems}
\bvolume{35},
\bfpage{11327}--\blpage{11340}
(\byear{2022})
\end{barticle}
\endbibitem

%%% 153
\bibitem[\protect\citeauthoryear{Yao et~al.}{2026}]{yao2026langmarl}
\begin{barticle}
\bauthor{\bsnm{Yao}, \binits{H.}},
\bauthor{\bsnm{Da}, \binits{L.}},
\bauthor{\bsnm{Liu}, \binits{X.}},
\bauthor{\bsnm{Fleming}, \binits{C.}},
\bauthor{\bsnm{Chen}, \binits{T.}},
\bauthor{\bsnm{Wei}, \binits{H.}}:
\batitle{{LangMARL}: Natural language multi-agent reinforcement learning}.
\bjtitle{arXiv preprint arXiv:2604.00722}
(\byear{2026})
\doiurl{10.48550/arXiv.2604.00722}
\end{barticle}
\endbibitem

%%% 154
\bibitem[\protect\citeauthoryear{Yu et~al.}{2022}]{yu2022model}
\begin{barticle}
\bauthor{\bsnm{Yu}, \binits{X.}},
\bauthor{\bsnm{Jiang}, \binits{J.}},
\bauthor{\bsnm{Zhang}, \binits{W.}},
\bauthor{\bsnm{Jiang}, \binits{H.}},
\bauthor{\bsnm{Lu}, \binits{Z.}}:
\batitle{Model-based opponent modeling}.
\bjtitle{Advances in Neural Information Processing Systems}
\bvolume{35},
\bfpage{28208}--\blpage{28221}
(\byear{2022})
\end{barticle}
\endbibitem

%%% 155
\bibitem[\protect\citeauthoryear{Yang et~al.}{2023}]{yang2023versatile}
\begin{botherref}
\oauthor{\bsnm{Yang}, \binits{X.}},
\oauthor{\bsnm{Liu}, \binits{Z.}},
\oauthor{\bsnm{Jiang}, \binits{W.}},
\oauthor{\bsnm{Zhang}, \binits{C.}},
\oauthor{\bsnm{Zhao}, \binits{L.}},
\oauthor{\bsnm{Song}, \binits{L.}},
\oauthor{\bsnm{Bian}, \binits{J.}}:
A versatile multi-agent reinforcement learning benchmark for inventory
  management.
arXiv preprint arXiv:2306.07542
(2023)
\end{botherref}
\endbibitem

%%% 156
\bibitem[\protect\citeauthoryear{Yang et~al.}{2018}]{yang2018mean}
\begin{bchapter}
\bauthor{\bsnm{Yang}, \binits{Y.}},
\bauthor{\bsnm{Luo}, \binits{R.}},
\bauthor{\bsnm{Li}, \binits{M.}},
\bauthor{\bsnm{Zhou}, \binits{M.}},
\bauthor{\bsnm{Zhang}, \binits{W.}},
\bauthor{\bsnm{Wang}, \binits{J.}}:
\bctitle{Mean field multi-agent reinforcement learning}.
In: \bbtitle{International Conference on Machine Learning},
pp. \bfpage{5571}--\blpage{5580}
(\byear{2018}).
\bcomment{PMLR}
\end{bchapter}
\endbibitem

%%% 157
\bibitem[\protect\citeauthoryear{Yuan et~al.}{2024}]{yuan2024multiagent}
\begin{botherref}
\oauthor{\bsnm{Yuan}, \binits{L.}},
\oauthor{\bsnm{Li}, \binits{L.}},
\oauthor{\bsnm{Zhang}, \binits{Z.}},
\oauthor{\bsnm{Zhang}, \binits{F.}},
\oauthor{\bsnm{Guan}, \binits{C.}},
\oauthor{\bsnm{Yu}, \binits{Y.}}:
Multiagent continual coordination via progressive task contextualization.
IEEE Transactions on Neural Networks and Learning Systems
(2024)
\end{botherref}
\endbibitem

%%% 158
\bibitem[\protect\citeauthoryear{Yang et~al.}{2021}]{yang2021believe}
\begin{barticle}
\bauthor{\bsnm{Yang}, \binits{Y.}},
\bauthor{\bsnm{Ma}, \binits{X.}},
\bauthor{\bsnm{Li}, \binits{C.}},
\bauthor{\bsnm{Zheng}, \binits{Z.}},
\bauthor{\bsnm{Zhang}, \binits{Q.}},
\bauthor{\bsnm{Huang}, \binits{G.}},
\bauthor{\bsnm{Yang}, \binits{J.}},
\bauthor{\bsnm{Zhao}, \binits{Q.}}:
\batitle{Believe what you see: Implicit constraint approach for offline
  multi-agent reinforcement learning}.
\bjtitle{Advances in Neural Information Processing Systems}
\bvolume{34},
\bfpage{10299}--\blpage{10312}
(\byear{2021})
\end{barticle}
\endbibitem

%%% 159
\bibitem[\protect\citeauthoryear{Yu et~al.}{2022}]{mappo}
\begin{bchapter}
\bauthor{\bsnm{Yu}, \binits{C.}},
\bauthor{\bsnm{Velu}, \binits{A.}},
\bauthor{\bsnm{Vinitsky}, \binits{E.}},
\bauthor{\bsnm{Gao}, \binits{J.}},
\bauthor{\bsnm{Wang}, \binits{Y.}},
\bauthor{\bsnm{Bayen}, \binits{A.}},
\bauthor{\bsnm{Wu}, \binits{Y.}}:
\bctitle{The surprising effectiveness of {PPO} in cooperative multi-agent
  games}.
In: \bbtitle{Advances in Neural Information Processing Systems},
pp. \bfpage{24611}--\blpage{24624}
(\byear{2022})
\end{bchapter}
\endbibitem

%%% 160
\bibitem[\protect\citeauthoryear{Yu et~al.}{2024}]{yu2024relation}
\begin{botherref}
\oauthor{\bsnm{Yu}, \binits{Y.}},
\oauthor{\bsnm{Yang}, \binits{L.}},
\oauthor{\bsnm{Guo}, \binits{Z.}},
\oauthor{\bsnm{Ren}, \binits{Y.}},
\oauthor{\bsnm{Yin}, \binits{Q.}},
\oauthor{\bsnm{Zhang}, \binits{J.}},
\oauthor{\bsnm{Huang}, \binits{K.}}:
Relation-aware learning for multi-task multi-agent cooperative games.
IEEE Transactions on Games
(2024)
\end{botherref}
\endbibitem

%%% 161
\bibitem[\protect\citeauthoryear{Yu et~al.}{2023}]{yu2023prioritized}
\begin{bchapter}
\bauthor{\bsnm{Yu}, \binits{Y.}},
\bauthor{\bsnm{Yin}, \binits{Q.}},
\bauthor{\bsnm{Zhang}, \binits{J.}},
\bauthor{\bsnm{Huang}, \binits{K.}}:
\bctitle{Prioritized tasks mining for multi-task cooperative multi-agent
  reinforcement learning}.
In: \bbtitle{Proceedings of the 2023 International Conference on Autonomous
  Agents and Multiagent Systems}
(\byear{2023})
\end{bchapter}
\endbibitem

%%% 162
\bibitem[\protect\citeauthoryear{Ying et~al.}{2023}]{ying2023scalable}
\begin{barticle}
\bauthor{\bsnm{Ying}, \binits{D.}},
\bauthor{\bsnm{Zhang}, \binits{Y.}},
\bauthor{\bsnm{Ding}, \binits{Y.}},
\bauthor{\bsnm{Koppel}, \binits{A.}},
\bauthor{\bsnm{Lavaei}, \binits{J.}}:
\batitle{Scalable primal-dual actor-critic method for safe multi-agent rl with
  general utilities}.
\bjtitle{Advances in Neural Information Processing Systems}
\bvolume{36},
\bfpage{36524}--\blpage{36539}
(\byear{2023})
\end{barticle}
\endbibitem

%%% 163
\bibitem[\protect\citeauthoryear{Yuan et~al.}{2023}]{yuan2023survey}
\begin{botherref}
\oauthor{\bsnm{Yuan}, \binits{L.}},
\oauthor{\bsnm{Zhang}, \binits{Z.}},
\oauthor{\bsnm{Li}, \binits{L.}},
\oauthor{\bsnm{Guan}, \binits{C.}},
\oauthor{\bsnm{Yu}, \binits{Y.}}:
A survey of progress on cooperative multi-agent reinforcement learning in open
  environment.
arXiv preprint arXiv:2312.01058
(2023)
\end{botherref}
\endbibitem

%%% 164
\bibitem[\protect\citeauthoryear{Zhang et~al.}{2019}]{zhang2019mamps}
\begin{botherref}
\oauthor{\bsnm{Zhang}, \binits{W.}},
\oauthor{\bsnm{Bastani}, \binits{O.}},
\oauthor{\bsnm{Kumar}, \binits{V.}}:
Mamps: Safe multi-agent reinforcement learning via model predictive shielding.
arXiv preprint arXiv:1910.12639
(2019)
\end{botherref}
\endbibitem

%%% 165
\bibitem[\protect\citeauthoryear{Zhang et~al.}{2022}]{zhang2022efficient}
\begin{botherref}
\oauthor{\bsnm{Zhang}, \binits{B.}},
\oauthor{\bsnm{Bai}, \binits{Y.}},
\oauthor{\bsnm{Xu}, \binits{Z.}},
\oauthor{\bsnm{Li}, \binits{D.}},
\oauthor{\bsnm{Fan}, \binits{G.}}:
Efficient cooperation strategy generation in multi-agent video games via
  hypergraph neural network.
arXiv:2203.03265
(2022)
\end{botherref}
\endbibitem

%%% 166
\bibitem[\protect\citeauthoryear{Zhu et~al.}{2025}]{zhu2025multiagentbench}
\begin{botherref}
\oauthor{\bsnm{Zhu}, \binits{K.}},
\oauthor{\bsnm{Du}, \binits{H.}},
\oauthor{\bsnm{Hong}, \binits{Z.}},
\oauthor{\bsnm{Yang}, \binits{X.}},
\oauthor{\bsnm{Guo}, \binits{S.}},
\oauthor{\bsnm{Wang}, \binits{Z.}},
\oauthor{\bsnm{Wang}, \binits{Z.}},
\oauthor{\bsnm{Qian}, \binits{C.}},
\oauthor{\bsnm{Tang}, \binits{X.}},
\oauthor{\bsnm{Ji}, \binits{H.}}, et al.:
Multiagentbench: Evaluating the collaboration and competition of llm agents.
arXiv preprint arXiv:2503.01935
(2025)
\end{botherref}
\endbibitem

%%% 167
\bibitem[\protect\citeauthoryear{Zhu et~al.}{2024}]{zhu2024survey}
\begin{barticle}
\bauthor{\bsnm{Zhu}, \binits{C.}},
\bauthor{\bsnm{Dastani}, \binits{M.}},
\bauthor{\bsnm{Wang}, \binits{S.}}:
\batitle{A survey of multi-agent deep reinforcement learning with
  communication}.
\bjtitle{Autonomous Agents and Multi-Agent Systems}
\bvolume{38}(\bissue{1}),
\bfpage{4}
(\byear{2024})
\end{barticle}
\endbibitem

%%% 168
\bibitem[\protect\citeauthoryear{Zhang et~al.}{2019}]{zhang2019cityflow}
\begin{bchapter}
\bauthor{\bsnm{Zhang}, \binits{H.}},
\bauthor{\bsnm{Feng}, \binits{S.}},
\bauthor{\bsnm{Liu}, \binits{C.}},
\bauthor{\bsnm{Ding}, \binits{Y.}},
\bauthor{\bsnm{Zhu}, \binits{Y.}},
\bauthor{\bsnm{Zhou}, \binits{Z.}},
\bauthor{\bsnm{Zhang}, \binits{W.}},
\bauthor{\bsnm{Yu}, \binits{Y.}},
\bauthor{\bsnm{Jin}, \binits{H.}},
\bauthor{\bsnm{Li}, \binits{Z.}}:
\bctitle{Cityflow: A multi-agent reinforcement learning environment for large
  scale city traffic scenario}.
In: \bbtitle{The World Wide Web Conference},
pp. \bfpage{3620}--\blpage{3624}
(\byear{2019})
\end{bchapter}
\endbibitem

%%% 169
\bibitem[\protect\citeauthoryear{Zhuang et~al.}{2025}]{zhuang2025pokerbench}
\begin{botherref}
\oauthor{\bsnm{Zhuang}, \binits{R.}},
\oauthor{\bsnm{Gupta}, \binits{A.}},
\oauthor{\bsnm{Yang}, \binits{R.}},
\oauthor{\bsnm{Rahane}, \binits{A.}},
\oauthor{\bsnm{Li}, \binits{Z.}},
\oauthor{\bsnm{Anumanchipalli}, \binits{G.}}:
Pokerbench: Training large language models to become professional poker
  players.
arXiv preprint arXiv:2501.08328
(2025)
\end{botherref}
\endbibitem

%%% 170
\bibitem[\protect\citeauthoryear{Zhu et~al.}{2024}]{zhu2024multi}
\begin{botherref}
\oauthor{\bsnm{Zhu}, \binits{Y.}},
\oauthor{\bsnm{Huang}, \binits{S.}},
\oauthor{\bsnm{Zuo}, \binits{B.}},
\oauthor{\bsnm{Zhao}, \binits{D.}},
\oauthor{\bsnm{Sun}, \binits{C.}}:
Multi-task multi-agent reinforcement learning with task-entity transformers and
  value decomposition training.
IEEE Transactions on Automation Science and Engineering
(2024)
\end{botherref}
\endbibitem

%%% 171
\bibitem[\protect\citeauthoryear{Zhong et~al.}{2024}]{harl}
\begin{barticle}
\bauthor{\bsnm{Zhong}, \binits{Y.}},
\bauthor{\bsnm{Kuba}, \binits{J.G.}},
\bauthor{\bsnm{Feng}, \binits{X.}},
\bauthor{\bsnm{Hu}, \binits{S.}},
\bauthor{\bsnm{Ji}, \binits{J.}},
\bauthor{\bsnm{Yang}, \binits{Y.}}:
\batitle{Heterogeneous-agent reinforcement learning}.
\bjtitle{Journal of Machine Learning Research}
\bvolume{25}(\bissue{32}),
\bfpage{1}--\blpage{67}
(\byear{2024})
\end{barticle}
\endbibitem

%%% 172
\bibitem[\protect\citeauthoryear{Zhang et~al.}{2021}]{zhang2021centralized}
\begin{botherref}
\oauthor{\bsnm{Zhang}, \binits{Q.}},
\oauthor{\bsnm{Lu}, \binits{C.}},
\oauthor{\bsnm{Garg}, \binits{A.}},
\oauthor{\bsnm{Foerster}, \binits{J.}}:
Centralized model and exploration policy for multi-agent rl.
arXiv preprint arXiv:2107.06434
(2021)
\end{botherref}
\endbibitem

%%% 173
\bibitem[\protect\citeauthoryear{Zhang et~al.}{2022}]{zhang2022automatic}
\begin{barticle}
\bauthor{\bsnm{Zhang}, \binits{T.}},
\bauthor{\bsnm{Liu}, \binits{Z.}},
\bauthor{\bsnm{Pu}, \binits{Z.}},
\bauthor{\bsnm{Yi}, \binits{J.}}:
\batitle{Automatic curriculum learning for large-scale cooperative multiagent
  systems}.
\bjtitle{IEEE Transactions on Emerging Topics in Computational Intelligence}
\bvolume{7}(\bissue{3}),
\bfpage{912}--\blpage{930}
(\byear{2022})
\end{barticle}
\endbibitem

%%% 174
\bibitem[\protect\citeauthoryear{Zhou et~al.}{2020}]{SMARTS}
\begin{botherref}
\oauthor{\bsnm{Zhou}, \binits{M.}},
\oauthor{\bsnm{Luo}, \binits{J.}},
\oauthor{\bsnm{Villella}, \binits{J.}},
\oauthor{\bsnm{Yang}, \binits{Y.}},
\oauthor{\bsnm{Rusu}, \binits{D.}},
\oauthor{\bsnm{Miao}, \binits{J.}},
\oauthor{\bsnm{Zhang}, \binits{W.}},
\oauthor{\bsnm{Alban}, \binits{M.}},
\oauthor{\bsnm{Fadakar}, \binits{I.}},
\oauthor{\bsnm{Chen}, \binits{Z.}},
\oauthor{\bsnm{Huang}, \binits{A.C.}},
\oauthor{\bsnm{Wen}, \binits{Y.}},
\oauthor{\bsnm{Hassanzadeh}, \binits{K.}},
\oauthor{\bsnm{Graves}, \binits{D.}},
\oauthor{\bsnm{Chen}, \binits{D.}},
\oauthor{\bsnm{Zhu}, \binits{Z.}},
\oauthor{\bsnm{Nguyen}, \binits{N.}},
\oauthor{\bsnm{Elsayed}, \binits{M.}},
\oauthor{\bsnm{Shao}, \binits{K.}},
\oauthor{\bsnm{Ahilan}, \binits{S.}},
\oauthor{\bsnm{Zhang}, \binits{B.}},
\oauthor{\bsnm{Wu}, \binits{J.}},
\oauthor{\bsnm{Fu}, \binits{Z.}},
\oauthor{\bsnm{Rezaee}, \binits{K.}},
\oauthor{\bsnm{Yadmellat}, \binits{P.}},
\oauthor{\bsnm{Rohani}, \binits{M.}},
\oauthor{\bsnm{Nieves}, \binits{N.P.}},
\oauthor{\bsnm{Ni}, \binits{Y.}},
\oauthor{\bsnm{Banijamali}, \binits{S.}},
\oauthor{\bsnm{Rivers}, \binits{A.C.}},
\oauthor{\bsnm{Tian}, \binits{Z.}},
\oauthor{\bsnm{Palenicek}, \binits{D.}},
\oauthor{\bsnm{Ammar}, \binits{H.}},
\oauthor{\bsnm{Zhang}, \binits{H.}},
\oauthor{\bsnm{Liu}, \binits{W.}},
\oauthor{\bsnm{Hao}, \binits{J.}},
\oauthor{\bsnm{Wang}, \binits{J.}}:
SMARTS: Scalable Multi-Agent Reinforcement Learning Training School for
  Autonomous Driving
(2020).
\url{https://arxiv.org/abs/2010.09776}
\end{botherref}
\endbibitem

%%% 175
\bibitem[\protect\citeauthoryear{Zhang et~al.}{2021}]{zhang2021model}
\begin{botherref}
\oauthor{\bsnm{Zhang}, \binits{W.}},
\oauthor{\bsnm{Wang}, \binits{X.}},
\oauthor{\bsnm{Shen}, \binits{J.}},
\oauthor{\bsnm{Zhou}, \binits{M.}}:
Model-based multi-agent policy optimization with adaptive opponent-wise
  rollouts.
arXiv preprint arXiv:2105.03363
(2021)
\end{botherref}
\endbibitem

%%% 176
\bibitem[\protect\citeauthoryear{Zhang et~al.}{2018}]{zhang2018networked}
\begin{bchapter}
\bauthor{\bsnm{Zhang}, \binits{K.}},
\bauthor{\bsnm{Yang}, \binits{Z.}},
\bauthor{\bsnm{Basar}, \binits{T.}}:
\bctitle{Networked multi-agent reinforcement learning in continuous spaces}.
In: \bbtitle{2018 IEEE Conference on Decision and Control (CDC)},
pp. \bfpage{2771}--\blpage{2776}
(\byear{2018}).
\bcomment{IEEE}
\end{bchapter}
\endbibitem

%%% 177
\bibitem[\protect\citeauthoryear{Zhang et~al.}{2021}]{zhang2021multi}
\begin{botherref}
\oauthor{\bsnm{Zhang}, \binits{K.}},
\oauthor{\bsnm{Yang}, \binits{Z.}},
\oauthor{\bsnm{Ba{\c{s}}ar}, \binits{T.}}:
Multi-agent reinforcement learning: A selective overview of theories and
  algorithms.
Handbook of reinforcement learning and control,
321--384
(2021)
\end{botherref}
\endbibitem

%%% 178
\bibitem[\protect\citeauthoryear{Zheng et~al.}{2018}]{zheng2018magent}
\begin{bchapter}
\bauthor{\bsnm{Zheng}, \binits{L.}},
\bauthor{\bsnm{Yang}, \binits{J.}},
\bauthor{\bsnm{Cai}, \binits{H.}},
\bauthor{\bsnm{Zhou}, \binits{M.}},
\bauthor{\bsnm{Zhang}, \binits{W.}},
\bauthor{\bsnm{Wang}, \binits{J.}},
\bauthor{\bsnm{Yu}, \binits{Y.}}:
\bctitle{Magent: A many-agent reinforcement learning platform for artificial
  collective intelligence}.
In: \bbtitle{Proceedings of the AAAI Conference on Artificial Intelligence},
vol. \bseriesno{32}
(\byear{2018})
\end{bchapter}
\endbibitem

%%% 179
\bibitem[\protect\citeauthoryear{Zhang et~al.}{2018}]{zhang2018fully}
\begin{bchapter}
\bauthor{\bsnm{Zhang}, \binits{K.}},
\bauthor{\bsnm{Yang}, \binits{Z.}},
\bauthor{\bsnm{Liu}, \binits{H.}},
\bauthor{\bsnm{Zhang}, \binits{T.}},
\bauthor{\bsnm{Basar}, \binits{T.}}:
\bctitle{Fully decentralized multi-agent reinforcement learning with networked
  agents}.
In: \bbtitle{International Conference on Machine Learning},
pp. \bfpage{5872}--\blpage{5881}
(\byear{2018}).
\bcomment{PMLR}
\end{bchapter}
\endbibitem

%%% 180
\bibitem[\protect\citeauthoryear{Zhou et~al.}{2023}]{zhou2023cooperative}
\begin{botherref}
\oauthor{\bsnm{Zhou}, \binits{T.}},
\oauthor{\bsnm{Zhang}, \binits{F.}},
\oauthor{\bsnm{Shao}, \binits{K.}},
\oauthor{\bsnm{Dai}, \binits{Z.}},
\oauthor{\bsnm{Li}, \binits{K.}},
\oauthor{\bsnm{Huang}, \binits{W.}},
\oauthor{\bsnm{Wang}, \binits{W.}},
\oauthor{\bsnm{Wang}, \binits{B.}},
\oauthor{\bsnm{Li}, \binits{D.}},
\oauthor{\bsnm{Liu}, \binits{W.}}, et al.:
Cooperative multi-agent transfer learning with coalition pattern decomposition.
IEEE Transactions on Games
(2023)
\end{botherref}
\endbibitem

%%% 181
\bibitem[\protect\citeauthoryear{Zhang et~al.}{2023}]{zhang2023gobigger}
\begin{bchapter}
\bauthor{\bsnm{Zhang}, \binits{M.}},
\bauthor{\bsnm{Zhang}, \binits{S.}},
\bauthor{\bsnm{Yang}, \binits{Z.}},
\bauthor{\bsnm{Chen}, \binits{L.}},
\bauthor{\bsnm{Zheng}, \binits{J.}},
\bauthor{\bsnm{Yang}, \binits{C.}},
\bauthor{\bsnm{Li}, \binits{C.}},
\bauthor{\bsnm{Zhou}, \binits{H.}},
\bauthor{\bsnm{Niu}, \binits{Y.}},
\bauthor{\bsnm{Liu}, \binits{Y.}}:
\bctitle{Gobigger: A scalable platform for cooperative-competitive multi-agent
  interactive simulation}.
In: \bbtitle{The Eleventh International Conference on Learning Representations}
(\byear{2023})
\end{bchapter}
\endbibitem

\end{thebibliography}

\end{document}